\documentclass[]{fairmeta}
\usepackage[table,xcdraw,dvipsnames]{xcolor}
\usepackage{wrapfig}
\usepackage{mathpazo}
\usepackage{tgpagella}
\usepackage{bbding}
\usepackage{graphicx}
\usepackage{pgfplots}
\usepackage{makecell}
\usepackage{enumitem}
\usepackage{mathtools}
\usepackage{xfrac}
\usepackage{nicefrac}
\usepackage{amssymb}
\newcolumntype{L}[1]{>{\raggedright\let\newline\\\arraybackslash\hspace{0pt}}m{#1}}
\newcolumntype{C}[1]{>{\centering\let\newline\\\arraybackslash\hspace{0pt}}m{#1}}
\newcolumntype{R}[1]{>{\raggedleft\let\newline\\\arraybackslash\hspace{0pt}}m{#1}}
\newcolumntype{Y}{>{\centering\arraybackslash}X}
\usepackage{lipsum}
\usepackage{algorithm}
\usepackage{algpseudocode}

\title{Mixture of States: Routing Token-Level Dynamics for Multimodal Generation}
\author[1,2,\dagger]{Haozhe Liu}
\author[2 \dagger]{Ding Liu}
\author[1,2, \ast]{Mingchen Zhuge}
\author[2, \ast]{Zijian Zhou}
\author[2, \ast]{Tian Xie}
\author[2]{Sen He}
\author[2]{Yukang Yang}
\author[1, 2]{Shuming Liu}
\author[2]{Yuren Cong}
\author[2]{Jiadong Guo}
\author[2]{Hongyu Xu}
\author[2]{Ke Xu}
\author[2]{Kam-Woh Ng}
\author[2]{Juan~C.~Pérez}
\author[2]{Juan-Manuel~Pérez-Rúa}
\author[2]{Tao Xiang}
\author[2]{Wei Liu}
\author[2, \ast]{Shikun Liu}
\author[1, \ast]{J\"{u}rgen Schmidhuber}
\affiliation[1]{KAUST}
\affiliation[2]{Meta AI}
\contribution[\dagger]{Joint First Authors}
\contribution[\ast]{Core Contributors}

\abstract{
We introduce MoS (Mixture of States), a novel fusion paradigm for multimodal diffusion models that merges modalities using flexible, state-based interactions. The core of MoS is a learnable, token-wise router that creates denoising timestep- and input-dependent interactions between modalities' hidden states, precisely aligning token-level features with the diffusion trajectory. This router sparsely selects the top-$k$ hidden states and is trained with an $\epsilon$-greedy strategy, efficiently selecting contextual features with minimal learnable parameters and negligible computational overhead. We validate our design with text-to-image generation (MoS-Image) and editing (MoS-Editing), which achieve state-of-the-art results. With only 3B to 5B parameters, our models match or surpass counterparts up to $4\times$ larger. These findings establish MoS as a flexible and compute-efficient paradigm for scaling multimodal diffusion models.
}

\date{\today}
\correspondence{\email{haozhe.liu@kaust.edu.sa}; \email{dingliu@meta.com} (Project Lead)}

\begin{document}

\maketitle
\begin{figure}[ht!]
    \centering
    \begin{subfigure}[t]{0.495\linewidth}
        \includegraphics[width=\linewidth]{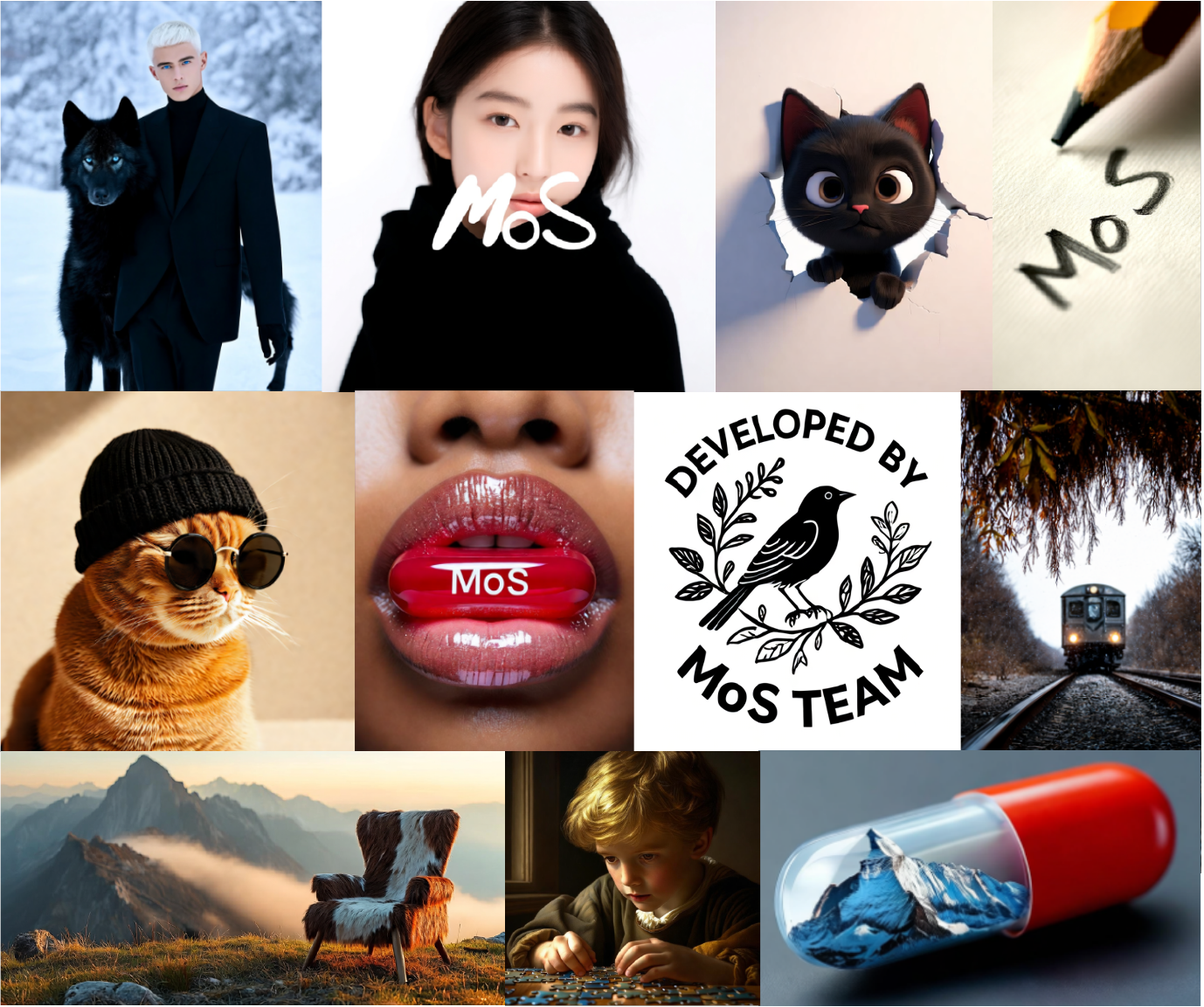}
        \caption*{MoS-Image}
    \end{subfigure}\hfill
    \begin{subfigure}[t]{0.495\linewidth}
        \includegraphics[width=\linewidth]{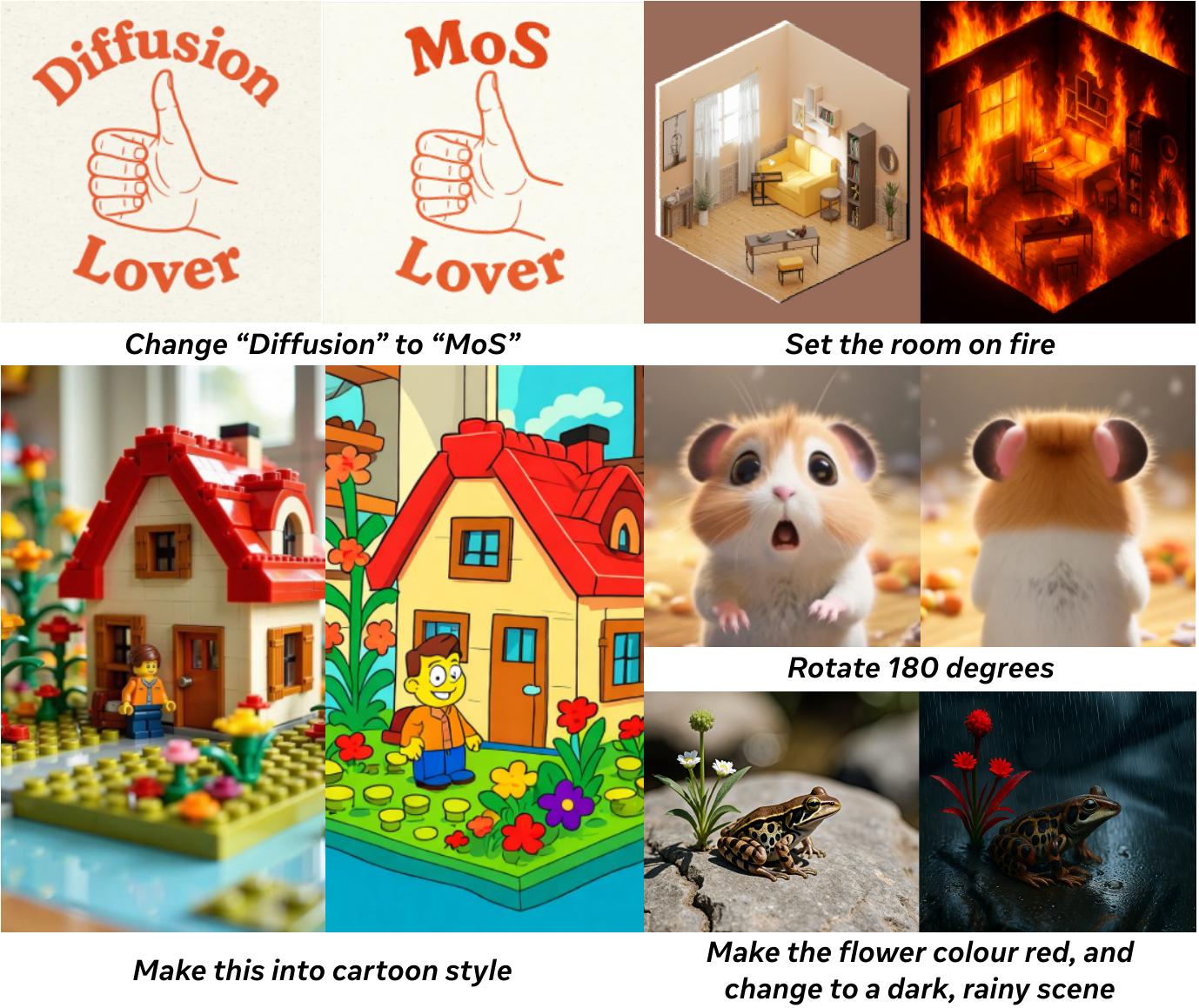}
        \caption*{MoS-Edit}
    \end{subfigure}\hfill
    \caption{\textbf{Generation examples by MoS-Image (left) and MoS-Edit (right).} MoS introduces a learnable, token-wise router that efficiently aggregates feature states across modalities. This allows for high-quality visual synthesis, producing photorealistic and stylized outputs from text and image inputs with precise control and quality. }
    \label{fig:highlight}
\end{figure}

\section{Introduction}
Multimodal generation is a fundamental application of modern AI, enabling models to synthesize high-quality visual content such as images and videos from conditional inputs \citep{rombach2022high,gao2025seedream,baldridge2024imagen,polyak2024movie,wu2025qwen,liu2025mardini}. In this paper, we focus on \textbf{text-to-image generation} and \textbf{instruction-based image editing} tasks, a domain where a central challenge lies in {\it effectively aligning textual and visual signals}. This problem is non-trivial, as the modalities rely on different modeling objectives: text models are typically trained with contrastive learning \citep{radford2021learning}, masked-token prediction \citep{devlin2019bert, raffel2020t5}, or next-token prediction \citep{team2024gemma,touvron2023llama1,brown2020language}, whereas visual models often adopt diffusion-based generation \citep{ho2020denoising,neal2001annealed,jarzynski1997equilibrium} or flow matching \citep{lipman2023flow,esser2024sd3}. Consequently, alignment requires bridging not only heterogeneous representations but also distinct designs across modalities. 

Prior studies address this challenge through various hand-crafted designs. These dominant fusion techniques, as well as the method we propose, are built upon the transformer architecture \citep{vaswani2017transformer,dosovitskiy2020vit,schmidhuber1992learning}, which serves as a powerful backbone for modeling both textual and visual representations. The primary strategies include: i) \textbf{Cross-attention} methods \citep{rombach2022high,vaswani2017transformer,chen2023pixart,xie2024sana} insert new attention blocks into the visual model, projecting text embeddings onto key-value vectors to enable cross-modal token interactions. ii) \textbf{Self-attention} methods \citep{esser2024scaling,chen2025dit,qin2025lumina} instead concatenate text and visual tokens into a unified sequence, processed by shared attention layers. While this allows for deeper, bidirectional fusion than cross-attention and often yield stronger performance, its computational cost is often prohibitive, scaling quadratically with the combined sequence length; iii) \textbf{MoT (Mixture-of-Transformers)}, a more recent method \citep{deng2025emerging,liao2025mogao,liang2025mixtureoftransformers,shi2024lmfusion}, establishing layer-wise cross-modal connections by sharing key–value vectors between corresponding text and visual blocks. This method facilitates a finer-grained interaction, but its rigid, layer-by-layer design imposes a strong architectural constraint: the text and visual backbones must be {\it symmetric, with a one-to-one block correspondence.}

Through our ablations, we have identified three critical design principles for improving text-visual representation alignment that challenge the hand-crafted/fixed-interaction paradigms of prior work:
\begin{itemize}
\item \textit{Layer selection should be adaptive, not fixed.}
We find that using a single fixed layer, typically the final-layer feature from the text branch, as commonly adopted in cross- and self-attention methods, is suboptimal. Furthermore, the rigid one-to-one layer correspondence of MoT assumes that text and visual features align symmetrically, an assumption for which we find no experimental support. This suggests that diffusion models do not consume language features in a strictly sequential or layer-aligned manner, making a {\it flexible selection} mechanism essential.
\item \textit{Conditional signals should be dynamic and timestep-dependent.} 
We validate that the common design in modern text-to-image models, which encodes the text embedding once and keep it static, creates an "information mismatch" with the dynamic nature of the denoising process. We argue that the conditional guidance needs to {\it adapt as the input noise level and denoising step change}.
\item \textit{Conditional signals should be token-specific.}
Our findings indicate that it is more effective to allow each token to source its representation adaptively from different layers, rather than using a single, shared layer embedding to represent all tokens uniformly. This supports a {\it more granular, token-level view} of context conditioning.
\end{itemize}

\begin{figure}[t!]
    \centering
    \begin{subfigure}[t]{0.23\linewidth}
        \includegraphics[width=\linewidth]{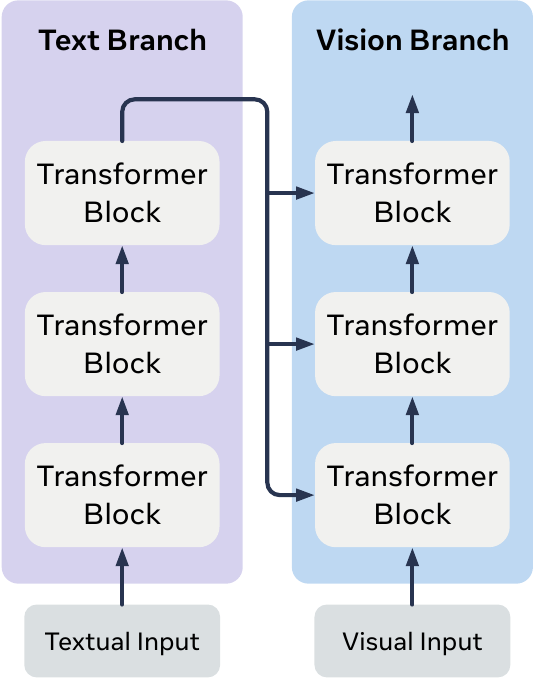}
        \caption{Cross-Attention}
    \end{subfigure}\hfill
    \begin{subfigure}[t]{0.23\linewidth}
        \includegraphics[width=\linewidth]{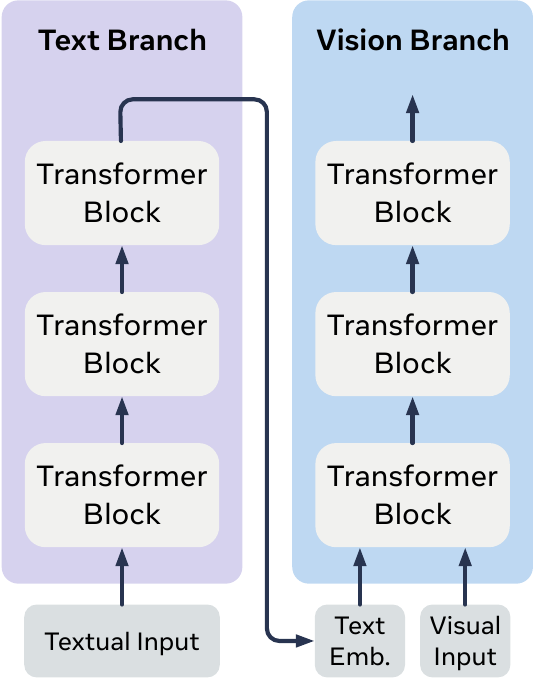}
        \caption{Self-Attention}
    \end{subfigure}\hfill
    \begin{subfigure}[t]{0.23\linewidth}
        \includegraphics[width=\linewidth]{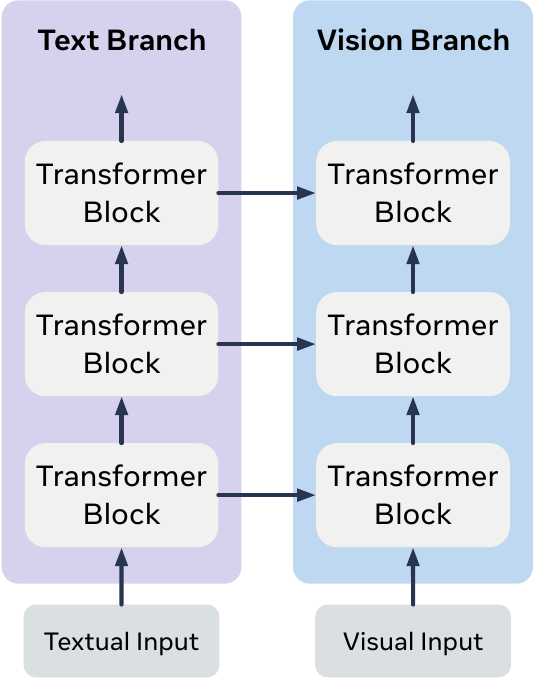}
        \caption{MoT}
    \end{subfigure}\hfill
    \begin{subfigure}[t]{0.253\linewidth}
        \includegraphics[width=\linewidth]{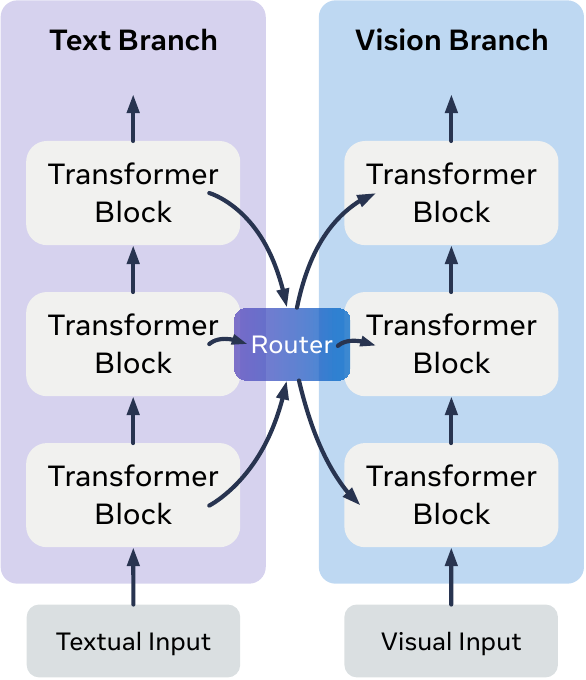}
        \caption{MoS (Ours)}
    \end{subfigure}
    \caption{\textbf{MoS enables sparse and dynamic interactions across modalities and transformers.} We illustrate MoS with text-to-image generation. Previous approaches, such as \textbf{(a) cross-attention} and \textbf{(b) self-attention}, typically provide only the final text encoder block's embedding as input to the visual branch, limiting the richness of cross-modal information. \textbf{(c) MoT (Mixture-of-Transformers)} attempts finer-grained interaction by passing outputs from all text blocks in a rigid, layer-by-layer fashion. In contrast, our proposed \textbf{(d) MoS (Mixture of States)} employs a learnable sparse interaction that dynamically links {\it any} text block to {\it any} visual block. The routing adapts to the current input, comprising the text prompt, visual latents, and denoising step embeddings, enabling flexible and efficient multimodal fusion. 
    }
    \label{fig:motivation}
    \vspace{-0.6cm}
\end{figure}

To this end, we introduce Mixture of States (MoS), a new framework that enables multimodal interaction to adapt dynamically to the input and denoising step.
Unlike prior fixed-interaction designs, MoS grants the vision branch access to {\it all textual hidden states} across {\it all layers} and employs a learnable, token-wise router to selectively aggregate features at each denoising step. As shown in Fig.~\ref{fig:motivation}, this sparse yet dynamic routing allows {\it any visual token, at any denoising step and within any transformer block, to attend to tokens from any layer of the text encoder.} This effectively bridges the gap between textual representations and visual diffusion dynamics.

We systematically explore the design space of MoS, including its input formulation, architectural configuration, and training strategy. Building on this foundation, we present a family of multimodal generation models that support two multimodal generation tasks: image generation (MoS-Image) and image editing (MoS-Edit).

\textbf{In all, our contributions are threefold:}
\begin{itemize}
\item We propose {\it Mixture of States}, a novel and flexible fusion mechanism for multimodal diffusion models. Its design, characterized by dynamic, sparse, and state-based interactions, is inherently adaptive and aligns with the iterative diffusion process.
\item We introduce {\it MoS-Image} and {\it MoS-Edit} (3-5B parameters) that serve as a blueprint for scaling the MoS design. We demonstrate its ability to effectively merge asymmetric text and visual backbones, overcoming a key limitation of prior fusion methods.
\item Extensive evaluations show that MoS-based models achieve state-of-the-art performance on image generation and editing tasks at a similar parameter scale. Notably, our 5B parameter model matches or surpasses the performance of a 20B-parameter ($4\times$ larger) model \citep{wu2025qwen}, demonstrating exceptional computational efficiency.
\end{itemize}

\section{Related Work}
\paragraph{Text-to-Image Network Architecture}
Diffusion models \citep{ho2020denoising,neal2001annealed,jarzynski1997equilibrium,peebles2023scalable} have become a dominant paradigm for text-to-image generation \citep{rombach2022high,dhariwal2021diffusion,ramesh2022hierarchical,chen2024gentron,saharia2022photorealistic,dai2023emu,betker2023improving} owing to their scalability and stable training. In multimodal diffusion models, prompt embeddings derived from a frozen text encoder are incorporated through cross-attention \citep{chen2024gentron,podell2023sdxl,rombach2022high}, self-attention \citep{esser2024scaling,chen2025dit}, or layer-wise attention (e.g., MoT) \citep{liang2025mixtureoftransformers}.  A fundamental challenge in this design is the "static vs. dynamic" mismatch. The diffusion process is inherently dynamic, operating over numerous timesteps with varying noise levels and visual features \citep{liu2025faster,kahatapitiya2024adaptive}. However, the text encoder provides only a single, static representation of the prompt. While self- and layer-wise attention allow this conditional information to evolve within the visual backbone's blocks, the initial conditional signal provided to the model remains fixed. To address this limitation, our MoS framework introduces a learnable router. 
The router jointly considers the prompt, denoising step, and noised image to dynamically select and aggregate conditional embeddings, enabling true input- and time-dependent conditioning.

\paragraph{Unified Model}
Recent research \citep{team2024chameleon,wu2024next,ge2024seed,xie2024show} has increasingly sought to unify diverse tasks within a single framework. While following this unified design philosophy, it diverges in its training methodology. Instead of the common approach of jointly training all tasks in a single, complex stage \citep{deng2025emerging,xie2025show,zhou2024transfusion,chen2025janus,liao2025mogao,geng2025x}, we adopt a multi-stage training strategy. This approach provides significant flexibility and efficiency. Specifically, by freezing our text branch, we can focus computational resources purely on optimizing the multimodal generation components. This staged training also circumvents common challenges of joint training, such as throughput bottlenecks from mixed data batches and the difficulty of balancing diverse, and often competing, learning objectives across modalities. This strategy is consistent with other recent, successful large-scale models like BLIP-3o \citep{chen2025blip3}, MetaQuery \citep{pan2025transfer}, Qwen-Image \citep{wu2025qwen} and LMFusion \citep{shi2024lmfusion}, which also employ multi-stage training.

\paragraph{Dynamic Neural Networks}
MoS is also related to the principles of dynamic neural networks \citep{han2021dynamic,schmidhuber1992learning,jacobs1991adaptive,hampshire1989connectionist,ivakhnenko1966cybernetic,ivakhnenko2007polynomial}, in which the computational graph or parameter usage is conditioned on the input. A prominent example is the Mixture-of-Experts (MoE) \citep{hampshire1989connectionist,eigen2013learning,shazeer2017outrageously,lepikhin2020gshard,fedus2022switch,liu2023prismer,jiang2024mixtral,csordas2024moeut}, where tokens are adaptively processed by different "expert" sub-networks within each transformer block. Due to its sparsity, MoE efficiently scales model parameters while keeping computation tractable. Recent advances have explored other forms of dynamic computation, such as Mixture-of-Depths (MoD) \citep{raposo2024mixture}, which dynamically allocates compute across tokens and layers, and Mixture-of-Recursions (MoR) \citep{bae2025mixture}, which reuses layers recursively with variable-depth routing. These methods establish a shared, powerful principle: computation should be sparse, adaptive, and conditional on the input. However, this principle has largely been applied to {\it intra-model adaptivity}, i.e., routing tokens within a single large model. In contrast, MoS extends this principle to {\it inter-model collaboration}.

\paragraph{Mixture of Transformers (MoT)}
MoS is conceptually related to the Mixture of Transformers (MoT) architecture \citep{liang2025mixtureoftransformers}. In MoT, each modality is processed by an independent transformer, while a shared attention module in each block enables tokens to attend across modalities. This design has been widely adopted in multimodal research: LMFusion \citep{shi2024lmfusion} and PGV3 \citep{liu2024playground} use it to couple a frozen LLMs as the text encoder with a trainable diffusion transformer for strong text-to-image generation performance. More recently, Bagel \citep{deng2025emerging} and Mogao \citep{liao2025mogao} enable joint training of both transformers, unifying image understanding and generation.
Despite these advances, a critical limitation persists: MoT designs require identical hidden dimensions and a strict one-to-one block correspondence across modalities to share global attention. This rigid, symmetric constraint is highly inflexible, as various modalities may follow distinct scaling laws and design principles. MoS is designed to solve this specific problem. We replace the rigid global attention mechanism with a learnable, sparse router, which removes the identical-size constraint and enables adaptive, effective interactions between {\it asymmetric} transformers.

\section{Mixture of States: Unifying Asymmetric Transformers}
\subsection{Design Overview}
MoS adopts a dual-tower architecture design for multimodal generation: 
an {\it understanding} tower $\mathcal{U}$ and a {\it generation} tower $\mathcal{G}$.\footnote{We deliberately choose this more general terminology rather than the terms \textit{text} or \textit{visual} branch.  MoS is designed to support both image generation and image editing tasks, where the latter requires both branches to process the reference image. For conceptual rigor, we thus avoid using terms such as \textit{text branch}.} The understanding tower $\mathcal{U}$ processes multimodal context $c$ (text for text-to-image tasks; text+image for image editing tasks), producing contextual representations that guide the generation tower $\mathcal{G}$ during visual synthesis.  Our design departs from prior work in two key ways. First, unlike MoT, which learns multi-modal interactions with the key-value vectors in the same layer, we use full layer-level hidden states as the unit of interaction. This enables state transfer compatibility across heterogeneous transformers with different model depths. Second, rather than connecting $\mathcal{U}$ and $\mathcal{G}$ through conventional hand-crafted feature fusing mechanisms like self- or cross-attention, we introduce a learnable router $\mathcal{R}$ to dynamically mediate their interaction.

Following standard practice, the entire model is trained end-to-end with rectified flow matching: 
\begin{align}
\mathbb{E}_{c,t,z_0, z_1} \bigg[\big\lVert \mathcal{G}(z_t, t, \mathcal{R}(t, c, z_t, \mathcal{U}(c))) - v_t \big\rVert_2^2\bigg],
\end{align}
where $v_t=\nicefrac{dz_t}{dt}= z_1 - z_0$ denotes the target velocity at denoising step $t\in[0,1]$. The input $z_t$ is the linear interpolation between the image latent $z_0$ and the random noise $z_1\sim\mathcal{N}(0,I)$, given by $z_t = (1-t)z_0 + t z_1$. The target latent $z_0$ is obtained from the VAE encoder $z_0=\mathcal{E}(x)$. During training, the parameters of the pre-trained understanding tower $\mathcal{U}$ are frozen. Only the generation tower $\mathcal{G}$ and the router $\mathcal{R}$ are learnable and are initialized from scratch. This setup significantly reduces computational overhead and facilitates more tractable architectural exploration. 

\begin{figure}[t!]
    \centering
    \includegraphics[width=\linewidth]{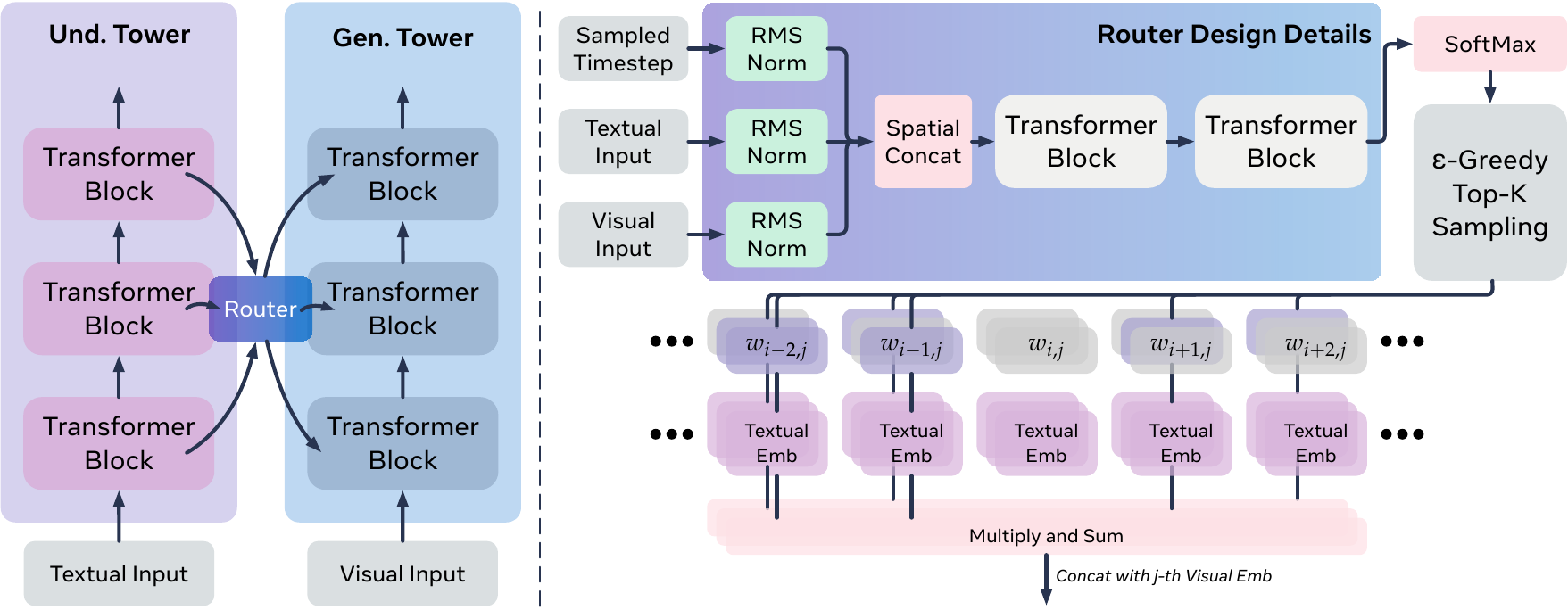}
    \caption{\textbf{MoS Design Details.} MoS introduces a new paradigm for multimodal interaction within transformer architectures. Rather than depending on manually designed fusion strategies, MoS employs a learned router to establish token-level sparse and dynamic connections between transformer blocks. For illustration, we use image generation as the running example and thus refer to the understanding-tower features as textual embeddings.
}
    \label{fig:main_pipeline}
\end{figure}

\subsection{MoS Design Details}
As shown in Fig.~\ref{fig:main_pipeline}, the MoS router $\mathcal{R}$ is the core component that governs collaboration between the two towers. It determines which hidden states from the understanding tower are transferred, how they are weighted, and to which layers of the generation tower they are delivered. We elaborate on this design along the following dimensions:

\paragraph{Router Input Space}
The router’s primary objective is to resolve the inherent {\it representation mismatch} between the understanding and generation towers. The understanding tower $\mathcal{U}$ processes the context $c$ in a single forward pass, producing a fixed set of representations. The generation tower $\mathcal{G}$, however, is highly dynamic, requiring step-specific guidance as its visual latent state $z_t$ evolves at each denoising timestep $t$. To address this, we incorporate the denoising step $t$  as a direct input to the router, enabling it to learn time-varying routing patterns. With additional input content, the complete router decision includes the timestep $t$, the noised image embedding $z_t$, and the context embedding $c$.  As the router is a lightweight transformer, it expects all inputs to be token sequences sharing a unified hidden size. 
The conditioning signal $c$ is processed through $\mathcal{U}$’s shared projection layer, followed by a linear layer for dimensional alignment. 
The timestep $t$ is represented using sinusoidal embeddings \citep{ho2020denoising} and projected into the same latent space. 
The latent $z_t$ shares $\mathcal{G}$’s patchify layer and is subsequently projected to the target dimensionality.

\paragraph{Router Output Space}
For each token in the context prompt, the router generates a corresponding logit matrix $\mathcal{W} = [w_{ij}] \in \mathbb{R}^{m\times n}$, where $m$ and $n$ denote the depths of the understanding and generation towers, respectively. Each entry $w_{ij}$ represents the learned affinity weight for routing the hidden states from the $i$-th layer of the understanding tower to the $j$-th layer of the generation tower. Building on prior analyses showing token-wise representations evolve and serve different roles across layers \citep{raposo2024mixture}, we hypothesize that routing decisions should also be token-specific. Therefore, in MoS, each token predicts its own distinct routing matrix $\mathcal{W}$, rather than sharing a single, uniform routing policy across the entire prompt.

\paragraph{Lightweight Router Design} 
To ensure efficiency, we adopt a lightweight router design. All input embeddings ($t$, $z_t$, and $c$)  are individually tokenized, normalized, and concatenated into a sequence. We then apply two transformer blocks with bidirectional self-attention to capture in-context semantics. Finally, we apply a projection layer on the context tokens to obtain the logit matrix. 

\paragraph{Sparsity and $\epsilon$-Greedy Exploration} 
In MoS, the router predicts the logit matrix to estimate the individual contributions of a context token’s hidden states from different blocks to the generation task. We expect the logit matrix to be sparse, as simply averaging all hidden states would obscure layer-specific information. To address this, we implement a sparse top-$k$ routing strategy.
Specifically, in each forward pass, the predicted logits $w_{ij}$ (for $i \in [1, m]$) are normalized using softmax. It then selects the top-$k$ hidden states from the understanding tower with the highest normalized weights. These selected hidden states are then reweighted and aggregated by their corresponding $w_{ij}$ to provide guidance to the $j$-th block of the generation tower. This selection process is performed independently for each generation block.

To prevent the router from converging to a sub-optimal local solution, we adopt an $\epsilon$-greedy strategy during training. At a pre-defined probability $\epsilon \in [0,1]$, the router selects $k$ random layers to encourage exploration; and with probability $1-\epsilon$, it follows the top-$k$ strategy described above.

\subsection{MoS for Image Generation}
For the text-to-image generation task, the generation tower follows the foundations of the diffusion process. 
During training, given a text-image pair $(c, z_0)$, we first extract all $m$ hidden states $\mathcal{S}^c_{i=1:m}$ from the understanding tower conditioned our input context prompt $c$:
\begin{align}
\mathcal{U}(c) = \{\mathcal{S}^c_i \mid i \in [1, m]\}.
\end{align}
A diffusion timestep $t$ is then sampled, yielding the corresponding noisy latent $z_t$. 
Conditioned on $c$, $t$, and $z_t$, the router predicts a logit matrix $\mathcal{W}$. By applying a softmax operation along each $j$-th column $w_{:,j}$, we obtain the final normalized weight matrix $\overline{\mathcal{W}} = [\overline{w}_{i,j}]$:
\begin{align}
    \overline{w}_{i,j} = \frac{\exp(w_{i,j})}{\sum_{i=1}^{m} \exp(w_{i,j})}.
\end{align}
For each $j$-th block in the generation tower $\mathcal{G}$, we identify the index set $I_j = \text{top-}k_\epsilon(\overline{w}_{1:m,j})$ using $\epsilon$-greedy top-$k$ selection. The conditional input is then computed as a weighted sum:
\begin{align}
    \mathbf{S}^c_j = \sum_{i \in I_j} \overline{w}_{ij} \cdot \mathcal{S}^c_i.
\end{align}
The aggregated hidden state $\mathbf{S}^c_j$ is passed through a linear projection layer, $\text{Proj}(\cdot)$, to match the dimensionality of the visual representation $\mathbf{S}^z_j$ at the $j$-th block. The projected features, $\text{Proj}(\mathbf{S}^c_j)$, are then concatenated with $\mathbf{S}^z_j$ 
\begin{wrapfigure}{r}{0.5\textwidth}
    \centering
    \vspace{-1em}
    \includegraphics[width=0.5\textwidth]{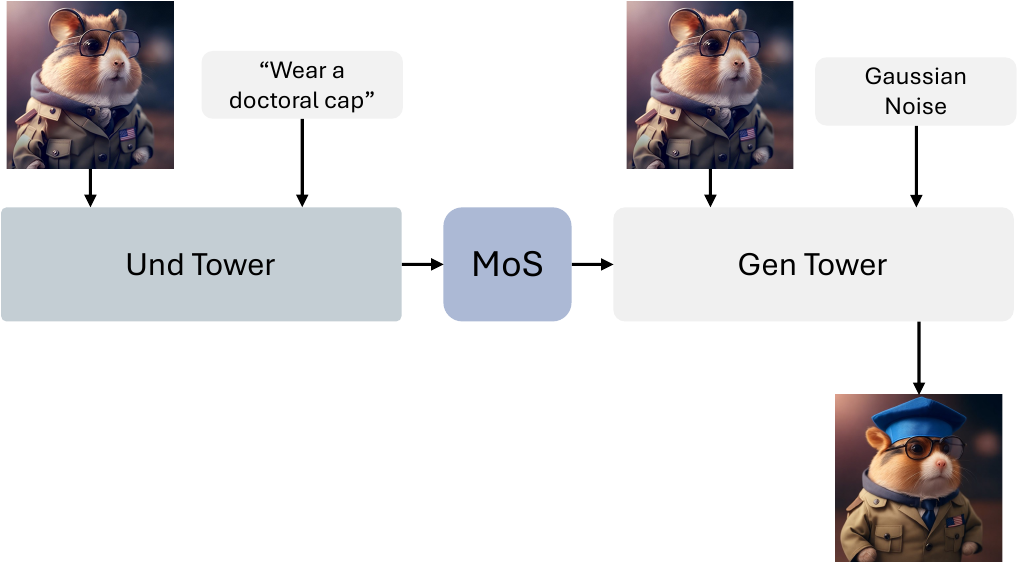} 
    \caption{\textbf{Image Editing Inference Pipeline.} Both the understanding and generation towers take the reference image as input, with their interaction facilitated through the MoS module.
    }
    \vspace{-4em}
    \label{fig:pipeline_image_editing}
\end{wrapfigure}
along the sequence dimension and jointly processed as in-context tokens within the block. During inference, this process is identical, but we set $\epsilon = 0$ with deterministic top-$k$ selection. The complete procedures for training and inference are summarized in Algorithms~\ref{alg:mos_training} and \ref{alg:mos_inference} in the Appendix.

\subsection{MoS for Image Editing}
Fig.~\ref{fig:pipeline_image_editing} illustrates the pipeline of MoS-Edit, following the same training strategy for MoS-Image but with a different input space. For this task, the model is given both a source reference image and a textual instruction (e.g., “wear a doctoral cap”). The understanding tower processes both the source image and the instruction to generate a rich, combined contextual representation $c$. This representation is then dynamically routed through MoS to the generation tower. 

During inference, the generation tower receives both the Gaussian noise (for the target image) and the clean reference image, and iteratively refines the latent representation conditioned on the guidance from MoS. Finally, the refined latent is decoded back into the final edited target image.

\subsection{Training MoS at Scale}
\paragraph{Training Strategy and Data Curation}
Following prior studies \citep{wu2025qwen, esser2024scaling}, we build our MoS-Image model upon the standard latent diffusion framework \citep{rombach2022high}, trained with a four-stage progressive schedule: 
\begin{itemize}
    \item Stage 1 (Low-Resolution): Leveraging a $16\times$ spatial compression rate from a pre-trained VAE \citep{wan2025wan} with a patch size of $2\times2$, we initiate training directly at a resolution of $512\times512$. This stage consumes 1400 A100-days on $\mathcal{O}(100\text{M})$ samples filtered by standard image quality metrics (e.g., resolution, caption length, aesthetic score).
    \item  Stage 2 (High Resolution): The model is then scaled to $1024\times1024$ resolution, training for a similar computational budget  on the same data. These first two stages establish the model's core capabilities on text-image alignment and high-fidelity image generation.
    \item Stage 3 (Aesthetic Tuning): We curate a high-quality subset of $\mathcal{O}(10\text{M})$ samples, with stricter data filtering on caption quality, aesthetic evaluation, and text-image alignment. The model is continually trained on these data for an additional 100 A100 days, enhancing its aesthetic fidelity and instruction-following capabilities.
    \item Stage 4 (Super-Resolution Tuning): We curate a high-quality dataset $\mathcal{O}(1\text{M})$ consisting of $2048 \times 2048$ resolution images. We fine-tune the model on this dataset to support 2K-resolution text-to-image generation, resulting in a final model with improved fidelity. This fine-tuning stage requires an additional 80 A100 GPU days.

\end{itemize}

Following this, we extend the model to MoS-Edit, which requires an additional 50 A100 days of training on $\mathcal{O}(1\text{M})$ paired image-editing data. In total, our MoS-Image model (Stages 1-4) requires $\sim$3,000 A100 days, substantially less than the 6,250 A100 days reported for one of the earliest large-scale text-to-image models, Stable Diffusion v1.5 \citep{rombach2022high}, demonstrating the efficiency of our architecture.

\paragraph{Model Configuration}
In this study, we develop two MoS variants: MoS-S and MoS-L.
Both models adopt the Wan 2.2 VAE \citep{wan2025wan} for compressing image inputs. Our MoS-S employs the 8B PLM-8B \citep{cho2025perceptionlm} as the frozen understanding tower and a 3B generation tower (trained from scratch); and our MoS-L employs the 14B InternVL-14B \citep{chen2024internvl} as the frozen understanding tower and a 5B generation tower (trained from scratch). For both variants, the router is designed with a lightweight 100M parameters, adding negligible overhead. For clarity, MoS-L/S refer to models with different parameter sizes, while MoS-Image/Edit denote variants targeting different tasks.

\section{Experiments}
\subsection{Systematic Ablations on Router Designs}
In this section, we systematically ablate the design choices of MoS. 
To balance computational cost, we conduct ablation studies with a lightweight configuration: models are trained on a randomly selected subset of pre-training data, and evaluation is performed on the MJHQ benchmark \citep{li2024playground}. 
We adopt FID \citep{heusel2017gans} and CLIP scores \citep{radford2021learning}, GenEval \citep{ghosh2023geneval} and DPG \citep{hu2024ella}, following a similar setup to \cite{esser2024scaling}.
Unless otherwise specified, all studies use an asymmetric design with LLaMA3.2-3B as the understanding tower and a 1B transformer for the generation tower. 
A key exception is the ablation comparing MoS with MoT. 
As MoT requires symmetric towers with identical depths, this specific study uses the smaller PLM-1B as the understanding tower, and instantiates an identical 1B architecture for the generation tower.

\paragraph{Router should be conditioned on dynamic, timestep-dependent signals.}
\begin{wraptable}{r}{0.3\textwidth}
\vspace{-1em}
\centering
\footnotesize
\setlength{\tabcolsep}{0.2em}  
\renewcommand{\arraystretch}{1.0}
\caption{\textbf{FID and CLIP results on MJHQ with different router's conditions.} Providing the router with the full dynamic state (prompt, noised latent, and timestep) yields the best performance.}
\label{tab:ab_input}
\begin{tabular}{ccccc}
\toprule
Prompt & Latent & Timestep & FID$\downarrow$ & CLIP$\uparrow$ \\
\midrule
      $\checkmark$                     &    $\times$   &    $\times$      & 21.12                   & 21.40                    \\
        $\checkmark$                    &   $\checkmark$      &     $\times$     & 21.89                   & 21.53                    \\
                           \rowcolor{metabg}
      $\checkmark$                    &   $\checkmark$      &     $\checkmark$         & \textbf{20.15}                   & \textbf{21.74}                   \\
                           \bottomrule
\end{tabular}
\vspace{-2em}
\end{wraptable}
We first validate our hypothesis that the router's conditional signal must be dynamic. 
We test three configurations: i) \textit{Prompt}: The router only receives the static prompt embedding $c$; ii) \textit{Prompt + Latent}: The router receives $c$ and the noised image latent $z_t$; iii) \textit{Prompt + Latent + Timestep}: The router receives the full dynamic states: $c$, $z_t$, and $t$.

The results in Table \ref{tab:ab_input} show a clear trend. 
Relying on the static prompt alone yields the weakest performance. 
Adding the dynamic state from the generation tower (the noised latent $z_t$ and timestep $t$) progressively improves both FID and CLIP scores. 
The optimal performance is achieved with the one that has access to all three inputs.

This confirms our design principle: the router must be aware of the diffusion process's dynamic state (noise level and timestep) to select the most effective features. 
This finding aligns with prior work \citep{liu2025faster}, showing that diffusion model behavior varies across denoising steps. 
We provide further validation in Appendix \ref{sec:router_visualize}, which visualizes how the router's guidance changes dynamically over different timesteps.

\begin{figure}[ht!]
\centering
\begin{minipage}[t]{0.70\textwidth}
\vspace{0pt}
    \centering
    \begin{subfigure}[t]{0.48\textwidth}
        \centering
        \includegraphics[width=\linewidth]{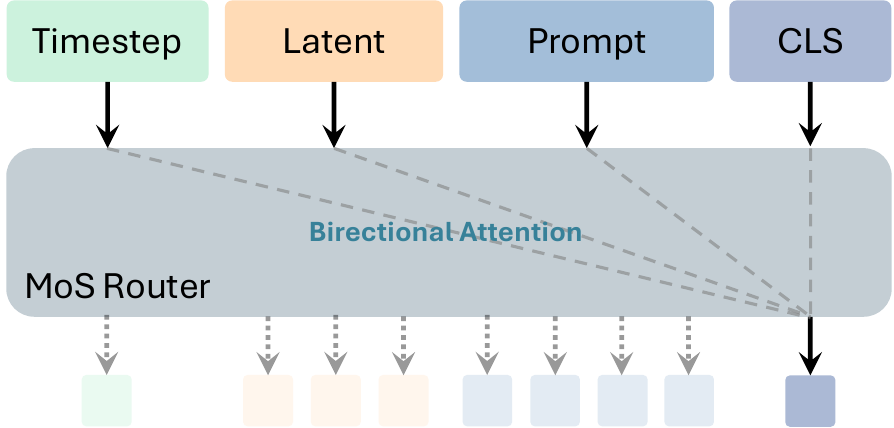}
        \caption{Sample-wise prediction.}
    \end{subfigure}\hfill
    \begin{subfigure}[t]{0.48\textwidth}
        \centering
        \includegraphics[width=\linewidth,trim=0 2 0 0,clip]{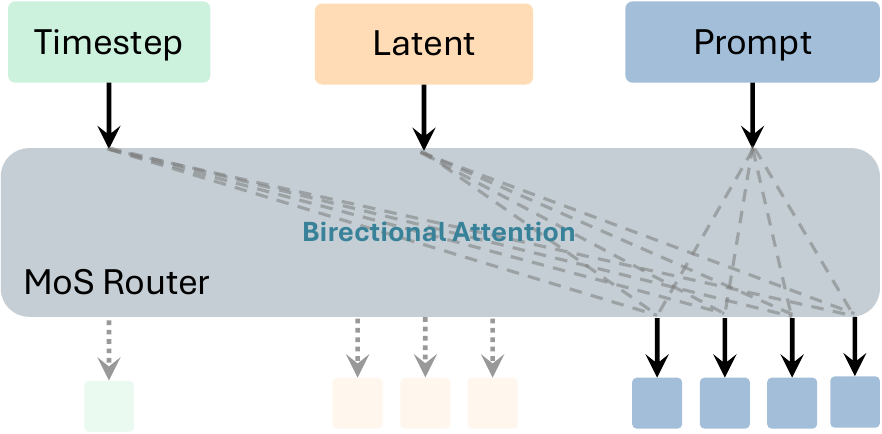}
        \caption{Token-specific prediction.}
    \end{subfigure}
    \caption{\textbf{Output configurations for the router.} (a) Predicts a single logit for the entire prompt; (b) predicts logits for each token.}
    \label{fig:output_space}
\end{minipage}
\hfill
\begin{minipage}[t]{0.28\textwidth}
\vspace{0pt}
    \centering
    \footnotesize
    \renewcommand{\arraystretch}{1.1}
    \captionof{table}{\textbf{FID and CLIP results on MJHQ with different router's prediction format.} The token-wise prediction consistently achieves superior results.}
    \vspace{1em}
    \begin{tabular}{lcc}
        \toprule
        & FID $\downarrow$ & CLIP $\uparrow$ \\
        \midrule
        Sample-wise & 21.66 & 21.48 \\
        \rowcolor{metabg}
        Token-specific & \textbf{20.17} & \textbf{21.63} \\
        \bottomrule
    \end{tabular}
    \label{tab:output_space}
\end{minipage}
\end{figure}

\paragraph{Router's prediction should be token-specific.}
Next, we investigate whether the router should predict a global logit matrix shared by all context tokens, or a token-specific matrix that assigns individualized routing decisions to each token.
As illustrated in Fig.~\ref{fig:output_space}, we compare two prediction formats for the router:
\begin{itemize}
\item \textit{Sample-wise Prediction}: We append a learnable \texttt{[CLS]} token to the input sequence. 
With full attention, this token’s final hidden state, which captures global context, is linearly projected to predict the logit matrix $\mathcal{W}$. 
In this setup, the entire prompt shares one routing solution.
\item \textit{Token-specific Prediction}: Instead of introducing a new token, we use the hidden states of the prompt embeddings directly. 
A shared linear projection maps each token’s representation to the logit matrix $\mathcal{W}$. 
Thus, each token in the prompt defines its own routing pattern.
\end{itemize}
As shown in Table~\ref{tab:output_space}, token-wise prediction outperforms the sample-wise alternative. 
This observation supports our hypothesis on token-level dynamics and aligns with prior studies \citep{bae2025mixture,raposo2024mixture} revealing that the behavior of tokens across LLMs varies significantly. 
We provide further validation in Appendix \ref{sec:router_visualize}, where visualizations show that the router naturally learns diverse, token-specific routing patterns without explicit regularization.

\begin{figure}[ht!]
\begin{minipage}[t]{0.20\textwidth}
\centering
\footnotesize
\setlength{\tabcolsep}{0.18em}  
\renewcommand{\arraystretch}{1.0}
\captionof{table}{\textbf{FID and CLIP results on MJHQ comparing hand-crafted routing and MoS.} MoS significantly outperforms the hand-crafted routing baseline.} %
\begin{tabular}{lcc}
\toprule
Model & FID $\downarrow$ & CLIP $\uparrow$ \\
\midrule
Hand-Crafted & 21.51 & 22.04 \\
\rowcolor{metabg}
MoS & \textbf{17.77} & \textbf{22.91} \\
\bottomrule
\end{tabular}
\label{tab:ab_handcrafted}
\end{minipage}
\hfill
\begin{minipage}[t]{0.25\textwidth}
\centering
\footnotesize
\setlength{\tabcolsep}{0.32em}  
\captionof{table}{\textbf{GenEval and DPG scores comparing the cross-attention baseline and MoS.} MoS consistently achieves better results than the cross-attention baseline.}
\begin{tabular}{lcc}
\toprule
Model & GenEval $\uparrow$ & DPG $\uparrow$ \\
\midrule
Cross-Attn & 0.74 & 83.40 \\
\rowcolor{metabg}
MoS & \textbf{0.79} & \textbf{85.61} \\
\bottomrule
\end{tabular}
\label{tab:ab_cross}
\end{minipage}
\hfill
\begin{minipage}[t]{0.5\textwidth}
\vspace{0pt}
\centering
\includegraphics[width=0.48\linewidth]{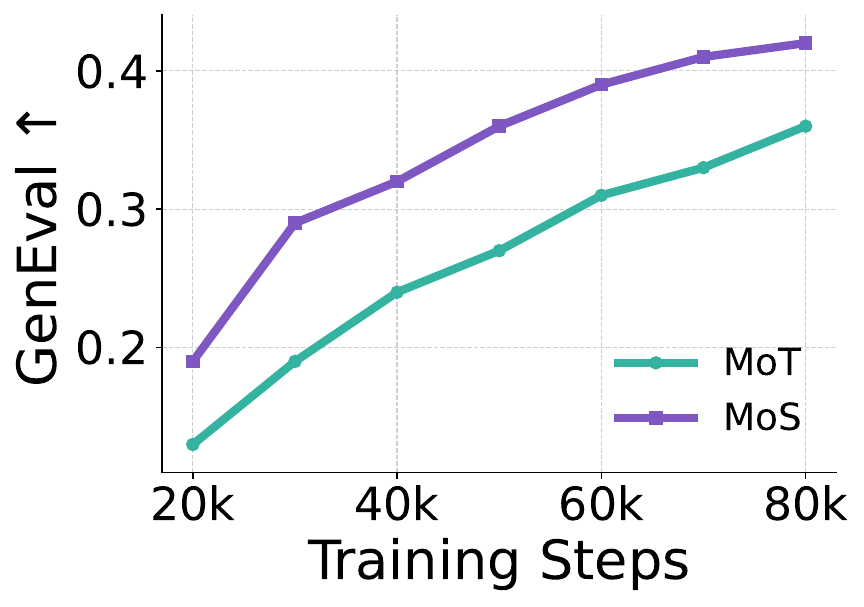}
\includegraphics[width=0.48\linewidth]{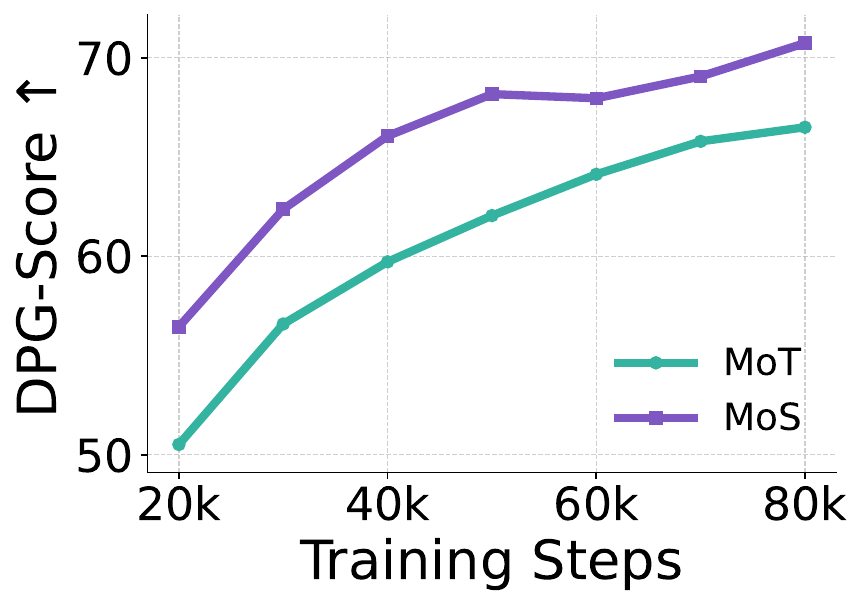}
\vspace{-0.5em}
\caption{\textbf{Ablation results on GenEval and DPG scores comparing the MoT baseline with MoS.} MoS consistently outperforms MoT across all training steps on both benchmarks.}
\label{fig:compare_with_mot}
\end{minipage}
\end{figure}

\paragraph{Layer selection should be adaptive, not fixed.}
To validate our hypothesis that features from different layers should be adaptively selected, we compare MoS against three fixed layer-selection baselines, under the same training recipe and data: 

\begin{itemize}
\item \textit{Hand-crafted Routing}: We first test a baseline with predefined, static connections (i.e., by evenly skipping a subset of text encoder layers). As shown in Table~\ref{tab:ab_handcrafted}, MoS surpasses this rigid hand-crafted design on both FID and CLIP, highlighting the importance of adaptive selection. A supplementary training-step analysis is in Appendix Fig.~\ref{fig:ab_78_results}.

\item \textit{Cross-Attention (Fixed-Layer)}: Next, we compare MoS with a 5B-parameter cross-attention model that uses customized text encoders \citep{xu2023demystifying,tay2022ul2,liu2024glyph}. Such models typically use a fixed, final-layer feature for conditioning. Table~\ref{tab:ab_cross} shows that while our cross-attention baseline is strong (0.74 GenEval, 83.40 DPG), MoS trained with the same data and parameter budget significantly outperforms it (0.79 GenEval, 85.61 DPG).

\item \textit{Mixture-of-Transformers (MoT)}: Finally, we compare MoS with MoT, which enforces a rigid one-to-one layer correspondence and has been shown to outperform cross- and self-attention \citep{tang2025exploring}. Under a fair comparison using identical parameters, data, and compute, Fig.~\ref{fig:compare_with_mot} shows that MoS consistently outperforms MoT across all training stages.
\end{itemize}

These quantitative results, supported by visualizations of the router’s learned patterns (Appendix \ref{sec:router_visualize}), demonstrate the clear superiority of adaptive, state-based selection. This aligns with prior work \citep{clark2019does,rai2024practical,raposo2024mixture}, which shows that token representations in LLMs play different functional roles across layers, implying that the fixed representation scheme might be inherently suboptimal.

\begin{figure}[ht!]
    \centering
    \begin{subfigure}{0.75\textwidth}
  \includegraphics[width=\linewidth]{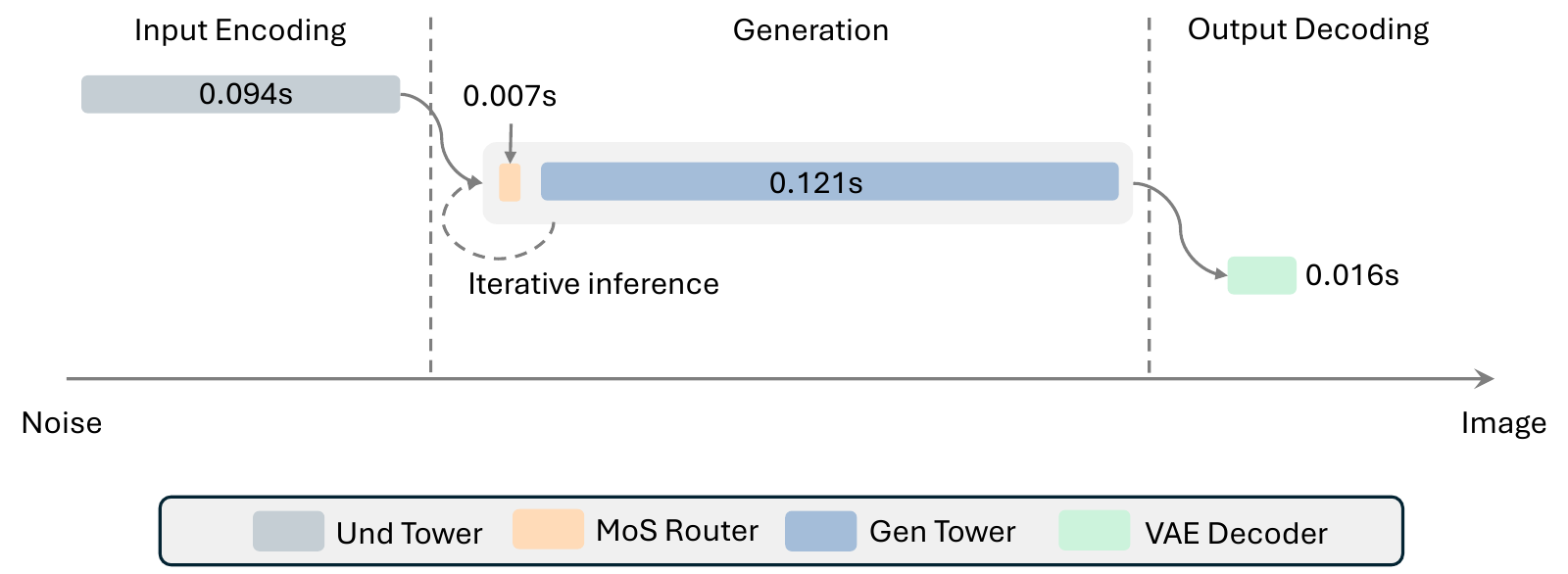}
  \caption*{(a) Latency contribution of individual components.}
    \end{subfigure}\hfill
    \begin{subfigure}{0.24\textwidth}
  \includegraphics[width=\linewidth]{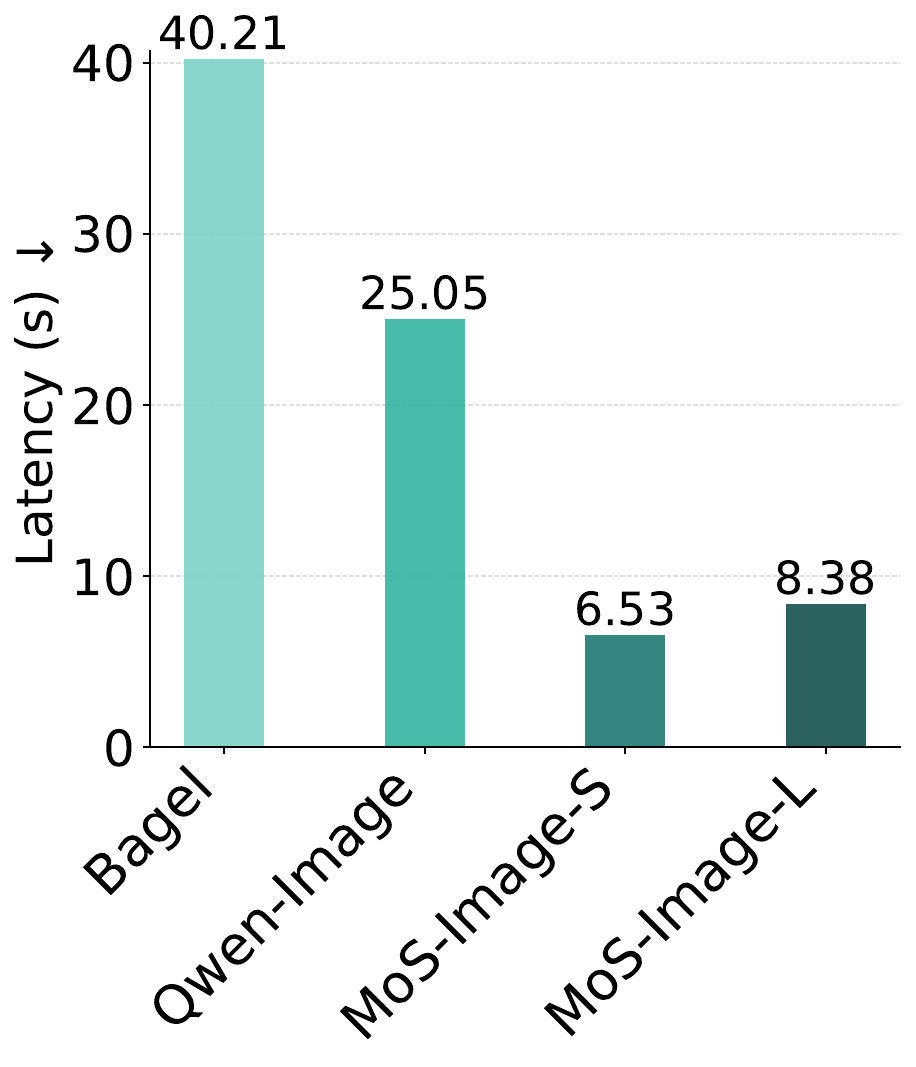}
  \caption*{(b) Latency comparison with other methods.}
    \end{subfigure}
\caption{\textbf{Latency analysis of MoS.} We evaluate the latency of MoS-Image-S for generating $1024^2$ images. The inference process includes three stages: (i) input encoding—embedding the input text and initializing noise; (ii) generation—iterative application of the generation tower and router; and (iii) decoding—transforming the latent code into the final image. We further compare MoS with existing methods on the same computational platform (a single A100 GPU). For reproducibility, baseline results are obtained using the default configurations in \texttt{diffusers} \citep{huggingface2022diffusers,JiaxinGe2025DiffusersBAGEL}. }
    \label{fig:efficiency}
\end{figure}

\paragraph{Router Efficiency}
The MoS router is designed to be a lightweight transformer, adding negligible computational overhead. We report end-to-end latency for generating a $1024\times 1024$ image on a single A100 GPU in Fig.~\ref{fig:efficiency}. With a 3B generation tower, the router contributes only 0.008s per iteration, a cost that is effectively insignificant. This fractional overhead becomes even smaller when MoS is paired with larger generation towers.

This lightweight design leads to improved overall latency performance. When benchmarked against two state-of-the-art image generation models: Qwen-Image \citep{wu2025qwen} and Bagel \citep{deng2025emerging}, MoS achieves lower end-to-end latency and thus higher throughput, demonstrating its efficiency.

\paragraph{Additional Discussions}
We provide extensive additional analyses in the Appendix. We further dissect the router's core design, including an ablation on which features are most beneficial to route (Appendix \ref{sec:opt_space}), an analysis of its architecture (Appendix \ref{sec:router_arch}), and an empirical validation of the $\epsilon$-greedy and top-$k$ selection strategies (Appendix \ref{sec:router_explore}). We also confirm the scalability and robustness of our framework: we verify that MoS adheres to standard scaling laws (Appendix \ref{sec:mos_scale}), demonstrate the specific advantages of our dual-tower design for image editing (Appendix \ref{sec:mos_editing}), and show that MoS benefits from common inference-time optimizations (Appendix \ref{sec:com_inf}).

\begin{table}[ht!]
\centering
\footnotesize
\setlength{\tabcolsep}{0.27em}  
\renewcommand{\arraystretch}{1.0}
\caption{\textbf{Performance of Foundational Diffusion Models on Text-to-Image and Image Editing.} The parameter size refers to the number of learnable parameters. This table summarizes state-of-the-art models across various multimodal interaction techniques. MoS achieves leading performance on most benchmarks and, in some cases, rivals product-level models with 20B or more parameters. The reported results are primarily drawn from \cite{wu2025qwen} and \cite{niu2025wise}.}
\label{tab:summary}
\begin{tabular}{llccccccc}
\toprule
\multirow{2}[2]{*}{SoTA Methods}   & \multirow{2}[2]{*}{\makecell[c]{Interaction\\ Type}}  & \multirow{2}[2]{*}{\makecell[c]{Model\\ Parameters}}  & \multicolumn{4}{c}{Image Generation}              &    \multicolumn{2}{c}{Image Editing}        \\
\cmidrule(lr){4-7} \cmidrule(l){8-9}
 &   &  & GenEval$^\uparrow_{[0\rightarrow1]}$ & DPG$^\uparrow_{[0\rightarrow100]}$   & WISE$^\uparrow_{[0\rightarrow1]}$ & oneIG$^\uparrow_{[0\rightarrow1]}$                         & GEdit$^\uparrow_{[0\rightarrow10]}$   & ImgEdit$^\uparrow_{[0\rightarrow5]}$           \\
\midrule

\textcolor{gray}{Qwen-Image\citep{wu2025qwen}}               & \textcolor{gray}{Self-Attn}   & \textcolor{gray}{\textbf{20B}}                   & \textcolor{gray}{0.87}    & \textcolor{gray}{\textbf{88.32}} & \textcolor{gray}{\textbf{0.62}} & \textcolor{gray}{\textbf{0.54}}                             & \textcolor{gray}{7.56}              & \textcolor{gray}{4.27}              \\
\midrule
SANA-1.5 \citep{xie2025sana}                & Cross-Attn              & 4.8B                  & 0.81    & 84.70 & -    & 0.33                             & -                 & -                 \\
FLUX.1{[}Dev{]} \citep{flux2024}         & Self-Attn  & 12B                   & 0.66    & 83.84 & 0.50 & 0.43                             & -                 & -                 \\
Bagel \citep{deng2025emerging}                   & MoT                     & 14B                   & 0.88    & -     & 0.52 & 0.36                             & 6.52              & 3.20              \\
\midrule
\rowcolor{metabg} 
MoS-S                   & MoS                     & 3B                    & 0.89    & 86.33 & 0.47 & 0.50                             & 7.41 & 4.17 \\
\rowcolor{metabg} 
MoS-L                   & MoS                     & 5B                    & \textbf{0.90}    & \textbf{87.01} & \textbf{0.54} & \textbf{0.52}  & \textbf{7.86}                 & \textbf{4.33}                \\
\bottomrule
\end{tabular}
\vspace{-1em}
\end{table}

\subsection{Results on Text-to-Image Generation}
We begin by evaluating MoS on standard text-to-image benchmarks. In Table \ref{tab:summary}, we report its performance on GenEval \citep{ghosh2023geneval}, DPG \citep{hu2024ella}, WISE \citep{niu2025wise}, and oneIG-EN \citep{chang2025oneig}. For comparison, we select 3 state-of-the-art methods representing different interaction types: SANA-1.5 (cross-attention) \citep{xie2025sana}, Flux.1[Dev] (self-attention) \citep{flux2024}, and Bagel (MoT) \citep{deng2025emerging}. We also include Qwen-Image \citep{wu2025qwen} for reference, which establishes a new state of the art for self-attention but uses more than four times the parameters of our model. The empirical results demonstrate that MoS-L, despite its smaller parameter count, consistently outperforms existing approaches across all benchmarks, highlighting the effectiveness of the MoS mechanism. 

In Appendix, we report full GenEval results in Table \ref{tab:geneval}, DPG-Bench results in Table \ref{tab:dpg_bench}, WISE results in Table \ref{tab:wisescore}, and OneIG performance in Table \ref{tab:oneig}. When benchmarking MoS, no prompt rewriting strategies (e.g., self-CoT \citep{deng2025emerging}) are applied; commercial models \citep{openai_dalle3_2025,kuai2025kolors2, recraft2024v3, google2025imagen, gao2025seedream,openai_gpt_image_1_2025} that may employ such techniques are shown in gray, as their exact generation pipelines and model sizes are unclear. Model size here refers only to learnable parameters, excluding text encoder or understanding tower parameters when frozen. Overall, MoS demonstrates strong performance, trailing only slightly behind Qwen-Image, which is substantially larger (20B vs 5B). Furthermore, Fig.~\ref{fig:vis_t2i} in the Appendix presents representative samples generated by MoS.

As shown in Fig.~\ref{fig:vis_t2i_comapare_with_others} (see Appendix), we further compare our generated samples with recent advanced methods. The selected examples are intentionally challenging for standard text-to-image models, as each contains more than four entities in a single scene, with fine-grained attributes such as location, color, and texture. To increase difficulty, we also include cases with dense visual text. The results show that our method aligns with input prompts more precisely and performs on par with, or in some cases surpasses, Qwen-Image, while significantly outperforming other baselines.

\subsection{Results on Instruction-based Image Editing}
We evaluate the image editing performance of MoS on two benchmarks: ImgEdit \citep{ye2025imgedit} and GEdit \citep{liu2025step1x_edit}. 
All evaluations are conducted automatically using GPT-4o \citep{hurst2024gpt}. The benchmarks cover a diverse set of dimensions, including object-level editing, scene-level editing, text editing and hybrid challenging cases.
As shown in Table~ \ref{tab:summary}, our 5B-parameter model achieves state-of-the-art performance on these two benchmarks. The detailed benchmarking results are provided in Table~\ref{tab:imgedit} for ImgEdit and Table~\ref{tab:gedit} for GEdit in the Appendix. We further provide visual comparisons on image editing tasks in Fig.~\ref{fig:vis_edit_comapare_with_others} (see Appendix), where MoS consistently produces results that precisely align with the given instructions and reference images, outperforming competing methods. 

\subsection{Results on Image Understanding}
Our model can be regarded as a two-stage framework for unifying the understanding and generation tasks: the first stage addresses understanding tasks, while the second focuses on generation. As the understanding tower is frozen in our study, its image/video understanding performance remains identical to previously reported results and is therefore omitted. Importantly, even without additional training, the understanding capacity strengthens the generation tower in several dimensions. For instance, with self-CoT \citep{deng2025emerging}, the understanding model progressively generates a reasoning-based caption, which then serves as contextual guidance for the generation tower to synthesize images. Following this approach, our performance on world knowledge reasoning improves on the WISE benchmark—from 0.54 to 0.65 for MoS-L and from 0.47 to 0.55 for MoS-S.

\section{Conclusion}
We present a new family of multimodal diffusion models based on Mixture of States (MoS) routing. By introducing a learnable router that dynamically selects hidden states across modalities, tokens, and denoising steps, MoS overcomes the rigid synchronization of prior architectures and supports asymmetric understanding and generation towers. This efficiency-oriented design executes the larger understanding model only once while dedicating the generation module to fine-grained detail synthesis. Extensive evaluations demonstrate that MoS achieves state-of-the-art performance across multiple benchmarks, rivaling or surpassing models up to four times larger, yet at a markedly reduced computational cost.
These results position MoS as both a practical and conceptual step toward scaling multimodal generative models, providing a flexible, efficient, and unified foundation for future research and deployment.

\section{Limitation and Future Studies}

\paragraph{One-Way to Dual-Way Setting.}
MoT has demonstrated strong scalability under early-fusion training. In contrast, while MoS shows promising results for multimodal generation, its effectiveness in early-fusion settings remains to be validated. A principled extension is to endow the router with multiple projection layers to establish bidirectional transformer connections. We defer this exploration to future work due to computational and data constraints.

\paragraph{Human Preference Alignment.} In this paper, we primarily adopt SFT as the post-training strategy for our models. Recent studies have explored applying CoT to multimodal generation \citep{guo2025can} or employing GRPO \citep{shao2024deepseekmath} to better align generated samples with human preferences \citep{liu2025flow}. Since our model’s behavior remains consistent with standard diffusion models, it can likewise benefit from such post-training techniques. We leave this direction for future work. 

\paragraph{Efficiency Improvement.} Our model is relatively smaller than prior state-of-the-art models, making it naturally more efficient. Nonetheless, it could be further accelerated through techniques such as low-precision quantization \citep{xie2024sana}, model distillation \citep{yin2024one}, or feature caching \citep{liu2025faster,kahatapitiya2024adaptive}. We leave these directions for future exploration.

\paragraph{Explainability.} The MoS router predicts the relative importance of each potential connection, which offers a basis for interpreting cross-modal interactions. While this property may provide insights into model explainability, such analysis lies beyond the scope of this work and is left for future investigation.

\paragraph{Visual Artifacts.} The primary goal of this paper is to address the challenge of instruction-following in multimodal generation. Nevertheless, our model still faces issues similar to other DiT or unified models, such as producing artifacts when the generated objects are very small (see our visualizations).

\section*{Acknowledgements}
The authors thank Yuhui Wang, Gordan (Guocheng) Qian, Jackson (Kuan-Chieh) Wang, Jinheng Xie, Enze Xie, Song Han, Chandan Akiti and Jinjie Mai for their valuable suggestions and contributions to the paper review. Haozhe Liu, Mingchen Zhuge, Shuming Liu and Jürgen Schmidhuber
were supported by funding from the King Abdullah University of Science and Technology (KAUST) - Center of Excellence for Generative AI under award number 5940 and the SDAIA-KAUST Center of Excellence in Data Science and Artificial Intelligence. 
\bibliographystyle{assets/plainnat}
\bibliography{references}

\begin{thebibliography}{119}
\providecommand{\natexlab}[1]{#1}
\providecommand{\url}[1]{\texttt{#1}}
\expandafter\ifx\csname urlstyle\endcsname\relax
  \providecommand{\doi}[1]{doi: #1}\else
  \providecommand{\doi}{doi: \begingroup \urlstyle{rm}\Url}\fi

\bibitem[AI and Team(2025)]{zhipu2025cogview4}
Zhipu AI and THUDM /~CogView Team.
\newblock Cogview4.
\newblock \url{https://github.com/THUDM/CogView4}, 2025.

\bibitem[Bae et~al.(2025)Bae, Kim, Bayat, Kim, Ha, Schuster, Fisch, Harutyunyan, Ji, Courville, et~al.]{bae2025mixture}
Sangmin Bae, Yujin Kim, Reza Bayat, Sungnyun Kim, Jiyoun Ha, Tal Schuster, Adam Fisch, Hrayr Harutyunyan, Ziwei Ji, Aaron Courville, et~al.
\newblock Mixture-of-recursions: Learning dynamic recursive depths for adaptive token-level computation.
\newblock \emph{arXiv preprint arXiv:2507.10524}, 2025.

\bibitem[Baldridge et~al.(2024)Baldridge, Bauer, Bhutani, Brichtova, Bunner, Castrejon, Chan, Chen, Dieleman, Du, et~al.]{baldridge2024imagen}
Jason Baldridge, Jakob Bauer, Mukul Bhutani, Nicole Brichtova, Andrew Bunner, Lluis Castrejon, Kelvin Chan, Yichang Chen, Sander Dieleman, Yuqing Du, et~al.
\newblock Imagen 3.
\newblock \emph{arXiv preprint arXiv:2408.07009}, 2024.

\bibitem[Batifol et~al.(2025)Batifol, Blattmann, Boesel, Consul, Diagne, Dockhorn, English, English, Esser, Kulal, et~al.]{batifol2025flux}
Stephen Batifol, Andreas Blattmann, Frederic Boesel, Saksham Consul, Cyril Diagne, Tim Dockhorn, Jack English, Zion English, Patrick Esser, Sumith Kulal, et~al.
\newblock Flux. 1 kontext: Flow matching for in-context image generation and editing in latent space.
\newblock \emph{arXiv e-prints}, pages arXiv--2506, 2025.

\bibitem[Betker et~al.(2023)Betker, Goh, Jing, Brooks, Wang, Li, Ouyang, Zhuang, Lee, Guo, et~al.]{betker2023improving}
James Betker, Gabriel Goh, Li~Jing, Tim Brooks, Jianfeng Wang, Linjie Li, Long Ouyang, Juntang Zhuang, Joyce Lee, Yufei Guo, et~al.
\newblock Improving image generation with better captions.
\newblock \emph{Computer Science. https://cdn. openai. com/papers/dall-e-3. pdf}, 2\penalty0 (3):\penalty0 8, 2023.

\bibitem[Brooks et~al.(2023)Brooks, Holynski, and Efros]{brooks2023instructpix2pix}
Tim Brooks, Aleksander Holynski, and Alexei~A Efros.
\newblock Instructpix2pix: Learning to follow image editing instructions.
\newblock In \emph{{Proceedings of the {IEEE} Conference on Computer Vision and Pattern Recognition ({CVPR})}}, pages 18392--18402, 2023.

\bibitem[Brown et~al.(2020)Brown, Mann, Ryder, Subbiah, Kaplan, Dhariwal, Neelakantan, Shyam, Sastry, Askell, et~al.]{brown2020language}
Tom Brown, Benjamin Mann, Nick Ryder, Melanie Subbiah, Jared~D Kaplan, Prafulla Dhariwal, Arvind Neelakantan, Pranav Shyam, Girish Sastry, Amanda Askell, et~al.
\newblock Language models are few-shot learners.
\newblock \emph{Advances in neural information processing systems}, 33:\penalty0 1877--1901, 2020.

\bibitem[Cai et~al.(2025)Cai, Chen, Chen, Li, Long, Pan, Qiu, Zhang, Gao, Xu, et~al.]{cai2025hidream}
Qi~Cai, Jingwen Chen, Yang Chen, Yehao Li, Fuchen Long, Yingwei Pan, Zhaofan Qiu, Yiheng Zhang, Fengbin Gao, Peihan Xu, et~al.
\newblock Hidream-i1: A high-efficient image generative foundation model with sparse diffusion transformer.
\newblock \emph{arXiv preprint arXiv:2505.22705}, 2025.

\bibitem[Chang et~al.(2025)Chang, Fang, Xing, Wu, Cheng, Wang, Zeng, Yu, and Chen]{chang2025oneig}
Jingjing Chang, Yixiao Fang, Peng Xing, Shuhan Wu, Wei Cheng, Rui Wang, Xianfang Zeng, Gang Yu, and Hai-Bao Chen.
\newblock Oneig-bench: Omni-dimensional nuanced evaluation for image generation.
\newblock \emph{arXiv preprint arxiv:2506.07977}, 2025.

\bibitem[Chen et~al.(2025{\natexlab{a}})Chen, Qian, Hu, Fu, Tong, Wang, Li, Zhang, Schwing, Liu, et~al.]{chen2025dit}
Chen Chen, Rui Qian, Wenze Hu, Tsu-Jui Fu, Jialing Tong, Xinze Wang, Lezhi Li, Bowen Zhang, Alex Schwing, Wei Liu, et~al.
\newblock Dit-air: Revisiting the efficiency of diffusion model architecture design in text to image generation.
\newblock \emph{arXiv preprint arXiv:2503.10618}, 2025{\natexlab{a}}.

\bibitem[Chen et~al.(2025{\natexlab{b}})Chen, Xu, Pan, Hu, Qin, Goldstein, Huang, Zhou, Xie, Savarese, et~al.]{chen2025blip3}
Jiuhai Chen, Zhiyang Xu, Xichen Pan, Yushi Hu, Can Qin, Tom Goldstein, Lifu Huang, Tianyi Zhou, Saining Xie, Silvio Savarese, et~al.
\newblock Blip3-o: A family of fully open unified multimodal models-architecture, training and dataset.
\newblock \emph{arXiv preprint arXiv:2505.09568}, 2025{\natexlab{b}}.

\bibitem[Chen et~al.(2023)Chen, Yu, Ge, Yao, Xie, Wu, Wang, Kwok, Luo, Lu, et~al.]{chen2023pixart}
Junsong Chen, Jincheng Yu, Chongjian Ge, Lewei Yao, Enze Xie, Yue Wu, Zhongdao Wang, James Kwok, Ping Luo, Huchuan Lu, et~al.
\newblock Pixart-alpha: Fast training of diffusion transformer for photorealistic text-to-image synthesis.
\newblock \emph{arXiv preprint arXiv:2310.00426}, 2023.

\bibitem[Chen et~al.(2024{\natexlab{a}})Chen, Ge, Xie, Wu, Yao, Ren, Wang, Luo, Lu, and Li]{chen2024pixart}
Junsong Chen, Chongjian Ge, Enze Xie, Yue Wu, Lewei Yao, Xiaozhe Ren, Zhongdao Wang, Ping Luo, Huchuan Lu, and Zhenguo Li.
\newblock Pixart-$\sigma$: Weak-to-strong training of diffusion transformer for 4k text-to-image generation.
\newblock In \emph{{Proceedings of the European Conference on Computer Vision ({ECCV})}}, pages 74--91. Springer, 2024{\natexlab{a}}.

\bibitem[Chen et~al.(2024{\natexlab{b}})Chen, Xu, Ren, Cong, He, Xie, Sinha, Luo, Xiang, and Perez-Rua]{chen2024gentron}
Shoufa Chen, Mengmeng Xu, Jiawei Ren, Yuren Cong, Sen He, Yanping Xie, Animesh Sinha, Ping Luo, Tao Xiang, and Juan-Manuel Perez-Rua.
\newblock Gentron: Diffusion transformers for image and video generation.
\newblock In \emph{{Proceedings of the {IEEE} Conference on Computer Vision and Pattern Recognition ({CVPR})}}, 2024{\natexlab{b}}.

\bibitem[Chen et~al.(2025{\natexlab{c}})Chen, Wu, Liu, Pan, Liu, Xie, Yu, and Ruan]{chen2025janus}
Xiaokang Chen, Zhiyu Wu, Xingchao Liu, Zizheng Pan, Wen Liu, Zhenda Xie, Xingkai Yu, and Chong Ruan.
\newblock Janus-pro: Unified multimodal understanding and generation with data and model scaling.
\newblock \emph{arXiv preprint arXiv:2501.17811}, 2025{\natexlab{c}}.

\bibitem[Chen et~al.(2024{\natexlab{c}})Chen, Wu, Wang, Su, Chen, Xing, Zhong, Zhang, Zhu, Lu, et~al.]{chen2024internvl}
Zhe Chen, Jiannan Wu, Wenhai Wang, Weijie Su, Guo Chen, Sen Xing, Muyan Zhong, Qinglong Zhang, Xizhou Zhu, Lewei Lu, et~al.
\newblock Internvl: Scaling up vision foundation models and aligning for generic visual-linguistic tasks.
\newblock In \emph{{Proceedings of the {IEEE} Conference on Computer Vision and Pattern Recognition ({CVPR})}}, pages 24185--24198, 2024{\natexlab{c}}.

\bibitem[Cho et~al.(2025)Cho, Madotto, Mavroudi, Afouras, Nagarajan, Maaz, Song, Ma, Hu, Jain, et~al.]{cho2025perceptionlm}
Jang~Hyun Cho, Andrea Madotto, Effrosyni Mavroudi, Triantafyllos Afouras, Tushar Nagarajan, Muhammad Maaz, Yale Song, Tengyu Ma, Shuming Hu, Suyog Jain, et~al.
\newblock Perceptionlm: Open-access data and models for detailed visual understanding.
\newblock \emph{arXiv preprint arXiv:2504.13180}, 2025.

\bibitem[Clark et~al.(2019)Clark, Khandelwal, Levy, and Manning]{clark2019does}
Kevin Clark, Urvashi Khandelwal, Omer Levy, and Christopher~D Manning.
\newblock What does bert look at? an analysis of bert's attention.
\newblock \emph{arXiv preprint arXiv:1906.04341}, 2019.

\bibitem[Csord{\'a}s et~al.(2024)Csord{\'a}s, Irie, Schmidhuber, Potts, and Manning]{csordas2024moeut}
R{\'o}bert Csord{\'a}s, Kazuki Irie, J{\"u}rgen Schmidhuber, Christopher Potts, and Christopher~D Manning.
\newblock Moeut: Mixture-of-experts universal transformers.
\newblock \emph{{Advances in Neural Information Processing Systems ({NeurIPS})}}, 37:\penalty0 28589--28614, 2024.

\bibitem[Dai et~al.(2023)Dai, Hou, Ma, Tsai, Wang, Wang, Zhang, Vandenhende, Wang, Dubey, et~al.]{dai2023emu}
Xiaoliang Dai, Ji~Hou, Chih-Yao Ma, Sam Tsai, Jialiang Wang, Rui Wang, Peizhao Zhang, Simon Vandenhende, Xiaofang Wang, Abhimanyu Dubey, et~al.
\newblock Emu: Enhancing image generation models using photogenic needles in a haystack.
\newblock \emph{arXiv preprint arXiv:2309.15807}, 2023.

\bibitem[Darcet et~al.(2023)Darcet, Oquab, Mairal, and Bojanowski]{darcet2023vision}
Timoth{\'e}e Darcet, Maxime Oquab, Julien Mairal, and Piotr Bojanowski.
\newblock Vision transformers need registers.
\newblock \emph{arXiv preprint arXiv:2309.16588}, 2023.

\bibitem[DeepMind(2025)]{google2025gemini2}
Google DeepMind.
\newblock {Gemini 2.0}.
\newblock \url{https://gemini.google.com/}, 2025.

\bibitem[Deng et~al.(2025)Deng, Zhu, Li, Gou, Li, Wang, Zhong, Yu, Nie, Song, et~al.]{deng2025emerging}
Chaorui Deng, Deyao Zhu, Kunchang Li, Chenhui Gou, Feng Li, Zeyu Wang, Shu Zhong, Weihao Yu, Xiaonan Nie, Ziang Song, et~al.
\newblock Emerging properties in unified multimodal pretraining.
\newblock \emph{arXiv preprint arXiv:2505.14683}, 2025.

\bibitem[Devlin et~al.(2019)Devlin, Chang, Lee, and Toutanova]{devlin2019bert}
Jacob Devlin, Ming-Wei Chang, Kenton Lee, and Kristina Toutanova.
\newblock Bert: Pre-training of deep bidirectional transformers for language understanding.
\newblock In \emph{Conference of the North American Chapter of the Association for Computational Linguistics: Human Language Technologies (Long and Short Papers)}, 2019.

\bibitem[Dhariwal and Nichol(2021)]{dhariwal2021diffusion}
Prafulla Dhariwal and Alexander Nichol.
\newblock Diffusion models beat {GANs} on image synthesis.
\newblock In \emph{{Advances in Neural Information Processing Systems ({NeurIPS})}}, 2021.

\bibitem[Dosovitskiy et~al.(2020)Dosovitskiy, Beyer, Kolesnikov, Weissenborn, Zhai, Unterthiner, Dehghani, Minderer, Heigold, Gelly, et~al.]{dosovitskiy2020vit}
Alexey Dosovitskiy, Lucas Beyer, Alexander Kolesnikov, Dirk Weissenborn, Xiaohua Zhai, Thomas Unterthiner, Mostafa Dehghani, Matthias Minderer, Georg Heigold, Sylvain Gelly, et~al.
\newblock An image is worth 16x16 words: Transformers for image recognition at scale.
\newblock In \emph{{Proceedings of the International Conference on Learning Representations ({ICLR})}}, 2020.

\bibitem[Eigen et~al.(2013)Eigen, Ranzato, and Sutskever]{eigen2013learning}
David Eigen, Marc'Aurelio Ranzato, and Ilya Sutskever.
\newblock Learning factored representations in a deep mixture of experts.
\newblock \emph{arXiv preprint arXiv:1312.4314}, 2013.

\bibitem[Esser et~al.(2024{\natexlab{a}})Esser, Kulal, Blattmann, Entezari, M{\"u}ller, Saini, Levi, Lorenz, Sauer, Boesel, et~al.]{esser2024scaling}
Patrick Esser, Sumith Kulal, Andreas Blattmann, Rahim Entezari, Jonas M{\"u}ller, Harry Saini, Yam Levi, Dominik Lorenz, Axel Sauer, Frederic Boesel, et~al.
\newblock Scaling rectified flow transformers for high-resolution image synthesis.
\newblock In \emph{{Proceedings of the International Conference on Machine Learning ({ICML})}}, 2024{\natexlab{a}}.

\bibitem[Esser et~al.(2024{\natexlab{b}})Esser, Kulal, Blattmann, Entezari, M{\"u}ller, Saini, Levi, Lorenz, Sauer, Boesel, et~al.]{esser2024sd3}
Patrick Esser, Sumith Kulal, Andreas Blattmann, Rahim Entezari, Jonas M{\"u}ller, Harry Saini, Yam Levi, Dominik Lorenz, Axel Sauer, Frederic Boesel, et~al.
\newblock Scaling rectified flow transformers for high-resolution image synthesis.
\newblock In \emph{{Proceedings of the International Conference on Machine Learning ({ICML})}}, 2024{\natexlab{b}}.

\bibitem[Face(2022)]{huggingface2022diffusers}
Hugging Face.
\newblock Diffusers: State-of-the-art diffusion models.
\newblock \url{https://github.com/huggingface/diffusers}, 2022.

\bibitem[Fedus et~al.(2022)Fedus, Zoph, and Shazeer]{fedus2022switch}
William Fedus, Barret Zoph, and Noam Shazeer.
\newblock Switch transformers: Scaling to trillion parameter models with simple and efficient sparsity.
\newblock \emph{{Journal of Machine Learning Research ({JMLR})}}, 23\penalty0 (120):\penalty0 1--39, 2022.

\bibitem[Gao et~al.(2025)Gao, Gong, Guo, Hou, Lai, Li, Li, Lian, Liao, Liu, et~al.]{gao2025seedream}
Yu~Gao, Lixue Gong, Qiushan Guo, Xiaoxia Hou, Zhichao Lai, Fanshi Li, Liang Li, Xiaochen Lian, Chao Liao, Liyang Liu, et~al.
\newblock Seedream 3.0 technical report.
\newblock \emph{arXiv preprint arXiv:2504.11346}, 2025.

\bibitem[Ge(2025)]{JiaxinGe2025DiffusersBAGEL}
Jiaxin Ge.
\newblock Diffusers-bagel: Bagel custom pipeline for diffusers (hugging face).
\newblock \url{https://huggingface.co/JiaxinGe/Diffusers-BAGEL}, 2025.
\newblock arXiv preprint arXiv:2505.14683.

\bibitem[Ge et~al.(2024)Ge, Zhao, Zhu, Ge, Yi, Song, Li, Ding, and Shan]{ge2024seed}
Yuying Ge, Sijie Zhao, Jinguo Zhu, Yixiao Ge, Kun Yi, Lin Song, Chen Li, Xiaohan Ding, and Ying Shan.
\newblock Seed-x: Multimodal models with unified multi-granularity comprehension and generation.
\newblock \emph{arXiv preprint arXiv:2404.14396}, 2024.

\bibitem[Geng et~al.(2025)Geng, Wang, Ma, Li, Rao, Gu, Zhong, Lu, Hu, Zhang, et~al.]{geng2025x}
Zigang Geng, Yibing Wang, Yeyao Ma, Chen Li, Yongming Rao, Shuyang Gu, Zhao Zhong, Qinglin Lu, Han Hu, Xiaosong Zhang, et~al.
\newblock X-omni: Reinforcement learning makes discrete autoregressive image generative models great again.
\newblock \emph{arXiv preprint arXiv:2507.22058}, 2025.

\bibitem[Ghosh et~al.(2023)Ghosh, Hajishirzi, and Schmidt]{ghosh2023geneval}
Dhruba Ghosh, Hannaneh Hajishirzi, and Ludwig Schmidt.
\newblock Geneval: An object-focused framework for evaluating text-to-image alignment.
\newblock \emph{{Advances in Neural Information Processing Systems ({NeurIPS})}}, 36:\penalty0 52132--52152, 2023.

\bibitem[Google(2025)]{google2025imagen}
Google.
\newblock Imagen.
\newblock \url{https://deepmind.google/models/imagen/}, 2025.

\bibitem[Guo et~al.(2025)Guo, Zhang, Tong, Zhao, Huang, Zhang, Zhang, Liu, Zhang, Gao, et~al.]{guo2025can}
Ziyu Guo, Renrui Zhang, Chengzhuo Tong, Zhizheng Zhao, Rui Huang, Haoquan Zhang, Manyuan Zhang, Jiaming Liu, Shanghang Zhang, Peng Gao, et~al.
\newblock Can we generate images with cot? let's verify and reinforce image generation step by step.
\newblock \emph{arXiv preprint arXiv:2501.13926}, 2025.

\bibitem[Hampshire and Waibel(1989)]{hampshire1989connectionist}
John Hampshire and Alex Waibel.
\newblock Connectionist architectures for multi-speaker phoneme recognition.
\newblock \emph{Advances in neural information processing systems}, 2, 1989.

\bibitem[Han et~al.(2025)Han, Chen, Zhao, Wang, Zhao, Yang, He, Yue, and Jiang]{han2025vision}
Jiaming Han, Hao Chen, Yang Zhao, Hanyu Wang, Qi~Zhao, Ziyan Yang, Hao He, Xiangyu Yue, and Lu~Jiang.
\newblock Vision as a dialect: Unifying visual understanding and generation via text-aligned representations.
\newblock \emph{arXiv preprint arXiv:2506.18898}, 2025.

\bibitem[Han et~al.(2021)Han, Huang, Song, Yang, Wang, and Wang]{han2021dynamic}
Yizeng Han, Gao Huang, Shiji Song, Le~Yang, Honghui Wang, and Yulin Wang.
\newblock Dynamic neural networks: A survey.
\newblock \emph{{{IEEE} Transactions on Pattern Analysis and Machine Intelligence ({PAMI})}}, 44\penalty0 (11):\penalty0 7436--7456, 2021.

\bibitem[Henry et~al.(2020)Henry, Dachapally, Pawar, and Chen]{henry2020query}
Alex Henry, Prudhvi~Raj Dachapally, Shubham Pawar, and Yuxuan Chen.
\newblock Query-key normalization for transformers.
\newblock \emph{arXiv preprint arXiv:2010.04245}, 2020.

\bibitem[Heusel et~al.(2017)Heusel, Ramsauer, Unterthiner, Nessler, and Hochreiter]{heusel2017gans}
Martin Heusel, Hubert Ramsauer, Thomas Unterthiner, Bernhard Nessler, and Sepp Hochreiter.
\newblock Gans trained by a two time-scale update rule converge to a local nash equilibrium.
\newblock \emph{{Advances in Neural Information Processing Systems ({NeurIPS})}}, 30, 2017.

\bibitem[Ho et~al.(2020)Ho, Jain, and Abbeel]{ho2020denoising}
Jonathan Ho, Ajay Jain, and Pieter Abbeel.
\newblock Denoising diffusion probabilistic models.
\newblock In \emph{{Advances in Neural Information Processing Systems ({NeurIPS})}}, 2020.

\bibitem[Hsu et~al.(2025)Hsu, Dai, Kothapalli, Song, Tang, Zhu, Shimizu, Sahni, Ning, Chen, and Wang]{hsu2025ligerkernel}
Pin-Lun Hsu, Yun Dai, Vignesh Kothapalli, Qingquan Song, Shao Tang, Siyu Zhu, Steven Shimizu, Shivam Sahni, Haowen Ning, Yanning Chen, and Zhipeng Wang.
\newblock Liger-kernel: Efficient triton kernels for {LLM} training.
\newblock In \emph{Championing Open-source DEvelopment in ML Workshop @ ICML25}, 2025.
\newblock \url{https://openreview.net/forum?id=36SjAIT42G}.

\bibitem[Hu et~al.(2024)Hu, Wang, Fang, Fu, Cheng, and Yu]{hu2024ella}
Xiwei Hu, Rui Wang, Yixiao Fang, Bin Fu, Pei Cheng, and Gang Yu.
\newblock Ella: Equip diffusion models with llm for enhanced semantic alignment.
\newblock \emph{arXiv preprint arXiv:2403.05135}, 2024.

\bibitem[Hurst et~al.(2024)Hurst, Lerer, Goucher, Perelman, Ramesh, Clark, Ostrow, Welihinda, Hayes, Radford, et~al.]{hurst2024gpt}
Aaron Hurst, Adam Lerer, Adam~P Goucher, Adam Perelman, Aditya Ramesh, Aidan Clark, AJ~Ostrow, Akila Welihinda, Alan Hayes, Alec Radford, et~al.
\newblock Gpt-4o system card.
\newblock \emph{arXiv preprint arXiv:2410.21276}, 2024.

\bibitem[Ivakhnenko and Lapa(1966)]{ivakhnenko1966cybernetic}
Alekse{\u{i}}~Grigor'evich Ivakhnenko and Valentin~Grigor'evich Lapa.
\newblock Cybernetic predicting devices.
\newblock \emph{Technical Report}, 1966.

\bibitem[Ivakhnenko(2007)]{ivakhnenko2007polynomial}
Alexey~Grigorevich Ivakhnenko.
\newblock Polynomial theory of complex systems.
\newblock \emph{IEEE transactions on Systems, Man, and Cybernetics}, \penalty0 (4):\penalty0 364--378, 2007.

\bibitem[Jacobs et~al.(1991)Jacobs, Jordan, Nowlan, and Hinton]{jacobs1991adaptive}
Robert~A Jacobs, Michael~I Jordan, Steven~J Nowlan, and Geoffrey~E Hinton.
\newblock Adaptive mixtures of local experts.
\newblock \emph{Neural computation}, 3\penalty0 (1):\penalty0 79--87, 1991.

\bibitem[Jarzynski(1997)]{jarzynski1997equilibrium}
Christopher Jarzynski.
\newblock Equilibrium free-energy differences from nonequilibrium measurements: A master-equation approach.
\newblock \emph{Physical Review E}, 1997.

\bibitem[Jiang et~al.(2024)Jiang, Sablayrolles, Roux, Mensch, Savary, Bamford, Chaplot, Casas, Hanna, Bressand, et~al.]{jiang2024mixtral}
Albert~Q Jiang, Alexandre Sablayrolles, Antoine Roux, Arthur Mensch, Blanche Savary, Chris Bamford, Devendra~Singh Chaplot, Diego de~las Casas, Emma~Bou Hanna, Florian Bressand, et~al.
\newblock Mixtral of experts.
\newblock \emph{arXiv preprint arXiv:2401.04088}, 2024.

\bibitem[Kahatapitiya et~al.(2025)Kahatapitiya, Liu, He, Liu, Jia, Zhang, Ryoo, and Xie]{kahatapitiya2024adaptive}
Kumara Kahatapitiya, Haozhe Liu, Sen He, Ding Liu, Menglin Jia, Chenyang Zhang, Michael~S Ryoo, and Tian Xie.
\newblock Adaptive caching for faster video generation with diffusion transformers.
\newblock \emph{{Proceedings of the International Conference on Computer Vision ({ICCV})}}, 2025.

\bibitem[Labs(2024)]{flux2024}
Black~Forest Labs.
\newblock Flux.
\newblock \url{https://github.com/black-forest-labs/flux}, 2024.

\bibitem[Lepikhin et~al.(2020)Lepikhin, Lee, Xu, Chen, Firat, Huang, Krikun, Shazeer, and Chen]{lepikhin2020gshard}
Dmitry Lepikhin, HyoukJoong Lee, Yuanzhong Xu, Dehao Chen, Orhan Firat, Yanping Huang, Maxim Krikun, Noam Shazeer, and Zhifeng Chen.
\newblock Gshard: Scaling giant models with conditional computation and automatic sharding.
\newblock \emph{arXiv preprint arXiv:2006.16668}, 2020.

\bibitem[Li et~al.(2024{\natexlab{a}})Li, Kamko, Akhgari, Sabet, Xu, and Doshi]{li2024playground}
Daiqing Li, Aleks Kamko, Ehsan Akhgari, Ali Sabet, Linmiao Xu, and Suhail Doshi.
\newblock Playground v2.5: Three insights towards enhancing aesthetic quality in text-to-image generation, 2024{\natexlab{a}}.

\bibitem[Li et~al.(2024{\natexlab{b}})Li, Zhang, Lin, Xiong, Long, Deng, Zhang, Liu, Huang, Xiao, et~al.]{li2024hunyuan}
Zhimin Li, Jianwei Zhang, Qin Lin, Jiangfeng Xiong, Yanxin Long, Xinchi Deng, Yingfang Zhang, Xingchao Liu, Minbin Huang, Zedong Xiao, et~al.
\newblock Hunyuan-dit: A powerful multi-resolution diffusion transformer with fine-grained chinese understanding.
\newblock \emph{arXiv preprint arXiv:2405.08748}, 2024{\natexlab{b}}.

\bibitem[Liang et~al.(2025)Liang, YU, Luo, Iyer, Dong, Zhou, Ghosh, Lewis, tau Yih, Zettlemoyer, and Lin]{liang2025mixtureoftransformers}
Weixin Liang, LILI YU, Liang Luo, Srini Iyer, Ning Dong, Chunting Zhou, Gargi Ghosh, Mike Lewis, Wen tau Yih, Luke Zettlemoyer, and Xi~Victoria Lin.
\newblock Mixture-of-transformers: A sparse and scalable architecture for multi-modal foundation models.
\newblock \emph{Transactions on Machine Laerning Research ({TMLR})}, 2025.
\newblock ISSN 2835-8856.
\newblock \url{https://openreview.net/forum?id=Nu6N69i8SB}.

\bibitem[Liao et~al.(2025)Liao, Liu, Wang, Luo, Zhang, Zhao, Wu, Li, Tian, and Huang]{liao2025mogao}
Chao Liao, Liyang Liu, Xun Wang, Zhengxiong Luo, Xinyu Zhang, Wenliang Zhao, Jie Wu, Liang Li, Zhi Tian, and Weilin Huang.
\newblock Mogao: An omni foundation model for interleaved multi-modal generation.
\newblock \emph{arXiv preprint arXiv:2505.05472}, 2025.

\bibitem[Lin et~al.(2025)Lin, Li, Cheng, Niu, Ye, He, Yuan, Yu, Wang, Ge, et~al.]{lin2025uniworld}
Bin Lin, Zongjian Li, Xinhua Cheng, Yuwei Niu, Yang Ye, Xianyi He, Shenghai Yuan, Wangbo Yu, Shaodong Wang, Yunyang Ge, et~al.
\newblock Uniworld: High-resolution semantic encoders for unified visual understanding and generation.
\newblock \emph{arXiv preprint arXiv:2506.03147}, 2025.

\bibitem[Lin et~al.(2024)Lin, Liu, Li, and Yang]{lin2024common}
Shanchuan Lin, Bingchen Liu, Jiashi Li, and Xiao Yang.
\newblock Common diffusion noise schedules and sample steps are flawed.
\newblock In \emph{{{IEEE} Winter Conference on Applications of Computer Vision (WACV)}}, pages 5404--5411, 2024.

\bibitem[Lipman et~al.(2023)Lipman, Chen, Ben-Hamu, Nickel, and Le]{lipman2023flow}
Yaron Lipman, Ricky~TQ Chen, Heli Ben-Hamu, Maximilian Nickel, and Matt Le.
\newblock Flow matching for generative modeling.
\newblock In \emph{{Proceedings of the International Conference on Learning Representations ({ICLR})}}, 2023.

\bibitem[Liu et~al.(2024{\natexlab{a}})Liu, Akhgari, Visheratin, Kamko, Xu, Shrirao, Lambert, Souza, Doshi, and Li]{liu2024playground}
Bingchen Liu, Ehsan Akhgari, Alexander Visheratin, Aleks Kamko, Linmiao Xu, Shivam Shrirao, Chase Lambert, Joao Souza, Suhail Doshi, and Daiqing Li.
\newblock Playground v3: Improving text-to-image alignment with deep-fusion large language models.
\newblock \emph{arXiv preprint arXiv:2409.10695}, 2024{\natexlab{a}}.

\bibitem[Liu et~al.(2025{\natexlab{a}})Liu, Liu, Zhou, Xu, Xie, Han, Perez, Liu, Kahatapitiya, Jia, et~al.]{liu2025mardini}
Haozhe Liu, Shikun Liu, Zijian Zhou, Mengmeng Xu, Yanping Xie, Xiao Han, Juan~Camilo Perez, Ding Liu, Kumara Kahatapitiya, Menglin Jia, et~al.
\newblock Mardini: Masked auto-regressive diffusion for video generation at scale.
\newblock \emph{Transactions on Machine Laerning Research ({TMLR})}, 2025{\natexlab{a}}.

\bibitem[Liu et~al.(2025{\natexlab{b}})Liu, Zhang, Xie, Faccio, Xu, Xiang, Shou, Perez-Rua, and Schmidhuber]{liu2025faster}
Haozhe Liu, Wentian Zhang, Jinheng Xie, Francesco Faccio, Mengmeng Xu, Tao Xiang, Mike~Zheng Shou, Juan-Manuel Perez-Rua, and J{\"u}rgen Schmidhuber.
\newblock Faster diffusion through temporal attention decomposition.
\newblock \emph{Transactions on Machine Laerning Research ({TMLR})}, 2025{\natexlab{b}}.

\bibitem[Liu et~al.(2025{\natexlab{c}})Liu, Liu, Liang, Li, Liu, Wang, Wan, Zhang, and Ouyang]{liu2025flow}
Jie Liu, Gongye Liu, Jiajun Liang, Yangguang Li, Jiaheng Liu, Xintao Wang, Pengfei Wan, Di~Zhang, and Wanli Ouyang.
\newblock Flow-grpo: Training flow matching models via online rl.
\newblock \emph{arXiv preprint arXiv:2505.05470}, 2025{\natexlab{c}}.

\bibitem[Liu et~al.(2023)Liu, Fan, Johns, Yu, Xiao, and Anandkumar]{liu2023prismer}
Shikun Liu, Linxi Fan, Edward Johns, Zhiding Yu, Chaowei Xiao, and Anima Anandkumar.
\newblock Prismer: A vision-language model with an ensemble of experts.
\newblock \emph{arXiv preprint arXiv:2303.02506}, 2023.

\bibitem[Liu et~al.(2025{\natexlab{d}})Liu, Han, Xing, Yin, Wang, Cheng, Liao, Wang, Fu, Han, Li, Peng, Sun, Wu, Cai, Ge, Ming, Xia, Zeng, Zhu, Jiao, Zhang, Yu, and Jiang]{liu2025step1x_edit}
Shiyu Liu, Yucheng Han, Peng Xing, Fukun Yin, Rui Wang, Wei Cheng, Jiaqi Liao, Yingming Wang, Honghao Fu, Chunrui Han, Guopeng Li, Yuang Peng, Quan Sun, Jingwei Wu, Yan Cai, Zheng Ge, Ranchen Ming, Lei Xia, Xianfang Zeng, Yibo Zhu, Binxing Jiao, Xiangyu Zhang, Gang Yu, and Daxin Jiang.
\newblock Step1x-edit: A practical framework for general image editing.
\newblock \emph{arXiv preprint arXiv:2504.17761}, 2025{\natexlab{d}}.

\bibitem[Liu et~al.(2024{\natexlab{b}})Liu, Liang, Liang, Luo, Li, Huang, and Yuan]{liu2024glyph}
Zeyu Liu, Weicong Liang, Zhanhao Liang, Chong Luo, Ji~Li, Gao Huang, and Yuhui Yuan.
\newblock Glyph-byt5: A customized text encoder for accurate visual text rendering.
\newblock In \emph{European Conference on Computer Vision}, pages 361--377. Springer, 2024{\natexlab{b}}.

\bibitem[Neal(2001)]{neal2001annealed}
Radford~M Neal.
\newblock Annealed importance sampling.
\newblock \emph{Statistics and computing}, 2001.

\bibitem[Niu et~al.(2025)Niu, Ning, Zheng, Jin, Lin, Jin, Liao, Ning, Feng, Zhu, and Yuan]{niu2025wise}
Yuwei Niu, Munan Ning, Mengren Zheng, Weiyang Jin, Bin Lin, Peng Jin, Jiaqi Liao, Kunpeng Ning, Chaoran Feng, Bin Zhu, and Li~Yuan.
\newblock Wise: A world knowledge-informed semantic evaluation for text-to-image generation.
\newblock \emph{arXiv preprint arXiv:2503.07265}, 2025.

\bibitem[OpenAI(2025{\natexlab{a}})]{openai_dalle3_2025}
OpenAI.
\newblock Dall·e 3.
\newblock \url{https://openai.com/research/dall-e-3}, 2025{\natexlab{a}}.
\newblock Accessed: 2025-09-16.

\bibitem[OpenAI(2025{\natexlab{b}})]{openai_gpt_image_1_2025}
OpenAI.
\newblock gpt-image-1.
\newblock \url{https://openai.com/index/image-generation-api/}, April 2025{\natexlab{b}}.
\newblock OpenAI blog post introducing the model. Accessed: 2025-09-15.

\bibitem[Pan et~al.(2025)Pan, Shukla, Singh, Zhao, Mishra, Wang, Xu, Chen, Li, Juefei-Xu, et~al.]{pan2025transfer}
Xichen Pan, Satya~Narayan Shukla, Aashu Singh, Zhuokai Zhao, Shlok~Kumar Mishra, Jialiang Wang, Zhiyang Xu, Jiuhai Chen, Kunpeng Li, Felix Juefei-Xu, et~al.
\newblock Transfer between modalities with metaqueries.
\newblock \emph{arXiv preprint arXiv:2504.06256}, 2025.

\bibitem[Peebles and Xie(2023)]{peebles2023scalable}
William Peebles and Saining Xie.
\newblock Scalable diffusion models with transformers.
\newblock In \emph{{Proceedings of the International Conference on Computer Vision ({ICCV})}}, pages 4195--4205, 2023.

\bibitem[Podell et~al.(2023)Podell, English, Lacey, Blattmann, Dockhorn, M{\"u}ller, Penna, and Rombach]{podell2023sdxl}
Dustin Podell, Zion English, Kyle Lacey, Andreas Blattmann, Tim Dockhorn, Jonas M{\"u}ller, Joe Penna, and Robin Rombach.
\newblock Sdxl: Improving latent diffusion models for high-resolution image synthesis.
\newblock \emph{arXiv preprint arXiv:2307.01952}, 2023.

\bibitem[Polyak et~al.(2024{\natexlab{a}})Polyak, Zohar, Brown, Tjandra, Sinha, Lee, Vyas, Shi, Ma, Chuang, et~al.]{polyak2024movie}
Adam Polyak, Amit Zohar, Andrew Brown, Andros Tjandra, Animesh Sinha, Ann Lee, Apoorv Vyas, Bowen Shi, Chih-Yao Ma, Ching-Yao Chuang, et~al.
\newblock Movie gen: A cast of media foundation models.
\newblock \emph{arXiv preprint arXiv:2410.13720}, 2024{\natexlab{a}}.

\bibitem[Polyak et~al.(2024{\natexlab{b}})Polyak, Zohar, Brown, Tjandra, Sinha, Lee, Vyas, Shi, Ma, Chuang, et~al.]{polyak2024moviegen}
Adam Polyak, Amit Zohar, Andrew Brown, Andros Tjandra, Animesh Sinha, Ann Lee, Apoorv Vyas, Bowen Shi, Chih-Yao Ma, Ching-Yao Chuang, et~al.
\newblock Movie gen: A cast of media foundation models.
\newblock \emph{arXiv preprint arXiv:2410.13720}, 2024{\natexlab{b}}.

\bibitem[Qin et~al.(2025)Qin, Zhuo, Xin, Du, Li, Fu, Lu, Yuan, Li, Liu, et~al.]{qin2025lumina}
Qi~Qin, Le~Zhuo, Yi~Xin, Ruoyi Du, Zhen Li, Bin Fu, Yiting Lu, Jiakang Yuan, Xinyue Li, Dongyang Liu, et~al.
\newblock Lumina-image 2.0: A unified and efficient image generative framework.
\newblock \emph{arXiv preprint arXiv:2503.21758}, 2025.

\bibitem[Radford et~al.(2021)Radford, Kim, Hallacy, Ramesh, Goh, Agarwal, Sastry, Askell, Mishkin, Clark, et~al.]{radford2021learning}
Alec Radford, Jong~Wook Kim, Chris Hallacy, Aditya Ramesh, Gabriel Goh, Sandhini Agarwal, Girish Sastry, Amanda Askell, Pamela Mishkin, Jack Clark, et~al.
\newblock Learning transferable visual models from natural language supervision.
\newblock In \emph{{Proceedings of the International Conference on Machine Learning ({ICML})}}, pages 8748--8763. PmLR, 2021.

\bibitem[Raffel et~al.(2020)Raffel, Shazeer, Roberts, Lee, Narang, Matena, Zhou, Li, Liu, et~al.]{raffel2020t5}
Colin Raffel, Noam Shazeer, Adam Roberts, Katherine Lee, Sharan Narang, Michael Matena, Yanqi Zhou, Wei Li, Peter~J Liu, et~al.
\newblock Exploring the limits of transfer learning with a unified text-to-text transformer.
\newblock \emph{{Journal of Machine Learning Research ({JMLR})}}, 2020.

\bibitem[Rai et~al.(2024)Rai, Zhou, Feng, Saparov, and Yao]{rai2024practical}
Daking Rai, Yilun Zhou, Shi Feng, Abulhair Saparov, and Ziyu Yao.
\newblock A practical review of mechanistic interpretability for transformer-based language models.
\newblock \emph{arXiv preprint arXiv:2407.02646}, 2024.

\bibitem[Ramesh et~al.(2022)Ramesh, Dhariwal, Nichol, Chu, and Chen]{ramesh2022hierarchical}
Aditya Ramesh, Prafulla Dhariwal, Alex Nichol, Casey Chu, and Mark Chen.
\newblock Hierarchical text-conditional image generation with clip latents.
\newblock \emph{arXiv preprint arXiv:2204.06125}, 2022.

\bibitem[Raposo et~al.(2024)Raposo, Ritter, Richards, Lillicrap, Humphreys, and Santoro]{raposo2024mixture}
David Raposo, Sam Ritter, Blake Richards, Timothy Lillicrap, Peter~Conway Humphreys, and Adam Santoro.
\newblock Mixture-of-depths: Dynamically allocating compute in transformer-based language models.
\newblock \emph{arXiv preprint arXiv:2404.02258}, 2024.

\bibitem[Rombach et~al.(2022)Rombach, Blattmann, Lorenz, Esser, and Ommer]{rombach2022high}
Robin Rombach, Andreas Blattmann, Dominik Lorenz, Patrick Esser, and Bj{\"o}rn Ommer.
\newblock High-resolution image synthesis with latent diffusion models.
\newblock In \emph{{Proceedings of the {IEEE} Conference on Computer Vision and Pattern Recognition ({CVPR})}}, 2022.

\bibitem[Saharia et~al.(2022)Saharia, Chan, Saxena, Li, Whang, Denton, Ghasemipour, Gontijo~Lopes, Karagol~Ayan, Salimans, et~al.]{saharia2022photorealistic}
Chitwan Saharia, William Chan, Saurabh Saxena, Lala Li, Jay Whang, Emily~L Denton, Kamyar Ghasemipour, Raphael Gontijo~Lopes, Burcu Karagol~Ayan, Tim Salimans, et~al.
\newblock Photorealistic text-to-image diffusion models with deep language understanding.
\newblock In \emph{{Advances in Neural Information Processing Systems ({NeurIPS})}}, 2022.

\bibitem[Schmidhuber(1992)]{schmidhuber1992learning}
J{\"u}rgen Schmidhuber.
\newblock Learning to control fast-weight memories: An alternative to dynamic recurrent networks.
\newblock \emph{Neural Computation}, 4\penalty0 (1):\penalty0 131--139, 1992.

\bibitem[Shao et~al.(2024)Shao, Wang, Zhu, Xu, Song, Bi, Zhang, Zhang, Li, Wu, et~al.]{shao2024deepseekmath}
Zhihong Shao, Peiyi Wang, Qihao Zhu, Runxin Xu, Junxiao Song, Xiao Bi, Haowei Zhang, Mingchuan Zhang, YK~Li, Yang Wu, et~al.
\newblock Deepseekmath: Pushing the limits of mathematical reasoning in open language models.
\newblock \emph{arXiv preprint arXiv:2402.03300}, 2024.

\bibitem[Shazeer et~al.(2017)Shazeer, Mirhoseini, Maziarz, Davis, Le, Hinton, and Dean]{shazeer2017outrageously}
Noam Shazeer, Azalia Mirhoseini, Krzysztof Maziarz, Andy Davis, Quoc Le, Geoffrey Hinton, and Jeff Dean.
\newblock Outrageously large neural networks: The sparsely-gated mixture-of-experts layer.
\newblock \emph{arXiv preprint arXiv:1701.06538}, 2017.

\bibitem[Shi et~al.(2024)Shi, Han, Zhou, Liang, Lin, Zettlemoyer, and Yu]{shi2024lmfusion}
Weijia Shi, Xiaochuang Han, Chunting Zhou, Weixin Liang, Xi~Victoria Lin, Luke Zettlemoyer, and Lili Yu.
\newblock Lmfusion: Adapting pretrained language models for multimodal generation.
\newblock \emph{arXiv preprint arXiv:2412.15188}, 2024.

\bibitem[Sun et~al.(2025)Sun, Jiang, Zhao, and He]{sun2025noise}
Qiao Sun, Zhicheng Jiang, Hanhong Zhao, and Kaiming He.
\newblock Is noise conditioning necessary for denoising generative models?
\newblock In \emph{{Proceedings of the International Conference on Machine Learning ({ICML})}}, 2025.

\bibitem[Tang et~al.(2025)Tang, Zheng, Paul, and Xie]{tang2025exploring}
Bingda Tang, Boyang Zheng, Sayak Paul, and Saining Xie.
\newblock Exploring the deep fusion of large language models and diffusion transformers for text-to-image synthesis.
\newblock In \emph{{Proceedings of the {IEEE} Conference on Computer Vision and Pattern Recognition ({CVPR})}}, 2025.

\bibitem[Tay et~al.(2022)Tay, Dehghani, Tran, Garcia, Wei, Wang, Chung, Shakeri, Bahri, Schuster, et~al.]{tay2022ul2}
Yi~Tay, Mostafa Dehghani, Vinh~Q Tran, Xavier Garcia, Jason Wei, Xuezhi Wang, Hyung~Won Chung, Siamak Shakeri, Dara Bahri, Tal Schuster, et~al.
\newblock Ul2: Unifying language learning paradigms.
\newblock \emph{arXiv preprint arXiv:2205.05131}, 2022.

\bibitem[Team(2024{\natexlab{a}})]{team2024chameleon}
Chameleon Team.
\newblock Chameleon: Mixed-modal early-fusion foundation models.
\newblock \emph{arXiv preprint arXiv:2405.09818}, 2024{\natexlab{a}}.

\bibitem[Team et~al.(2024)Team, Riviere, Pathak, Sessa, Hardin, Bhupatiraju, Hussenot, Mesnard, Shahriari, Ram{\'e}, et~al.]{team2024gemma}
Gemma Team, Morgane Riviere, Shreya Pathak, Pier~Giuseppe Sessa, Cassidy Hardin, Surya Bhupatiraju, L{\'e}onard Hussenot, Thomas Mesnard, Bobak Shahriari, Alexandre Ram{\'e}, et~al.
\newblock Gemma 2: Improving open language models at a practical size.
\newblock \emph{arXiv preprint arXiv:2408.00118}, 2024.

\bibitem[Team(2025)]{kuai2025kolors2}
Kuaishou~Kolors Team.
\newblock {Kolors2.0}.
\newblock \url{https://app.klingai.com/}, 2025.

\bibitem[Team(2024{\natexlab{b}})]{recraft2024v3}
Recraft Team.
\newblock {Recraft v3}.
\newblock \url{https://www.recraft.ai/}, 2024{\natexlab{b}}.

\bibitem[Touvron et~al.(2023)Touvron, Lavril, Izacard, Martinet, Lachaux, Lacroix, Rozi{\`e}re, Goyal, Hambro, Azhar, et~al.]{touvron2023llama1}
Hugo Touvron, Thibaut Lavril, Gautier Izacard, Xavier Martinet, Marie-Anne Lachaux, Timoth{\'e}e Lacroix, Baptiste Rozi{\`e}re, Naman Goyal, Eric Hambro, Faisal Azhar, et~al.
\newblock Llama: Open and efficient foundation language models.
\newblock \emph{arXiv preprint arXiv:2302.13971}, 2023.

\bibitem[Vaswani et~al.(2017)Vaswani, Shazeer, Parmar, Uszkoreit, Jones, Gomez, Kaiser, and Polosukhin]{vaswani2017transformer}
Ashish Vaswani, Noam Shazeer, Niki Parmar, Jakob Uszkoreit, Llion Jones, Aidan~N Gomez, {\L}ukasz Kaiser, and Illia Polosukhin.
\newblock Attention is all you need.
\newblock \emph{{Advances in Neural Information Processing Systems ({NeurIPS})}}, 2017.

\bibitem[Wan et~al.(2025)Wan, Wang, Ai, Wen, Mao, Xie, Chen, Yu, Zhao, Yang, et~al.]{wan2025wan}
Team Wan, Ang Wang, Baole Ai, Bin Wen, Chaojie Mao, Chen-Wei Xie, Di~Chen, Feiwu Yu, Haiming Zhao, Jianxiao Yang, et~al.
\newblock Wan: Open and advanced large-scale video generative models.
\newblock \emph{arXiv preprint arXiv:2503.20314}, 2025.

\bibitem[Wang et~al.(2024)Wang, Zhang, Luo, Sun, Cui, Wang, Zhang, Wang, Li, Yu, et~al.]{wang2024emu3}
Xinlong Wang, Xiaosong Zhang, Zhengxiong Luo, Quan Sun, Yufeng Cui, Jinsheng Wang, Fan Zhang, Yueze Wang, Zhen Li, Qiying Yu, et~al.
\newblock Emu3: Next-token prediction is all you need.
\newblock \emph{arXiv preprint arXiv:2409.18869}, 2024.

\bibitem[Wu et~al.(2025{\natexlab{a}})Wu, Li, Zhou, Lin, Gao, Yan, Yin, Bai, Xu, Chen, et~al.]{wu2025qwen}
Chenfei Wu, Jiahao Li, Jingren Zhou, Junyang Lin, Kaiyuan Gao, Kun Yan, Sheng-ming Yin, Shuai Bai, Xiao Xu, Yilei Chen, et~al.
\newblock Qwen-image technical report.
\newblock \emph{arXiv preprint arXiv:2508.02324}, 2025{\natexlab{a}}.

\bibitem[Wu et~al.(2025{\natexlab{b}})Wu, Zheng, Yan, Xiao, Luo, Wang, Li, Jiang, Liu, Zhou, et~al.]{wu2025omnigen2}
Chenyuan Wu, Pengfei Zheng, Ruiran Yan, Shitao Xiao, Xin Luo, Yueze Wang, Wanli Li, Xiyan Jiang, Yexin Liu, Junjie Zhou, et~al.
\newblock Omnigen2: Exploration to advanced multimodal generation.
\newblock \emph{arXiv preprint arXiv:2506.18871}, 2025{\natexlab{b}}.

\bibitem[Wu et~al.(2024{\natexlab{a}})Wu, Fei, Qu, Ji, and Chua]{wu2024next}
Shengqiong Wu, Hao Fei, Leigang Qu, Wei Ji, and Tat-Seng Chua.
\newblock Next-gpt: Any-to-any multimodal llm.
\newblock In \emph{{Proceedings of the International Conference on Machine Learning ({ICML})}}, 2024{\natexlab{a}}.

\bibitem[Wu et~al.(2024{\natexlab{b}})Wu, Zhang, Chen, Tang, Li, Fang, Zhu, Xie, Yin, Yi, et~al.]{wu2024vila}
Yecheng Wu, Zhuoyang Zhang, Junyu Chen, Haotian Tang, Dacheng Li, Yunhao Fang, Ligeng Zhu, Enze Xie, Hongxu Yin, Li~Yi, et~al.
\newblock Vila-u: a unified foundation model integrating visual understanding and generation.
\newblock \emph{arXiv preprint arXiv:2409.04429}, 2024{\natexlab{b}}.

\bibitem[Xiao et~al.(2025)Xiao, Wang, Zhou, Yuan, Xing, Yan, Li, Wang, Huang, and Liu]{xiao2025omnigen}
Shitao Xiao, Yueze Wang, Junjie Zhou, Huaying Yuan, Xingrun Xing, Ruiran Yan, Chaofan Li, Shuting Wang, Tiejun Huang, and Zheng Liu.
\newblock Omnigen: Unified image generation.
\newblock In \emph{{Proceedings of the {IEEE} Conference on Computer Vision and Pattern Recognition ({CVPR})}}, pages 13294--13304, 2025.

\bibitem[Xie et~al.(2024)Xie, Chen, Chen, Cai, Tang, Lin, Zhang, Li, Zhu, Lu, et~al.]{xie2024sana}
Enze Xie, Junsong Chen, Junyu Chen, Han Cai, Haotian Tang, Yujun Lin, Zhekai Zhang, Muyang Li, Ligeng Zhu, Yao Lu, et~al.
\newblock Sana: Efficient high-resolution image synthesis with linear diffusion transformers.
\newblock \emph{arXiv preprint arXiv:2410.10629}, 2024.

\bibitem[Xie et~al.(2025{\natexlab{a}})Xie, Chen, Zhao, Yu, Zhu, Wu, Lin, Zhang, Li, Chen, et~al.]{xie2025sana}
Enze Xie, Junsong Chen, Yuyang Zhao, Jincheng Yu, Ligeng Zhu, Chengyue Wu, Yujun Lin, Zhekai Zhang, Muyang Li, Junyu Chen, et~al.
\newblock Sana 1.5: Efficient scaling of training-time and inference-time compute in linear diffusion transformer.
\newblock \emph{arXiv preprint arXiv:2501.18427}, 2025{\natexlab{a}}.

\bibitem[Xie et~al.(2025{\natexlab{b}})Xie, Mao, Bai, Zhang, Wang, Lin, Gu, Chen, Yang, and Shou]{xie2024show}
Jinheng Xie, Weijia Mao, Zechen Bai, David~Junhao Zhang, Weihao Wang, Kevin~Qinghong Lin, Yuchao Gu, Zhijie Chen, Zhenheng Yang, and Mike~Zheng Shou.
\newblock Show-o: One single transformer to unify multimodal understanding and generation.
\newblock \emph{{Proceedings of the International Conference on Learning Representations ({ICLR})}}, 2025{\natexlab{b}}.

\bibitem[Xie et~al.(2025{\natexlab{c}})Xie, Yang, and Shou]{xie2025show}
Jinheng Xie, Zhenheng Yang, and Mike~Zheng Shou.
\newblock Show-o2: Improved native unified multimodal models.
\newblock \emph{arXiv preprint arXiv:2506.15564}, 2025{\natexlab{c}}.

\bibitem[Xu et~al.(2023)Xu, Xie, Tan, Huang, Howes, Sharma, Li, Ghosh, Zettlemoyer, and Feichtenhofer]{xu2023demystifying}
Hu~Xu, Saining Xie, Xiaoqing~Ellen Tan, Po-Yao Huang, Russell Howes, Vasu Sharma, Shang-Wen Li, Gargi Ghosh, Luke Zettlemoyer, and Christoph Feichtenhofer.
\newblock Demystifying clip data.
\newblock \emph{arXiv preprint arXiv:2309.16671}, 2023.

\bibitem[Ye et~al.(2025)Ye, He, Li, Lin, Yuan, Yan, Hou, and Yuan]{ye2025imgedit}
Yang Ye, Xianyi He, Zongjian Li, Bin Lin, Shenghai Yuan, Zhiyuan Yan, Bohan Hou, and Li~Yuan.
\newblock Imgedit: A unified image editing dataset and benchmark.
\newblock \emph{arXiv preprint arXiv:2505.20275}, 2025.

\bibitem[Yin et~al.(2024)Yin, Gharbi, Zhang, Shechtman, Durand, Freeman, and Park]{yin2024one}
Tianwei Yin, Micha{\"e}l Gharbi, Richard Zhang, Eli Shechtman, Fredo Durand, William~T Freeman, and Taesung Park.
\newblock One-step diffusion with distribution matching distillation.
\newblock In \emph{{Proceedings of the {IEEE} Conference on Computer Vision and Pattern Recognition ({CVPR})}}, pages 6613--6623, 2024.

\bibitem[Yu et~al.(2025)Yu, Chow, Yue, Pan, Wu, Wan, Li, Tang, Zhang, and Zhuang]{yu2025anyedit}
Qifan Yu, Wei Chow, Zhongqi Yue, Kaihang Pan, Yang Wu, Xiaoyang Wan, Juncheng Li, Siliang Tang, Hanwang Zhang, and Yueting Zhuang.
\newblock Anyedit: Mastering unified high-quality image editing for any idea.
\newblock In \emph{{Proceedings of the {IEEE} Conference on Computer Vision and Pattern Recognition ({CVPR})}}, pages 26125--26135, 2025.

\bibitem[Zhang and Sennrich(2019)]{zhang2019root}
Biao Zhang and Rico Sennrich.
\newblock Root mean square layer normalization.
\newblock \emph{{Advances in Neural Information Processing Systems ({NeurIPS})}}, 32, 2019.

\bibitem[Zhang et~al.(2023)Zhang, Mo, Chen, Sun, and Su]{zhang2023magicbrush}
Kai Zhang, Lingbo Mo, Wenhu Chen, Huan Sun, and Yu~Su.
\newblock Magicbrush: A manually annotated dataset for instruction-guided image editing.
\newblock \emph{{Advances in Neural Information Processing Systems ({NeurIPS})}}, 36:\penalty0 31428--31449, 2023.

\bibitem[Zhang et~al.(2025)Zhang, Xie, Lu, Yang, and Yang]{zhang2025context}
Zechuan Zhang, Ji~Xie, Yu~Lu, Zongxin Yang, and Yi~Yang.
\newblock In-context edit: Enabling instructional image editing with in-context generation in large scale diffusion transformer.
\newblock \emph{arXiv preprint arXiv:2504.20690}, 2025.

\bibitem[Zhao et~al.(2024)Zhao, Ma, Chen, Si, Wu, An, Yu, Zhang, Li, and Chang]{zhao2024ultraedit}
Haozhe Zhao, Xiaojian~Shawn Ma, Liang Chen, Shuzheng Si, Rujie Wu, Kaikai An, Peiyu Yu, Minjia Zhang, Qing Li, and Baobao Chang.
\newblock Ultraedit: Instruction-based fine-grained image editing at scale.
\newblock \emph{{Advances in Neural Information Processing Systems ({NeurIPS})}}, 37:\penalty0 3058--3093, 2024.

\bibitem[Zhou et~al.(2024)Zhou, Yu, Babu, Tirumala, Yasunaga, Shamis, Kahn, Ma, Zettlemoyer, and Levy]{zhou2024transfusion}
Chunting Zhou, Lili Yu, Arun Babu, Kushal Tirumala, Michihiro Yasunaga, Leonid Shamis, Jacob Kahn, Xuezhe Ma, Luke Zettlemoyer, and Omer Levy.
\newblock Transfusion: Predict the next token and diffuse images with one multi-modal model.
\newblock \emph{arXiv preprint arXiv:2408.11039}, 2024.

\end{thebibliography}

\clearpage
\newpage
\beginappendix

\section{Implementation Details}
To support efficient and stable training, we introduce a set of optimizations spanning both system-level infrastructure and model design: 

\begin{itemize}
    \item \textit{QK-Norm:} To enhance training stability, we apply QK-Norm \citep{henry2020query} in each transformer block. Specifically, before the attention operation, we normalize the query and key vectors using RMS-Norm \citep{zhang2019root}.
    \item \textit{Modality-specific Norm:} We apply separate normalization layers for different modalities, which improves performance while maintaining training stability. 
    \item \textit{Token Registers:} Following \citet{darcet2023vision}, we introduce four auxiliary learnable tokens into the input sequence to enhance training stability, without assigning them explicit training objectives.
    \item \textit{Diffusion Step Sampling:} To accelerate convergence during low-resolution (512$\times$512) pretraining, we adopt logit-normal sampling, which is later replaced by mode sampling (scale = 0.8, shift = 3.0) to adapt to high-resolution (1024$\times$1024 and 2048 $\times$ 2048) training.
    \item \textit{Dropping timestep embeddings.} Motivated by recent findings \citep{tang2025exploring,sun2025noise} and the empirical analysis, we confirm that timestep embeddings provide negligible benefit to the diffusion model while introducing an overhead of $\sim$20\% parameters. For efficiency, we remove the timestep conditioning from the generation tower.
    \item \textit{FSDP:} We adopt Fully Sharded Data Parallel (FSDP) as our primary distributed training framework and enable activation checkpointing in the high-resolution stages.
    \item \textit{Low-Precision Training}: We employ a module-specific mixed-precision strategy. The VAE compressor is maintained in float32 to ensure numerical stability, while the understanding tower is set entirely in bfloat16. For the generation tower and router, we use bfloat16 for all-gather operations and float32 for gradient reduce-scatter.
    \item \textit{Triton Kernel Optimization:} To further improve training throughput, we employ custom Triton kernels, including an RMSNorm kernel and a fused FFN kernel. Our implementation builds on the liger-kernel \citep{hsu2025ligerkernel}.
    \item \textit{Bucket-wise Dynamic Resolution Training:} To support dynamic-resolution training, we adopt a resolution-driven bucket dataloader. Data samples are assigned online to buckets based on their resolution and aspect ratio. Once a bucket is filled, it is dispatched to the model for training. 
    \item \textit{Data Reweighting:} To achieve balanced performance across different dimensions, we adjust the mixing ratios of the training datasets. The optimal ratios are determined through grid search.
\end{itemize}

Regarding the hyperparameters, we use AdamW with a learning rate of $1\times10^{-4}$, weight decay of 0.01, and betas set to $(0.9, 0.95)$. The first 4k steps serve as a warm-up phase, where the learning rate is linearly increased to the target value. A cosine schedule is then applied to gradually decay the learning rate to $1.5\times10^{-5}$. The global batch size is dynamically set to 2048 or 1024, depending on available training resources. For each pre-training stage, we run 400k–1200k steps based on visual inspection of convergence, while HQ fine-tuning is performed for 50k steps. We use a top-k router ($k=2$) with $\epsilon$-greedy exploration ($\epsilon=0.05$). The training platform comprises both A100 and H100 GPUs; to standardize reporting, we compute the total training cost by counting 2 A100 days as equivalent to 1 H100 day.
\newpage

\section{Additional Discussions on Router Designs}
\begin{figure}[ht!]
    \centering
    \begin{subfigure}{0.48\linewidth}
        \includegraphics[width=\linewidth]{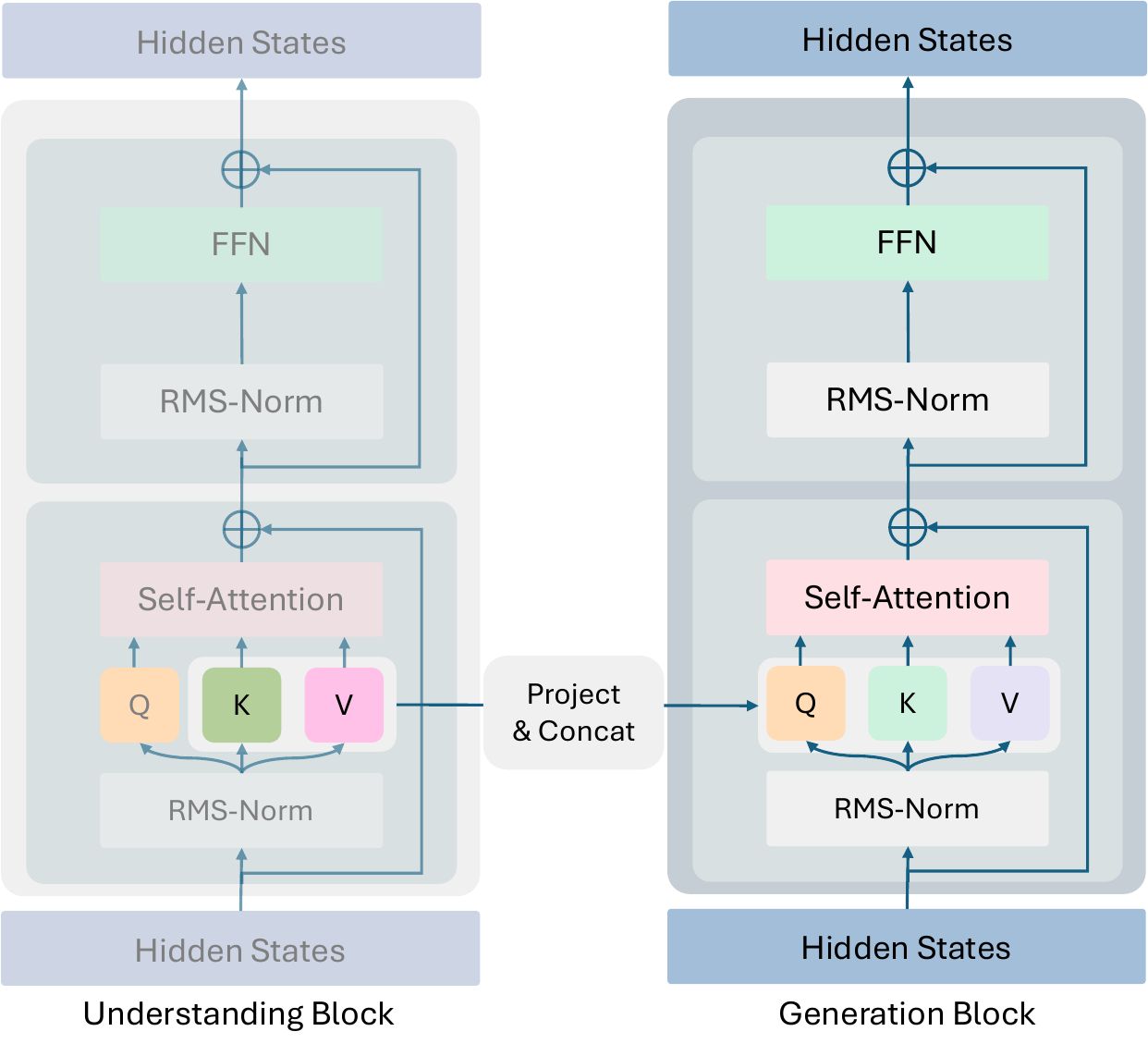}
                \caption{Global Attention}
    \end{subfigure}
    \begin{subfigure}{0.48\linewidth}
        \includegraphics[width=\linewidth]{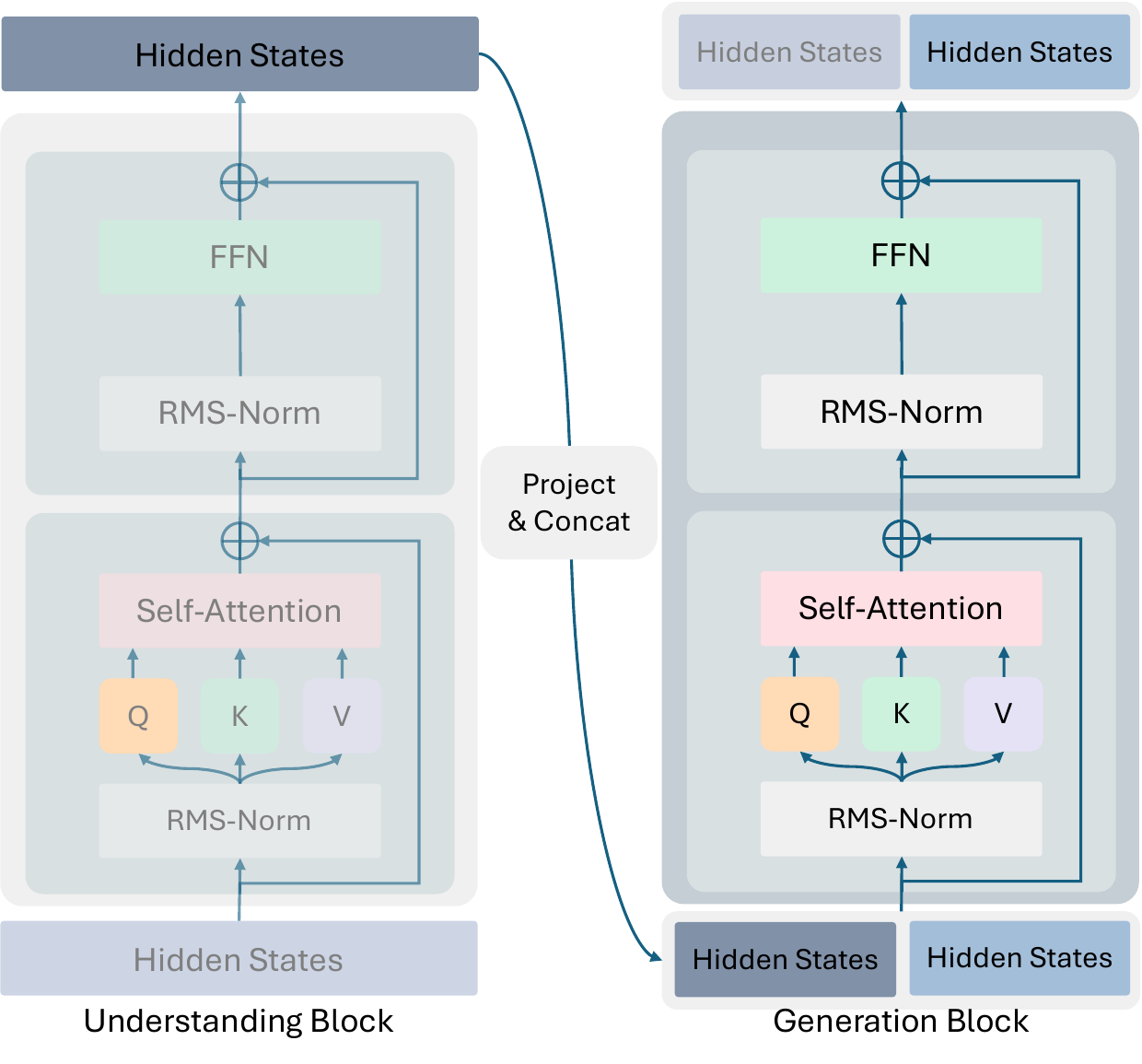}
            \caption{Global Hidden States}
    \end{subfigure}
    \caption{\textbf{Illustration of the router’s operation space.}
(a) In the global-attention scheme, keys and values from the understanding branch are projected to the generation branch for cross-modal interaction. 
(b) In the global hidden-state scheme, hidden representations from both branches are directly concatenated at the input of each transformer block.}
    \label{fig:ab1}
\end{figure}

This ablation study aims to determine the MoS router’s operation space, i.e., which features routed across transformers yield the optimal benefit.
We begin with the MoT architecture \citep{liang2025mixtureoftransformers}. To enable representation transfer across transformers, MoT introduces a global attention mechanism where keys, values, and queries are shared between towers.
In contrast to MoT, a competing approach fuses hidden states directly before sequence modeling. As shown in Fig.~\ref{fig:ab1} (a)-(b), these two design philosophies yield four candidates:
\begin{itemize}
\item \textit{Global Attention (Head-Projection)}\footnote{Here, we avoid using the term layer-wise attention to describe this operation, to maintain symmetry with the case of global hidden states. Nonetheless, the mechanism is essentially equivalent to the layer-wise attention discussed in previous sections.}: Apply a projection layer on each head dimension for the key and value vectors from the understanding tower, then concatenate with the generation tower’s key and value representations, respectively.
\item \textit{Global Attention (State-Projection)}: Apply the projection layer on the hidden dimension, split into multi-head vectors, and fuse with the generation tower’s features.
\item \textit{Global Hidden States (Independent-Projection)}: Concatenate hidden states in the generation tower, then apply separate key-value projection layers before attention.
\item \textit{Global Hidden States (Shared Projection)}: Concatenate hidden states, then apply a shared key–value projection layer.
\end{itemize}

\subsection{Router Operation Space}
\label{sec:opt_space}

\begin{figure}[t]
\centering
\begin{subfigure}{0.245\textwidth}
  \includegraphics[width=\linewidth]{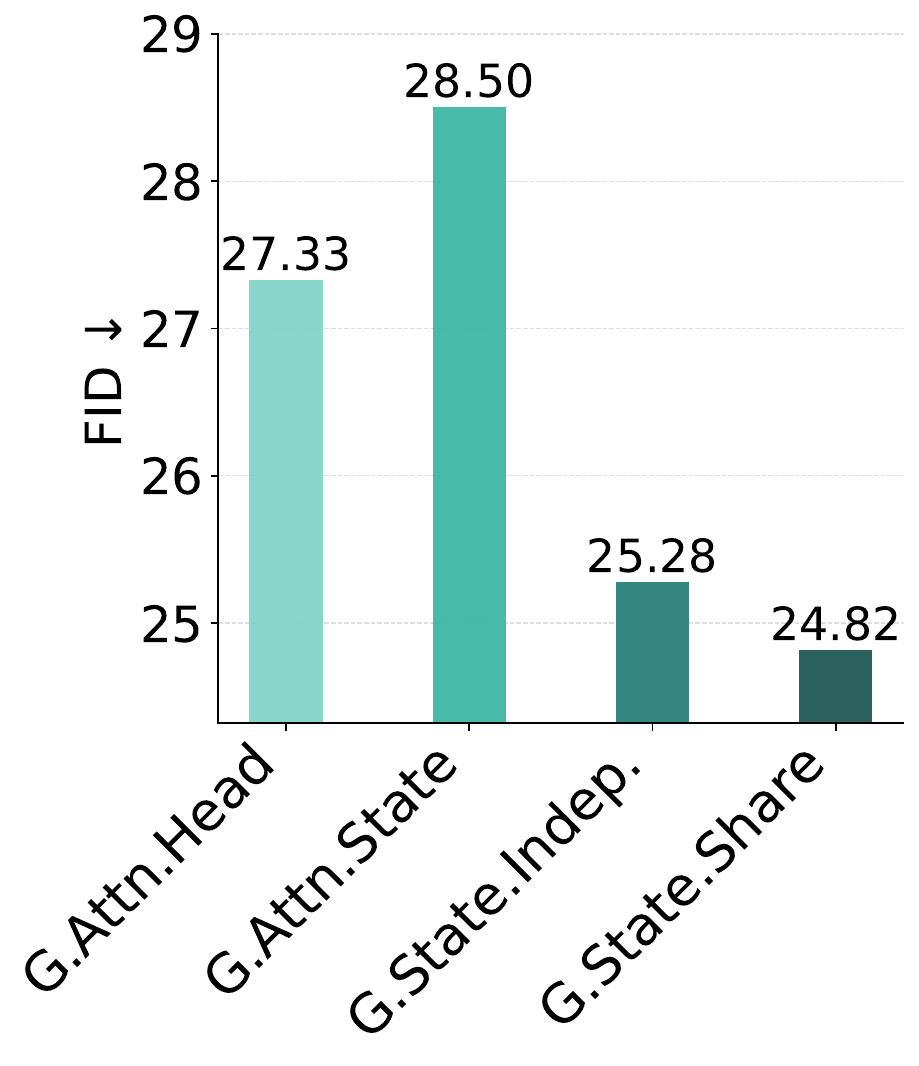}
  \caption*{(a) FID comparison across feature types used for interaction.}
\end{subfigure}\hfill
\begin{subfigure}{0.245\textwidth}
  \includegraphics[width=\linewidth]{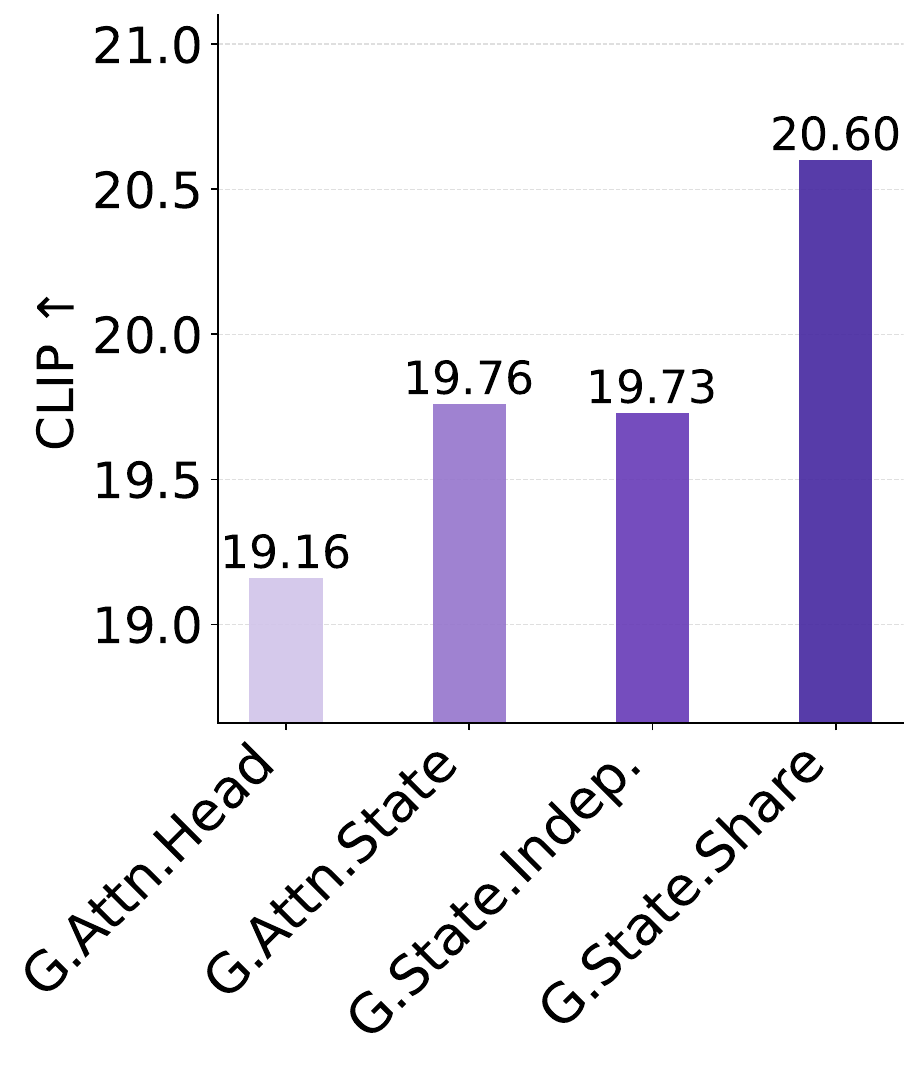}
  \caption*{(b) CLIP comparison across feature types used for interaction.}
\end{subfigure}\hfill
\begin{subfigure}{0.245\textwidth}
  \includegraphics[width=\linewidth]{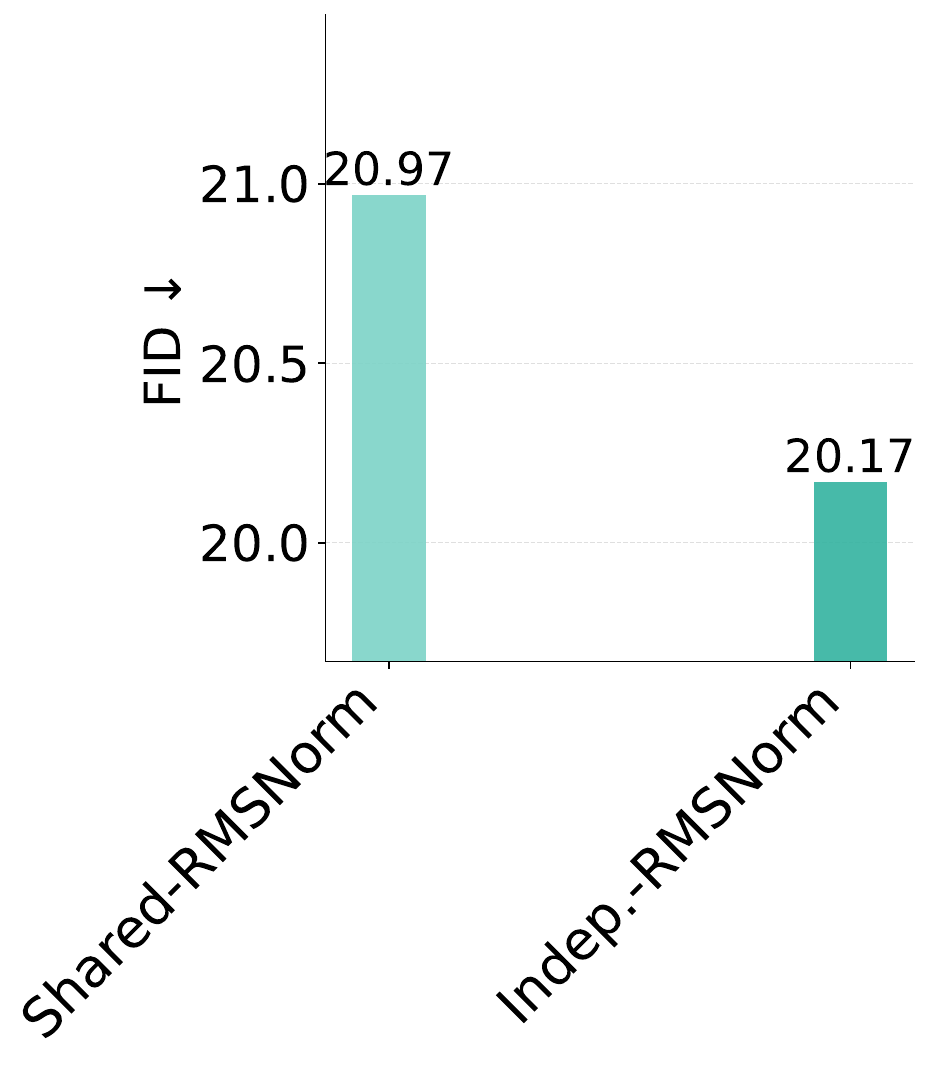}
  \caption*{(c) FID performance with and without separate normalization in the router.}
\end{subfigure}\hfill
\begin{subfigure}{0.245\textwidth}
  \includegraphics[width=\linewidth]{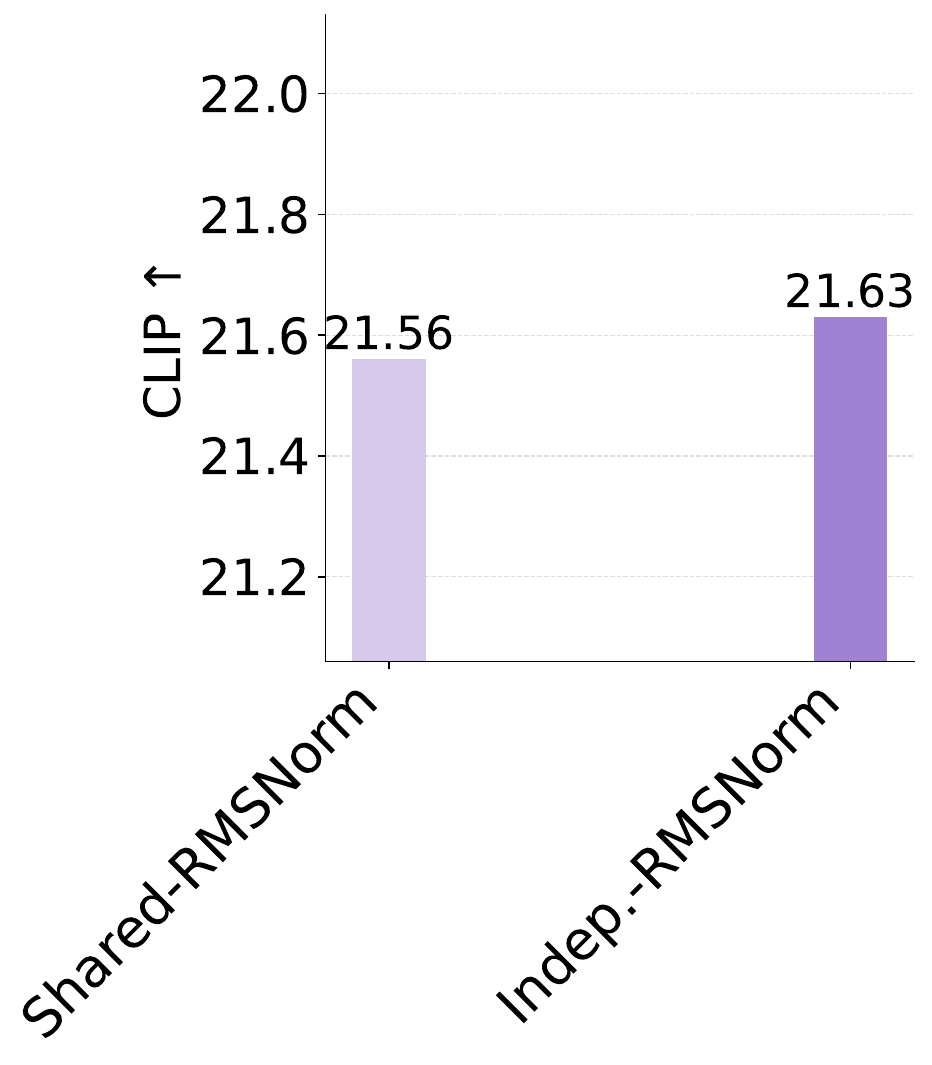}
  \caption*{(d) CLIP performance with and without separate normalization in the router.}
\end{subfigure}
\caption{\textbf{Ablation study results on FID and CLIP across the router's operation space and architectural design.}
(a)–(b) indicate that using hidden states as the router’s operation space outperforms using key/value features, while (c)–(d) show that applying modality-specific normalization to the router’s inputs further improves performance. }
\label{fig:ab_1234_results}
\end{figure}
As shown in Fig.~\ref{fig:ab_1234_results} (a)-(b), the empirical results clearly indicate that using global hidden states with a shared projection layer yields the optimal configuration. Note that the MoS router is excluded from this ablation.

\begin{figure}[t]
\centering
\begin{subfigure}{0.245\textwidth}
  \includegraphics[width=\linewidth]{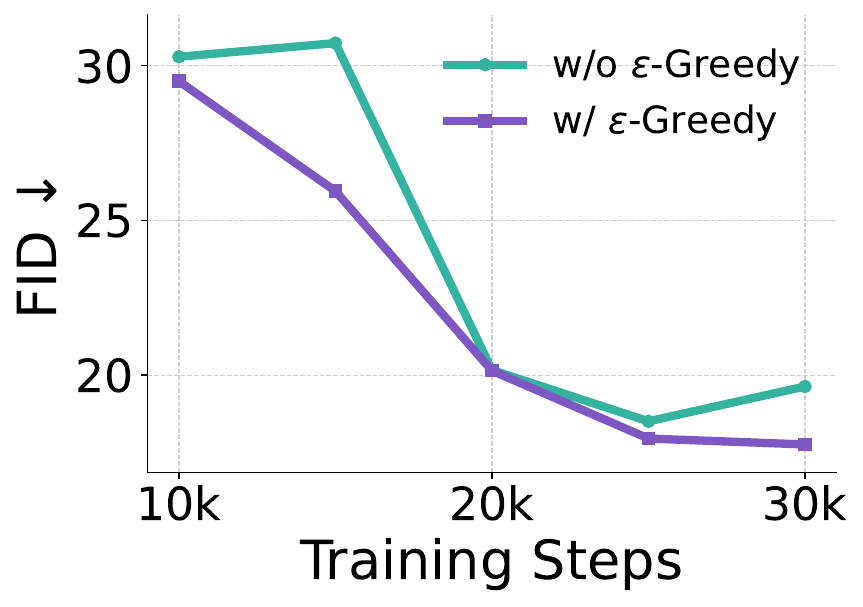}
  \caption*{(a) FID performance by using  $\epsilon$-greedy.}
\end{subfigure}\hfill
\begin{subfigure}{0.245\textwidth}
  \includegraphics[width=\linewidth]{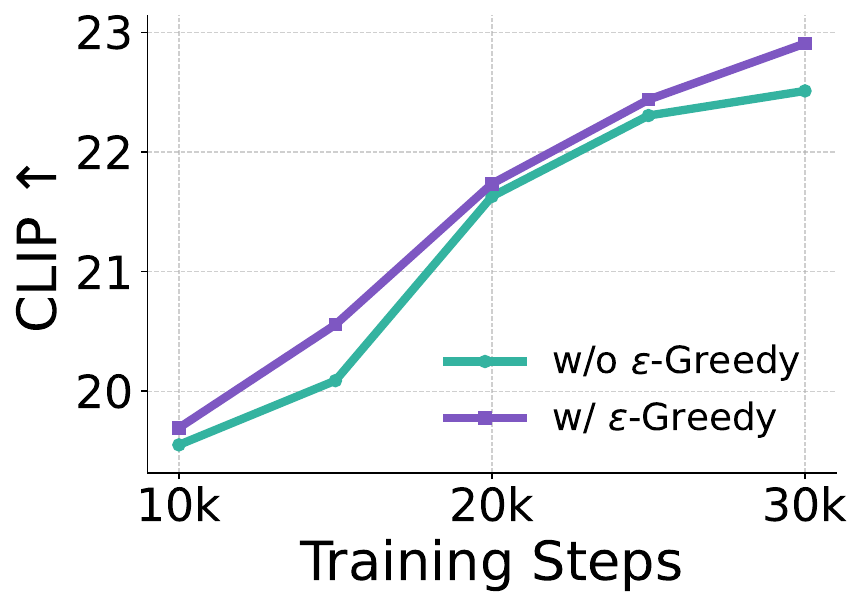}
  \caption*{(b) CLIP performance by using  $\epsilon$-greedy.}
\end{subfigure}\hfill
\begin{subfigure}{0.245\textwidth}
  \includegraphics[width=\linewidth]{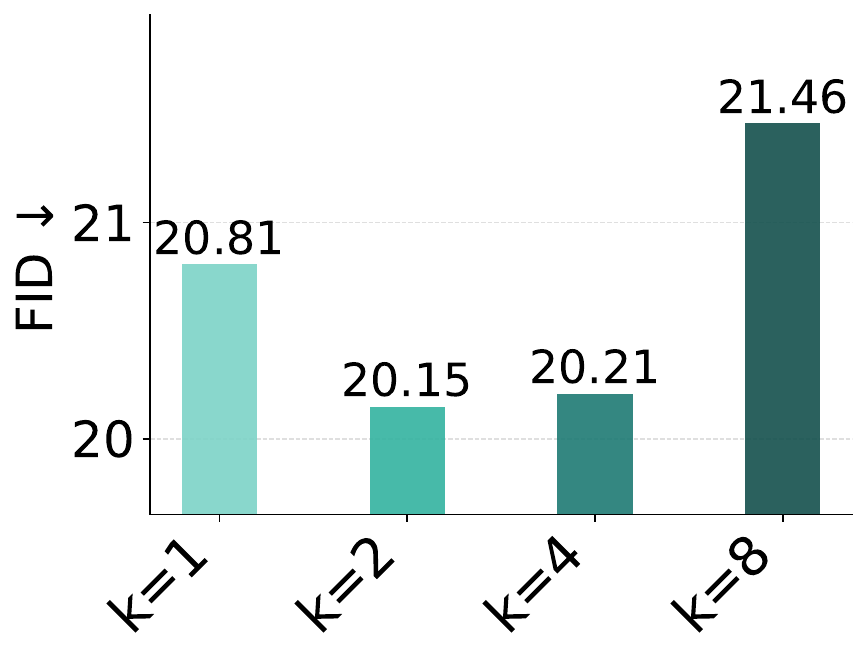}
  \caption*{(c) FID performance across different top-k settings.}
\end{subfigure}\hfill
\begin{subfigure}{0.245\textwidth}
  \includegraphics[width=\linewidth]{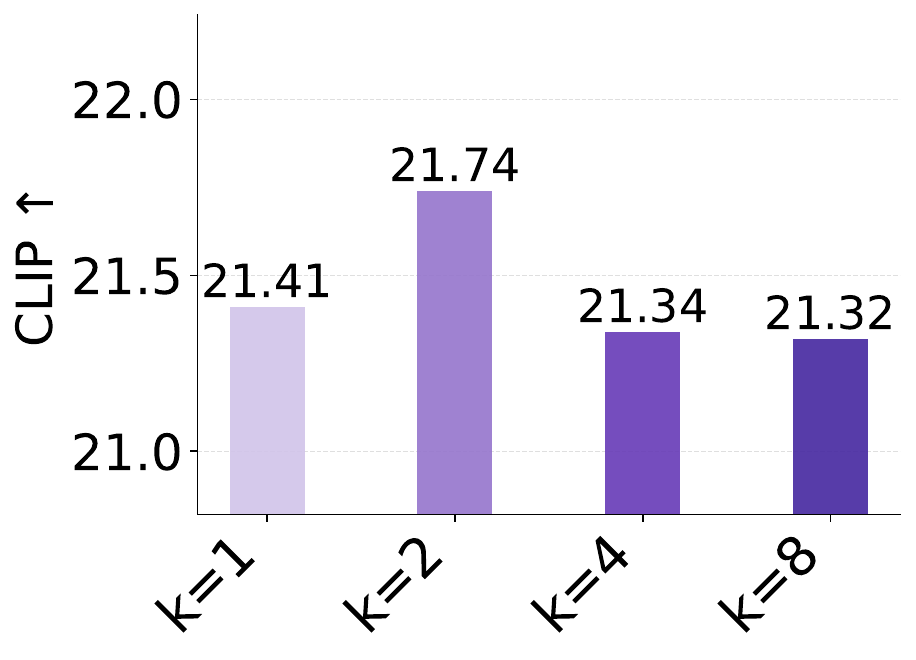}
  \caption*{(d) CLIP performance across different top-k settings.}
\end{subfigure}
\caption{\textbf{Ablation study on the $\epsilon$-greedy exploration strategy and sparsity settings.} 
Results show that incorporating $\epsilon$-greedy accelerates convergence, and that $k=2$ yields the best performance.}
\label{fig:ab_56_results}
\end{figure}

\begin{figure}[h]
\centering
\begin{subfigure}{0.245\textwidth}
  \includegraphics[width=\linewidth]{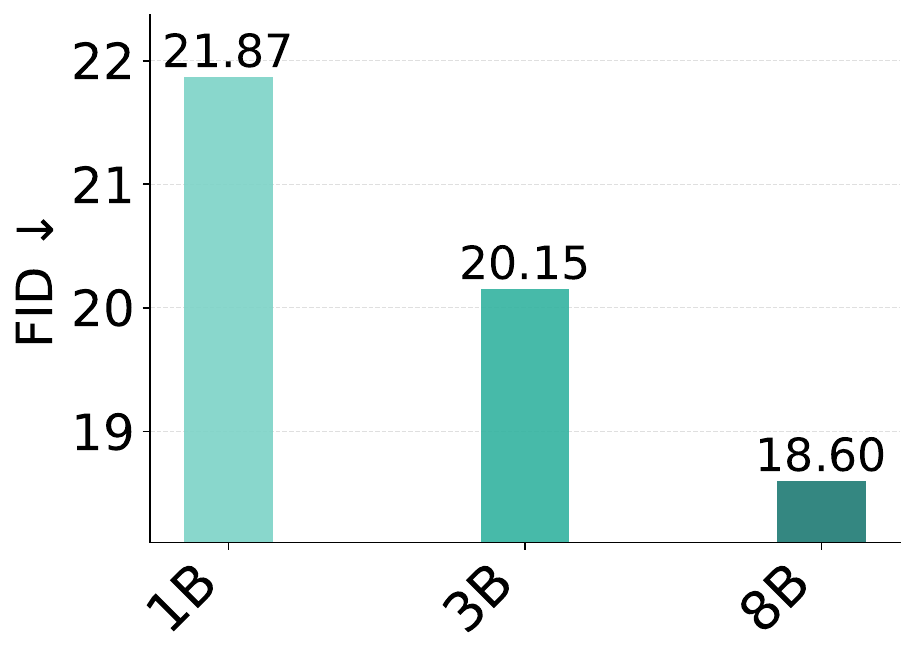}
  \caption*{(a) FID performance by using different Und. Tower.}
\end{subfigure}\hfill
\begin{subfigure}{0.245\textwidth}
  \includegraphics[width=\linewidth]{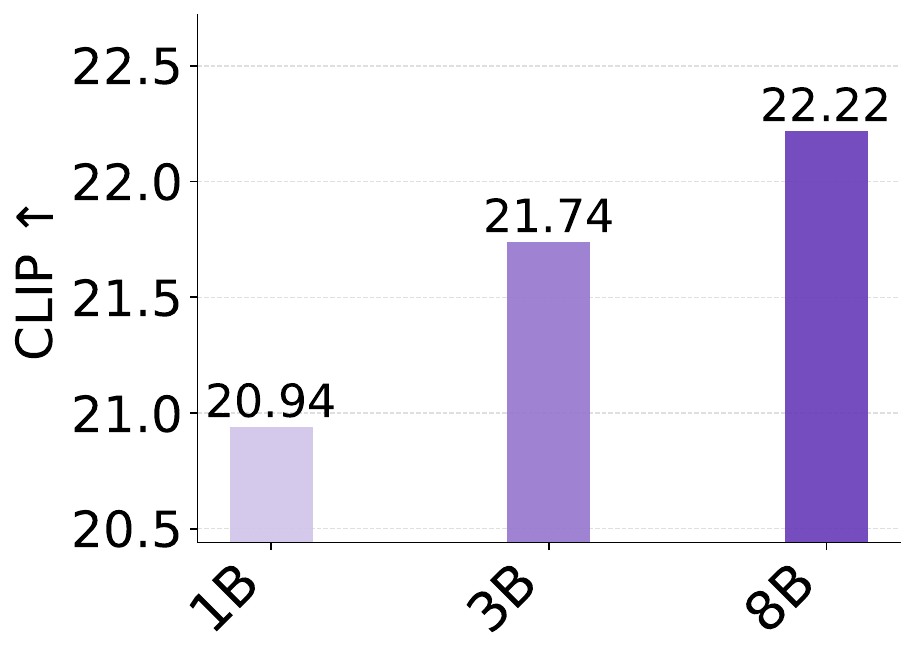}
  \caption*{(b) CLIP performance by using different Und. Tower.}
\end{subfigure}\hfill
\begin{subfigure}{0.245\textwidth}
  \includegraphics[width=\linewidth]{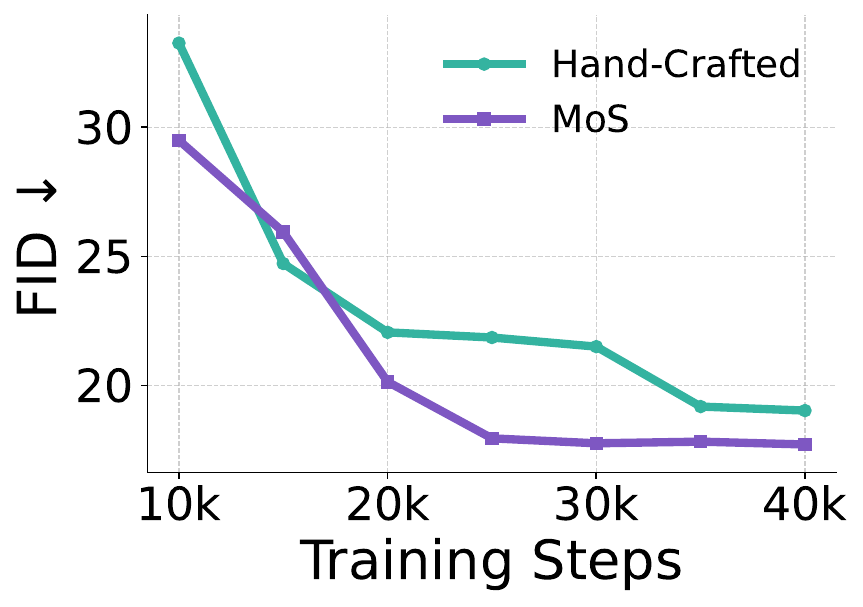}
  \caption*{(c) FID performance across different interaction setting.} 
\end{subfigure}\hfill
\begin{subfigure}{0.245\textwidth}
  \includegraphics[width=\linewidth]{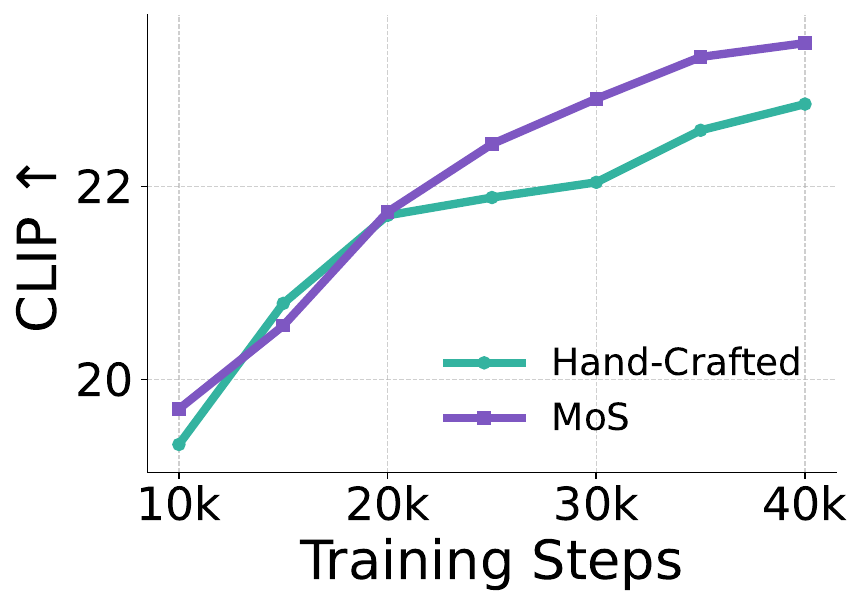}
  \caption*{(d) CLIP performance across different interaction setting.}
\end{subfigure}
\caption{\textbf{Ablation study results on FID and CLIP for understanding tower size and interaction types.} 
Our ablations show that MoS interaction consistently outperforms hand-crafted design, while also benefiting from scaling up the understanding tower.}
\label{fig:ab_78_results}
\end{figure}

\subsection{Router Architecture Design}
\label{sec:router_arch}
In the MoS router, token embeddings from different modalities are normalized using separate RMSNorm layers to align their representation scales. Our analysis shows that this design is a key factor in improving performance. To verify its effectiveness, we compare two configurations: (a) a shared RMSNorm applied to all modalities, and (b) separate RMSNorms for each modality. Fig.~\ref{fig:ab_1234_results}(c–d) shows that configuration (b) consistently outperforms (a) across all evaluation metrics. 

\begin{figure}[ht]
\centering
\begin{subfigure}{0.33\textwidth}
  \includegraphics[width=\linewidth]{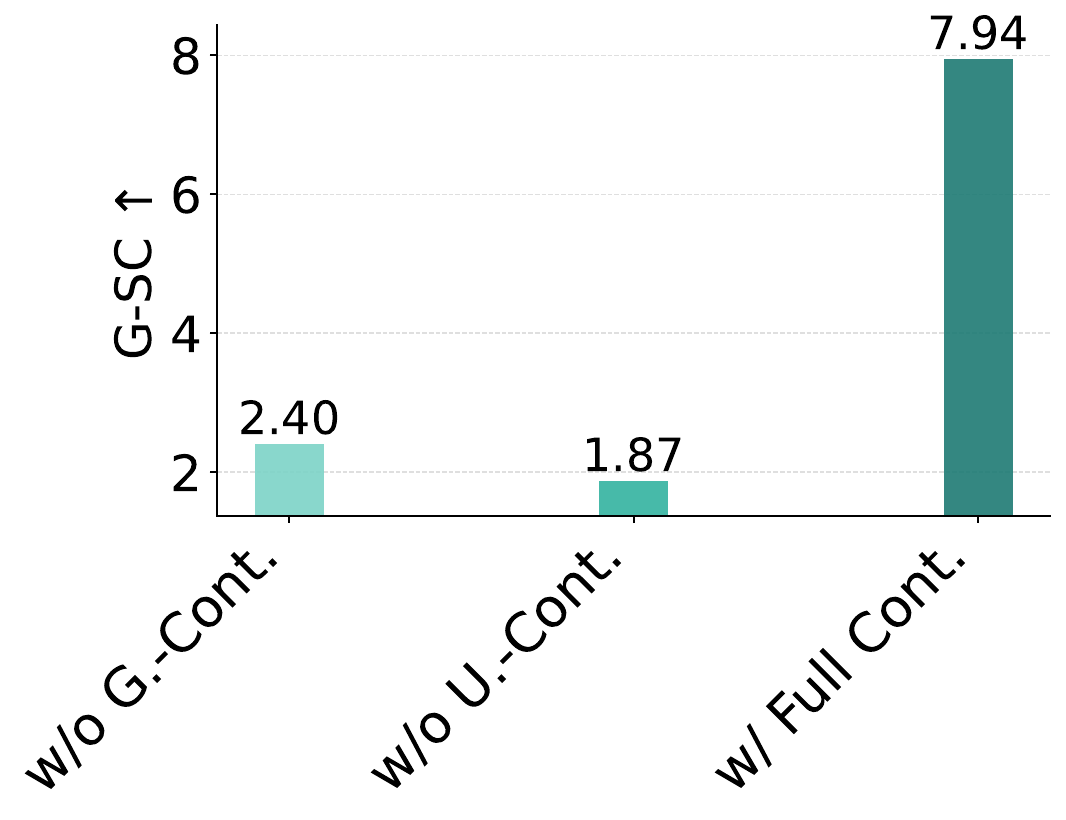}
  \caption*{(a) G-SC (Semantic Consistency) Performance with varying context.}
\end{subfigure}\hfill
\begin{subfigure}{0.33\textwidth}
  \includegraphics[width=\linewidth]{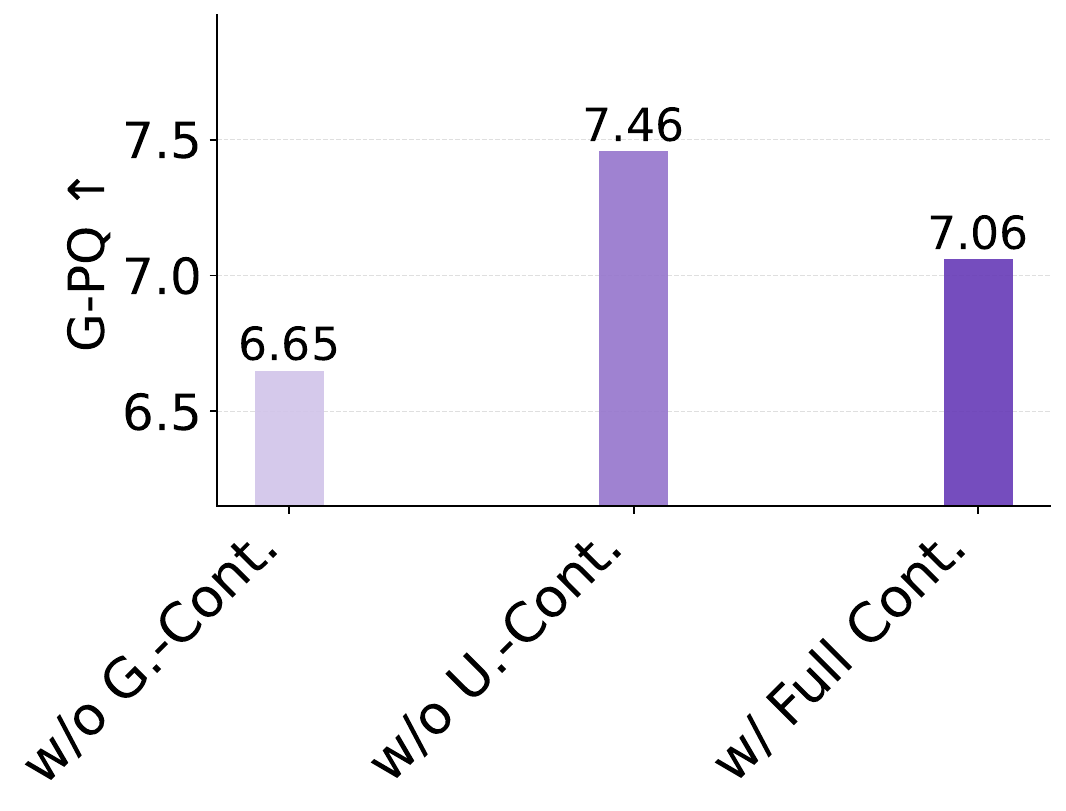}
  \caption*{(b) G-PQ (Perceptual Quality) Performance  with varying context.}
\end{subfigure}\hfill
\begin{subfigure}{0.33\textwidth}
  \includegraphics[width=\linewidth]{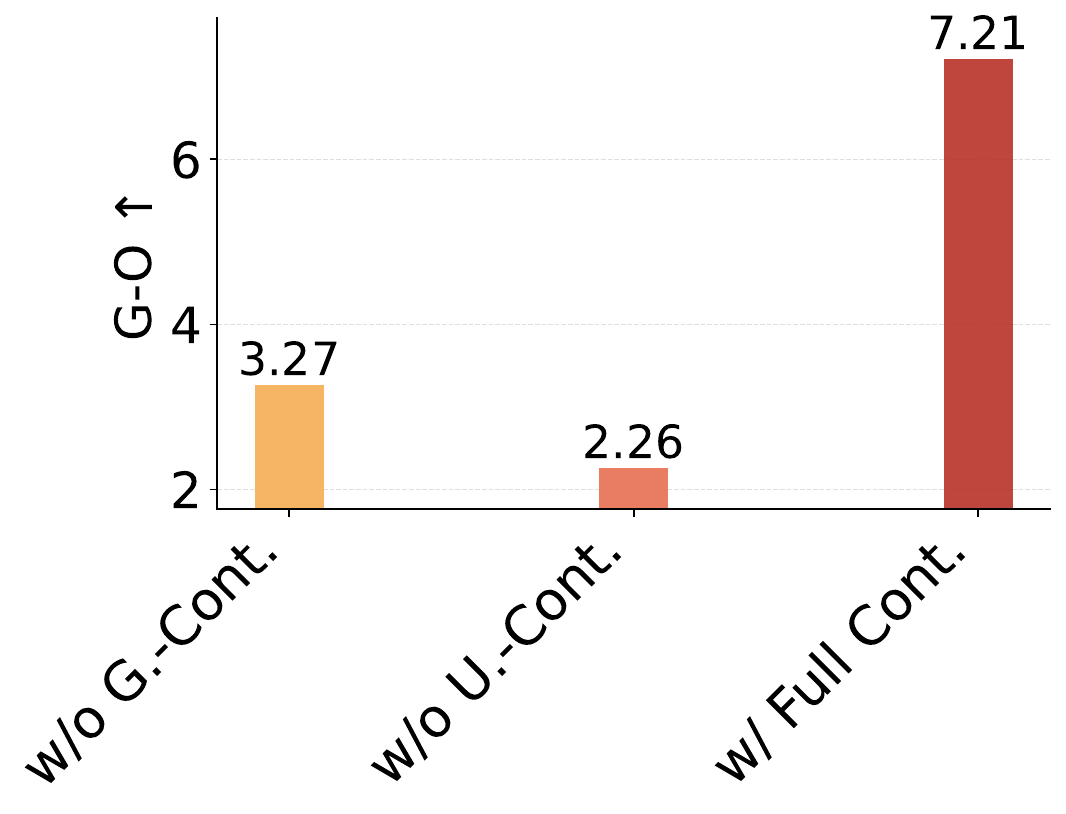}
  \caption*{(c) G-O (Overall Score) Performance with varying context.}
\end{subfigure}
\caption{\textbf{Ablation study results on GEdit-Bench \citep{liu2025step1x_edit}.} The best performance is obtained when the source images are provided to both the generation and editing towers.}
\label{fig:ab_9_results}
\end{figure}
\begin{figure}
\centering
\begin{subfigure}{0.245\textwidth}
  \includegraphics[width=\linewidth]{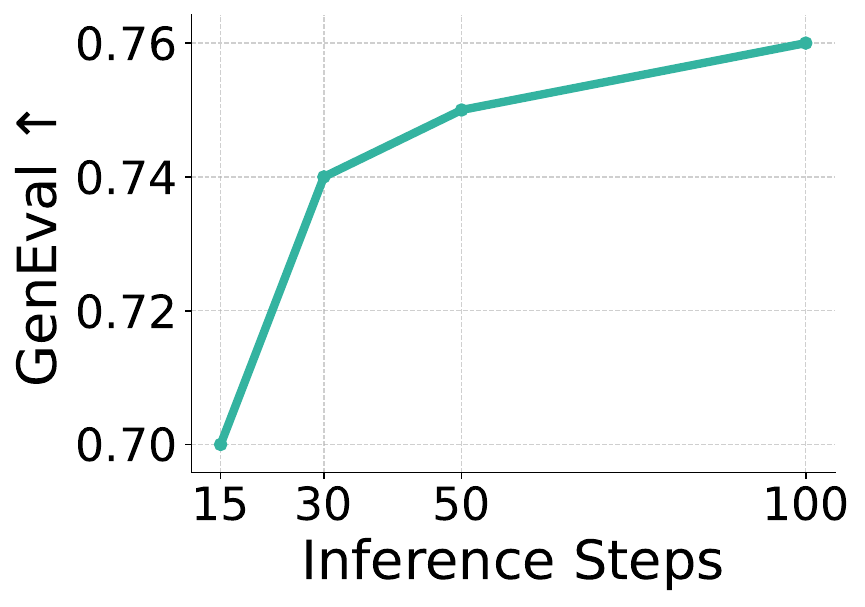}
  \caption*{(a) GenEval Score using different inference steps}
\end{subfigure}\hfill
\begin{subfigure}{0.245\textwidth}
  \includegraphics[width=\linewidth]{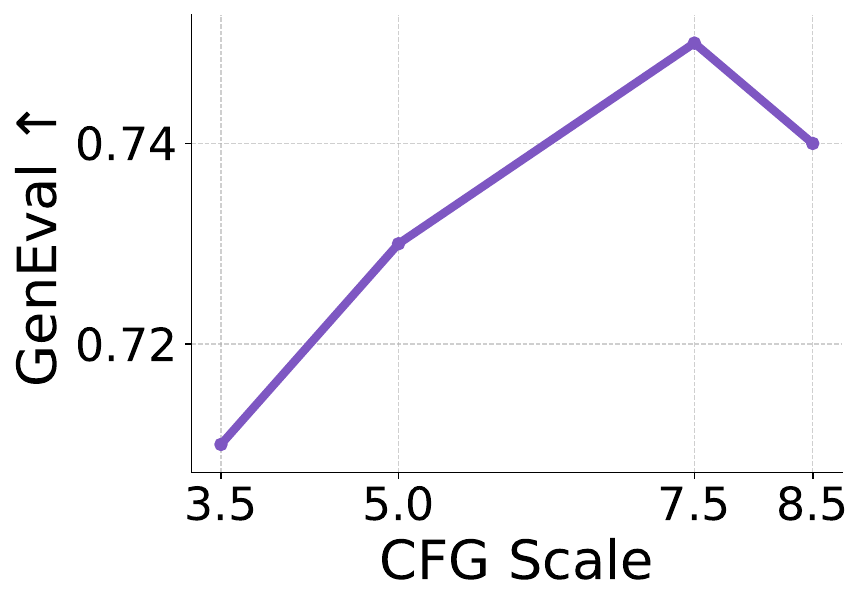}
  \caption*{(b) GenEval Score using different CFG scales.}
\end{subfigure}\hfill
\begin{subfigure}{0.245\textwidth}
  \includegraphics[width=\linewidth]{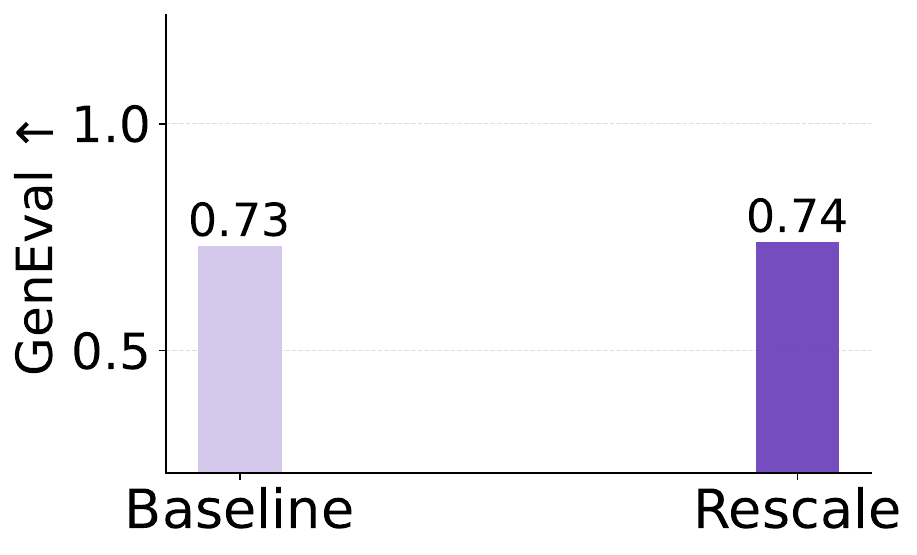}
  \vspace{-0.3em}
  \caption*{(c) GenEval Score using or not using rescale strategy.}
\end{subfigure}
\begin{subfigure}{0.245\textwidth}
  \includegraphics[width=\linewidth]{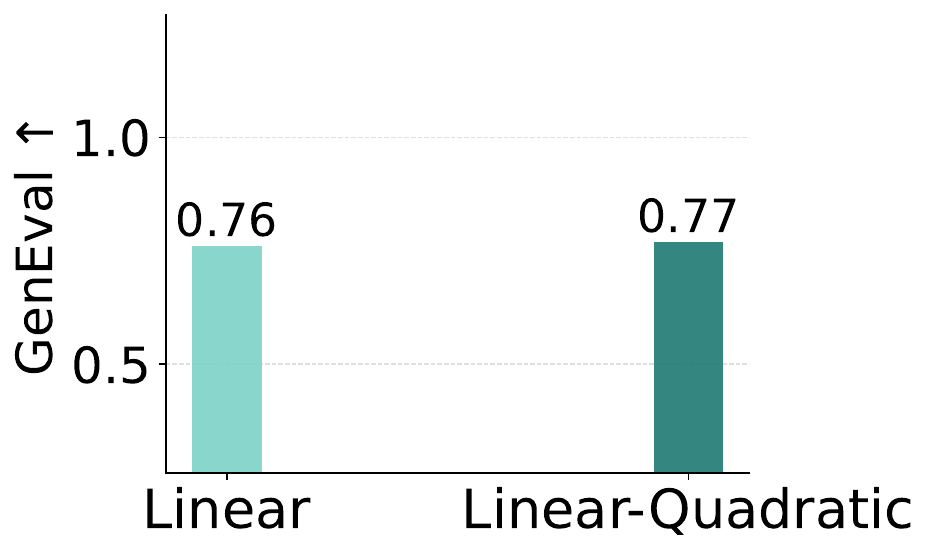}
  \vspace{-0.3em}
  \caption*{(d) GenEval Score using different schedulers.}
\end{subfigure}
\caption{\textbf{Ablation study results on GenEval \citep{ghosh2023geneval} with different inference strategies.} 
MoS exhibits behavior similarly to other diffusion models. Incorporating commonly adopted enhancements into the inference process consistently improves performance. }
\label{fig:ab_10_results}
\end{figure}

\subsection{$\epsilon$-greedy Strategy and Sparsity Design}
\label{sec:router_explore}
To evaluate the impact of applying $\epsilon$-greedy to the router’s output, we conduct an ablation study. Here, $\epsilon=0.05$, meaning that with a 5\% probability the router randomly selects a layer rather than following the predicted logits. This choice is informed by an empirically grid search. As shown in Fig.~\ref{fig:ab_56_results}, the results indicate that incorporating $\epsilon$-greedy notably accelerates convergence across training steps. 
Next, we study how many layers ($k$) should be consolidated to form the final guidance feature. As shown in Fig.~\ref{fig:ab_56_results}, $k=2$ consistently outperforms other candidates. This is reasonable: when $k=1$, the model may become trapped in a local view, as the router tends to overfit to a single layer; conversely, larger $k$ values dilute representations, over-flatten hidden states, and ultimately degrade performance.

\subsection{Scalability of MoS}
\label{sec:mos_scale}
Here, we validate the scalability of our model. Prior studies \citep{cai2025hidream,esser2024scaling,polyak2024moviegen,wu2025qwen} have demonstrated the effectiveness of scaling the generation tower (diffusion model). Since our approach does not alter the fundamental formulation of the diffusion process, it should likewise benefit from enlarging the generation tower. In this section, however, we focus on a complementary direction—scaling the understanding tower. Unlike MoT, MoS provides a more flexible framework that enables independent scaling of the understanding tower, thereby allowing the use of larger understanding models. As shown in Fig.~\ref{fig:ab_78_results} (a)-(b), we find that enlarging the text encoder yields consistent and stable improvements as its size increases. Moreover, since the understanding tower does not need to be updated across all training stages and its embeddings can be provided in a Producer–Consume manner, scaling the understanding tower emerges as a cost-efficient solution based on our empirical analysis.

\begin{figure}[h]
    \centering
    \includegraphics[width=\linewidth]{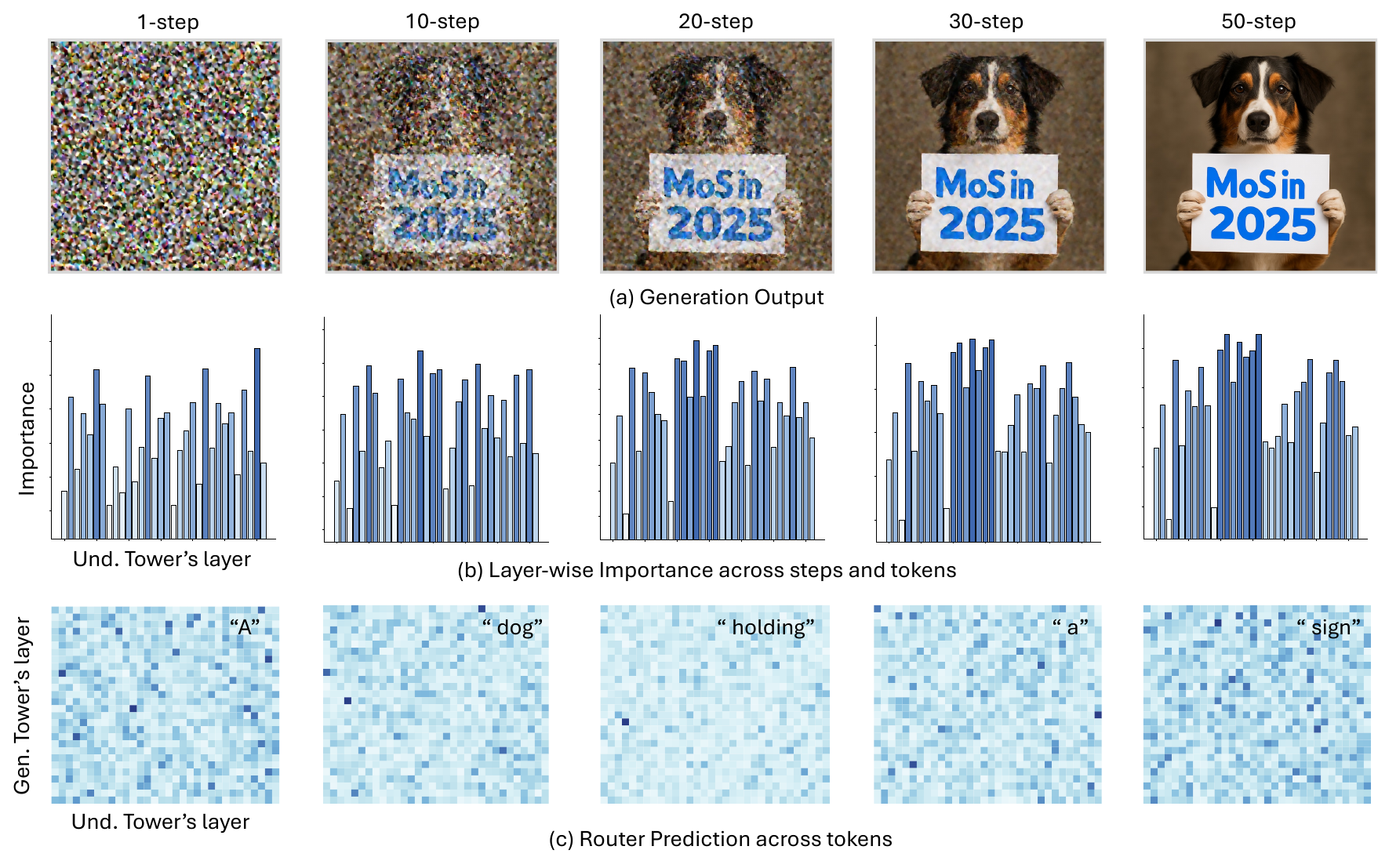}
        \vspace{-1em}
    \caption{\textbf{Visualization of the Router across Time Steps.} The results show that different tokens induce distinct connection patterns, indicating that the router dynamically adjusts its layer-to-layer routing based on token-specific semantics.}
    \label{fig:router_vis_per_steps}
\end{figure}

\subsection{MoS for Image Editing}
\label{sec:mos_editing}
Our design feeds the reference image into both the understanding and generation towers. 
We hypothesize that this allows the model to leverage semantic information from the understanding tower and low-level visual features from the generation tower, thereby enabling more precise and consistent editing.
To validate the effectiveness of our design, we compare three input configurations for image editing: 
(i) \textit{w/o generation-tower context}, where the reference image input to the generation tower is removed; 
(ii) \textit{w/o understanding-tower context}, where the reference image input to the understanding tower is removed; and 
(iii) \textit{w/ full context}, where both towers receive the source images.
As shown in Fig.~\ref{fig:ab_9_results}, we conduct experiments on GEdit-Bench \citep{liu2025step1x_edit}, which evaluates model editing performance across three dimensions: semantic consistency (G-SC), perceptual quality (G-PQ), and overall score (G-O). All metrics are obtained from GPT-4o–based automatic evaluations \citep{hurst2024gpt}, where higher values indicate better performance. The results demonstrate that incorporating reference images from both towers achieves the highest average score, consistent with our hypothesis.

\subsection{Ablation Study on Inference Strategy}
\label{sec:com_inf}
MoS incorporates denoising steps into its input, which may influence the underlying diffusion dynamics. We thereby empirically validate whether common inference enhancements can benefit MoS-based models. As shown in Fig.~\ref{fig:ab_10_results}, we evaluate an intermediate checkpoint of \textsc{MoS-S} (trained for 400k steps at $512 \times 512$ resolution) on GenEval \citep{ghosh2023geneval}.
The results show that increasing inference steps consistently improves generation quality, and CFG guidance can be applied in the typical range (5.0–7.5), similar to the other models \cite{podell2023sdxl}. We further observe that adopting a linear–quadratic scheduler \citep{polyak2024movie} and rescaling strategy \citep{lin2024common} yields slight additional gains.

\subsection{MoS Router Visualization Analysis} 
\label{sec:router_visualize}
To analyze the router’s behavior, we visualize its output patterns in Fig.~\ref{fig:router_vis_per_steps} using the caption “A dog holding a sign that says ‘MoS in 2025’” with \textsc{MoS-S}: 
i) The first row shows the denoising trajectory, where the model progressively refines the image from pure noise to the target output, guided by the input caption.
ii) The second row visualizes the average contribution of each understanding layer. To obtain this, we compute the router’s logit matrix---modeling the affinity between blocks in the understanding and generation towers---and average the weights across all generation blocks and tokens.
iii) The third row presents the router’s output at a fixed denoising step ($t{=}1$) for individual tokens, revealing that different tokens induce distinct routing patterns.
The results indicate that: 
\begin{itemize}
    \item The router’s predictions vary across denoising steps. In the early stages, features from layers of different depths are sparsely selected as the most influential. As the denoising process progresses, the weights of the middle layers gradually increase, leading to smoother importance distributions and reduced variation across steps. This trend is intuitive: at later stages, most semantic information has been established, and the model no longer requires highly specific features from individual layers. This observation is consistent with the findings reported in \cite{liu2025faster}.
    \item The router’s predictions also vary across tokens. As shown in Fig.~\ref{fig:router_vis_per_steps}, each token exhibits a distinct connection pattern, reflecting the router’s ability to adapt its routing strategy to token-specific semantics. This observation aligns with our ablation study, which demonstrates that token-wise prediction yields better performance than sample-wise prediction. 
    \item Since the router is jointly optimized with the generation tower, it can be regarded as a surrogate mechanism that approximates an optimal routing strategy. However, our analysis provides no evidence that the final-layer embedding serves as an effective solution. Similarly, we find no consistent pattern indicating a strict layer-to-layer correspondence in MoT. These findings suggest that previous designs may not fully leverage the capacity of the understanding tower, thereby supporting our hypothesis and underlying motivation.
\end{itemize}

\begin{table}[!h]
\centering
    \footnotesize
\setlength{\tabcolsep}{0.3em}  
\renewcommand{\arraystretch}{1.0}
\caption{\textbf{Performance of Foundational Image Generation Models on the GenEval Benchmark \citep{ghosh2023geneval}.} GenEval evaluates object-level prompt alignment in text-to-image models.
Greyed rows denote models with unclear configuration references, which hinders fair comparison. 
MoS-L attains the strongest results across several dimensions and delivers the best overall performance.}
\label{tab:geneval}
\begin{tabular}{lcccccccc}
\toprule 
Model             & \#Param     & Single Obj.$^\uparrow$ & Two Obj.$^\uparrow$ & Counting$^\uparrow$ & Colors$^\uparrow$ & Position$^\uparrow$ & Color Attri.$^\uparrow$ & Overall$^\uparrow$  \\ \midrule
\textcolor{gray}{GPT Image 1 {[}High{]}} \citep{openai_gpt_image_1_2025} & \textcolor{gray}{-} & \textcolor{gray}{0.99}        & \textcolor{gray}{0.92}     & \textcolor{gray}{0.85}     & \textcolor{gray}{0.92}   & \textcolor{gray}{0.75}     & \textcolor{gray}{0.61}         & \textcolor{gray}{0.84}    \\
\textcolor{gray}{Seedream 3.0} \citep{gao2025seedream}        &  \textcolor{gray}{-} & \textcolor{gray}{0.99}        & \textcolor{gray}{0.96}     & \textcolor{gray}{\textbf{0.91}}     & \textcolor{gray}{\textbf{0.93}}   & \textcolor{gray}{0.47}     & \textcolor{gray}{0.80}         & \textcolor{gray}{0.84}    \\
Qwen-Image \citep{wu2025qwen}       &   \textbf{20B}   & 0.99        & 0.92     & \textbf{0.89}     & 0.88   & 0.76     & 0.77         & 0.87    \\
\midrule
Emu3-Gen \citep{wang2024emu3}       &   8B   & 0.98        & 0.71     & 0.34     & 0.81   & 0.17     & 0.21         & 0.54    \\
SD3 Medium \citep{esser2024scaling} &    2B       & 0.98        & 0.74     & 0.63     & 0.67   & 0.34     & 0.36         & 0.62    \\
FLUX.1 {[}Dev{]} \citep{flux2024}   &  12B & 0.98        & 0.81     & 0.74     & 0.79   & 0.22     & 0.45         & 0.66    \\
SD3.5 Large \citep{esser2024scaling}    &   8.1B     & 0.98        & 0.89     & 0.73     & 0.83   & 0.34     & 0.47         & 0.71    \\
Lumina-Image 2.0 \citep{qin2025lumina}  &  2.6B  & -           & 0.87     & 0.67     & -      & -        & 0.62         & 0.73    \\
Show-O2 \citep{xie2025show}             &  7B & 1.00        & 0.87     & 0.58     & 0.92   & 0.52     & 0.62         & 0.76    \\
Janus-Pro \citep{chen2025janus}   &  7B     & 0.99        & 0.89     & 0.59     & 0.90   & 0.79     & 0.66         & 0.80    \\
SANA-1.5 \citep{xie2025sana}  &   4.8B     & 0.99        & 0.93     & 0.86     & 0.84   & 0.59     & 0.65         & 0.81    \\
HiDream-I1-Full \citep{cai2025hidream} &  17B  & 1.00        & 0.98     & 0.79     & 0.91   & 0.60     & 0.72         & 0.83    \\
TAR \citep{han2025vision} & 7B & 0.99 & 0.92 & 0.83 & 0.85 & 0.80 & 0.65 & 0.84 \\
Bagel  \citep{deng2025emerging}     &    14B       & 0.98        & 0.95     & 0.84     & \textbf{0.95}   & 0.78     & 0.77         & 0.88    \\
Mogao  \citep{liao2025mogao}        &    7B    & 1.00        & 0.97     & 0.83     & 0.93   & 0.84     & 0.80         & 0.89    \\
\midrule
\rowcolor{metabg}
MoS-S           & 3B      & 1.00        & 0.95     & 0.83     & 0.89   & 0.86     & \textbf{0.81}         & 0.89    \\
\rowcolor{metabg}
MoS-L        & 5B         & \textbf{1.00}        & \textbf{0.97}     & 0.82     & 0.91   & \textbf{0.88}     & 0.80         & \textbf{0.90}   
\\
\bottomrule
\end{tabular}
\end{table}
\begin{table}[!h]
\centering
    \footnotesize
\setlength{\tabcolsep}{1.0em}  
\renewcommand{\arraystretch}{1.0}
\caption{\textbf{Performance of Foundational Image Generation Models on the DPG Benchmark \citep{hu2024ella}.} DPG-Bench evaluates long-prompt alignment in text-to-image models. MoS-Image-L delivers state-of-the-art results across multiple dimensions, ranking just below Lumina.}
\label{tab:dpg_bench}
\begin{tabular}{lccccccc}
\toprule
Model        & \#Param          & Global$^\uparrow$ & Entity$^\uparrow$ & Attribute$^\uparrow$ & Relation$^\uparrow$                  & Other$^\uparrow$ & Overall$^\uparrow$                 \\
\midrule
\textcolor{gray}{DALL-E 3 \citep{openai_dalle3_2025} }   & \textcolor{gray}{-}          & \textcolor{gray}{90.97}  & \textcolor{gray}{89.61}  & \textcolor{gray}{88.39}     & \textcolor{gray}{90.58}                     & \textcolor{gray}{89.83} & \textcolor{gray}{83.50}                     \\
\textcolor{gray}{GPT Image 1 {[}High{]} \citep{openai_gpt_image_1_2025}} & \textcolor{gray}{-} & \textcolor{gray}{88.89}  & \textcolor{gray}{88.94}  & \textcolor{gray}{89.84}     & \textcolor{gray}{92.63}                     & \textcolor{gray}{90.96} & \textcolor{gray}{85.15}                     \\
\textcolor{gray}{Seedream 3.0  \citep{gao2025seedream} }     & \textcolor{gray}{-}    & \textcolor{gray}{\textbf{94.31}}  & \textcolor{gray}{\textbf{92.65}}  & \textcolor{gray}{91.36}     & \textcolor{gray}{92.78}                     & \textcolor{gray}{88.24} & \textcolor{gray}{88.27}                     \\
Qwen-Image \citep{wu2025qwen}    & \textbf{20B}         & 91.32  & 91.56  & 92.02     & 94.31                     & \textbf{92.73} & \textbf{88.32}                     \\
\midrule
SD v1.5 \citep{rombach2022high}     & 0.86B          & 74.63  & 74.23  & 75.39     & 73.49                     & 67.81 & 63.18                     \\
SDXL \citep{podell2023sdxl}        &  6.6B          & 83.27  & 82.43  & 80.91     & 86.76                     & 80.41 & 74.65                     \\
Playground v2.5 \citep{li2024playground}  & 6.6B     & 83.06  & 82.59  & 81.20     & 84.08                     & 83.50 & 75.47                     \\
Hunyuan-DiT \citep{li2024hunyuan}     & 1.5B      & 84.59  & 80.59  & 88.01     & 74.36                     & 86.41 & 78.87                     \\
PixArt-$\Sigma$  \citep{chen2024pixart}  &   0.6B         & 86.89  & 82.89  & 88.94     & 86.59                     & 87.68 & 80.54                     \\
BLIP-3o  \citep{chen2025blip3}      & 8B     & -      & -      & -         & -                         & -     & 81.60                     \\
Emu3-Gen \citep{wang2024emu3}     & 8B         & 85.21  & 86.68  & 86.84     & 90.22                     & 83.15 & 80.60                     \\
FLUX.1 {[}Dev{]} \citep{flux2024}  &  12B    & 74.35  & 90.00  & 88.96     & 90.87                     & 88.33 & 83.84                     \\
SD3 Medium  \citep{esser2024scaling}   & 2B        & 87.90  & 91.01  & 88.83     & 80.70                     & 88.68 & 84.08                     \\
Janus-Pro \citep{chen2025janus}   & 7B      & 86.90  & 88.90  & 89.40     & 89.32                     & 89.48 & 84.19                     \\
TAR \citep{han2025vision} & 7B     & 83.98   & 88.62  & 88.05     & 93.98                     & 84.86  & 84.19                     \\
Mogao  \citep{liao2025mogao}      & 7B       & 82.37  & 90.03  & 88.26     & 93.18                     & 85.40 & 84.33                     \\
HiDream-I1-Full \citep{cai2025hidream}    & 17B  & 76.44  & 90.22  & 89.48     & 93.74                     & 91.83 & 85.89                     \\
Show-o2-7B  \cite{xie2025show}     & 7B      & 89.00  & 91.78  & 89.96     & 91.81                     & 91.64 & 86.14                     \\
Lumina-Image 2.0 \citep{qin2025lumina} & 2.6B     & -      & 91.97  & 90.20     & \textbf{94.85}                     & -     & 87.20                     \\
\midrule
\rowcolor{metabg}
MoS-Image-S         & 3B        & 89.29  & \textbf{92.17}  & \textbf{92.09}     & 89.38                     & 90.18 & 86.33                     \\
\rowcolor{metabg}
MoS-Image-L         & 5B        & \textbf{91.74}  & 90.59  & 91.29     & 93.30 & 91.69 & 87.01 \\
\bottomrule
\end{tabular}
\end{table}
\begin{table}[!ht]
    \centering
        \footnotesize
    \setlength{\tabcolsep}{0.7em}  
    \renewcommand{\arraystretch}{1.0}
    \caption{\textbf{Performance on world knowledge reasoning with WISE \citep{niu2025wise}.}  
WISE evaluates complex semantic understanding and world knowledge in text-to-image generation.}
    \label{tab:wisescore}
    \begin{tabular}{lccccccccc}
    \toprule
 Model & \#Param  & Cultural$^\uparrow$  & Time$^\uparrow$     & Space$^\uparrow$    & Biology$^\uparrow$    & Physics$^\uparrow$ & Chemistry$^\uparrow$ & Overall$^\uparrow$ \\
    \midrule
\textcolor{gray}{GPT Image 1 [High] \citep{openai_gpt_image_1_2025}} & \textcolor{gray}{-} & \textcolor{gray}{0.81} & \textcolor{gray}{0.71} & \textcolor{gray}{0.89} & \textcolor{gray}{0.83} & \textcolor{gray}{0.79} & \textcolor{gray}{0.74} & \textcolor{gray}{0.80} \\
 Qwen-Image \citep{wu2025qwen} & \textbf{20B} & \textbf{0.62} & \textbf{0.63} & \textbf{0.77} & \textbf{0.57} & \textbf{0.75} & 0.40 & \textbf{0.62} \\
 \midrule
  VILA-U~\citep{wu2024vila} & 7B & 0.26 &0.33  & 0.37 &0.35  &0.39 &0.23 & 0.31\\
 SDv1.5~\citep{rombach2022high} & 0.86B & 0.34 & 0.35& 0.32&0.28 &0.29 &0.21 & 0.32\\
 Janus-Pro \citep{chen2025janus}   & 7B    & 0.30& 0.37& 0.49 & 0.36&0.42 &0.26 & 0.35 \\
 Emu3-Gen \citep{wang2024emu3}       &   8B  & 0.34 & 0.45 & 0.48 & 0.41  & 0.45 & 0.27 & 0.39 \\
 SDXL \citep{podell2023sdxl}        &  6.6B &0.43  & 0.48 &0.47  &0.44  &0.45 &0.27 & 0.43 \\
 SD3.5 Large \citep{esser2024scaling}    &   8.1B   & 0.44 &0.50 &0.58  & 0.44&0.52 &0.31 & 0.46 \\
 PixArt-Alpha~\citep{chen2024pixart} & 0.6B & 0.45  & 0.50& 0.48 & 0.49& 0.56 &0.34 & 0.47\\
 Playground v2.5 \citep{li2024playground}  & 6.6B  & 0.49  &0.58  & 0.55&0.43  & 0.48&0.33 & 0.49 \\
 FLUX.1 {[}Dev{]} \citep{flux2024}   &  12B & 0.48  &0.58 &0.62  &0.42  &0.51 & 0.35 & 0.50 \\
 BAGEL~\citep{deng2025emerging} & 14B & 0.44 & 0.55 & 0.68 & 0.44 & 0.60 & 0.39 & 0.52 \\
 UniWorld-V1~\citep{lin2025uniworld} & 12B  & 0.53 & 0.55 & 0.73 & 0.45 & 0.59	& 0.41 & 0.55 \\
 \midrule
 \rowcolor{metabg}
MoS-Image-S  & 3B & 0.40  & 0.50 & 0.65 & 0.43 & 0.63 & 0.37 & 0.47 \\
\rowcolor{metabg}
MoS-Image-L  & 5B & 0.47  & \underline{0.56} & \underline{0.74} & \underline{0.49} & \underline{0.64} & \textbf{0.44} & 0.54 \\
\bottomrule
    \end{tabular}
\end{table}

\begin{table}[!ht]
    \centering
    \footnotesize
    \setlength{\tabcolsep}{0.9em}  
    \renewcommand{\arraystretch}{1.0}
    \caption{\textbf{Quantitative results on OneIG~\citep{chang2025oneig}.} The overall score is averaged across five dimensions. With only 5B parameters, our model matches Imagen4 and trails recent commercial models by a small margin.}
    \label{tab:oneig}
    \begin{tabular}{lccccccc}
        \toprule
        Model & \# Param & Alignment$^\uparrow$& Text$^\uparrow$ & Reasoning$^\uparrow$ & Style$^\uparrow$ & Diversity$^\uparrow$ & Overall$^\uparrow$ \\
        \midrule
        \textcolor{gray}{Imagen3 \citep{baldridge2024imagen} } & \textcolor{gray}{-}&  \textcolor{gray}{0.84} & \textcolor{gray}{0.34} & \textcolor{gray}{0.31} & \textcolor{gray}{0.36} & \textcolor{gray}{0.19} & \textcolor{gray}{0.41} \\
        \textcolor{gray}{Kolors 2.0 \citep{kuai2025kolors2}}& \textcolor{gray}{-} & \textcolor{gray}{0.82} & \textcolor{gray}{0.43} & \textcolor{gray}{0.26} & \textcolor{gray}{0.36} & \textcolor{gray}{0.30} & \textcolor{gray}{0.43}\\
        \textcolor{gray}{Recraft V3 \citep{recraft2024v3}} & \textcolor{gray}{-} & \textcolor{gray}{0.81} & \textcolor{gray}{0.80} & \textcolor{gray}{0.32} & \textcolor{gray}{0.38} & \textcolor{gray}{0.21} &\textcolor{gray}{0.50}\\
        \textcolor{gray}{Imagen4 \citep{google2025imagen}} & \textcolor{gray}{-} & \textcolor{gray}{0.86} & \textcolor{gray}{0.81} & \textcolor{gray}{0.34} & \textcolor{gray}{0.38} & \textcolor{gray}{0.20} & \textcolor{gray}{0.52} \\
        \textcolor{gray}{Seedream 3.0 \citep{gao2025seedream}} & \textcolor{gray}{-} & \textcolor{gray}{0.82} & \textcolor{gray}{0.87} & \textcolor{gray}{0.28} & \textcolor{gray}{0.41} & \textcolor{gray}{0.28} & \textcolor{gray}{0.53} \\
        \textcolor{gray}{GPT Image 1 [High] \citep{openai_gpt_image_1_2025}} & \textcolor{gray}{-}  & \textcolor{gray}{0.85} & \textcolor{gray}{0.86} & \textcolor{gray}{\textbf{0.35}} & \textcolor{gray}{\textbf{0.46}} & \textcolor{gray}{0.15}& \textcolor{gray}{0.53}\\
        Qwen-Image \citep{wu2025qwen} & \textbf{20B}&  \textbf{0.88}  & \textbf{0.89} & \textbf{0.31} & \textbf{0.42} & 0.20 & \textbf{0.54} \\
        \midrule
        Janus-Pro \citep{chen2025janus}   & 7B      & 0.55  & 0.00  &   0.14     & 0.28 & 0.37 & 0.27\\
        BLIP3-o \citep{chen2025blip3} & 8B  & 0.71  & 0.01  &   0.22      & 0.36 & 0.23 & 0.31\\
        BAGEL \citep{deng2025emerging} & 14B & 0.77  & 0.24  &   0.17    & 0.37  & 0.25 & 0.36\\
        Show-o2 \citep{xie2025show} & 7B & 0.82 & 0.00 & 0.23 & 0.32 & 0.18 &0.31\\  
        SDv1.5~\citep{rombach2022high} & 0.86B & 0.57 & 0.01 & 0.21 & 0.38 & \textbf{0.43} & 0.32\\
        SDXL \citep{podell2023sdxl}        &  6.6B & 0.69 & 0.03 & 0.24 & 0.33 & 0.30 &0.32\\
        SANA-1.5\citep{xie2025sana} & 4.8B & 0.77 & 0.07 & 0.22 & 0.40 & 0.22 &0.33\\
        Lumina-Image 2.0 \citep{qin2025lumina} & 2.6B   & 0.82 & 0.11 & 0.27 & 0.35 & 0.22 & 0.35 \\
        SD3.5 Large \citep{esser2024scaling}    &   8.1B   & 0.81 & 0.63 & 0.29 & 0.35 & 0.23&0.46 \\
        FLUX.1 {[}Dev{]} \citep{flux2024}   &  12B & 0.79 & 0.52 & 0.25 & 0.37 & 0.24 &0.43 \\
        CogView4 \citep{zhipu2025cogview4} & 6B & 0.79 & 0.64 & 0.25 & 0.35 & 0.21 &0.45 \\
        OmniGen2 \citep{wu2025omnigen2} & 4B & 0.80 & 0.68 & 0.27 & 0.38 & 0.24 & 0.48 \\ 
        HiDream-I1-Full \citep{cai2025hidream}    & 17B  & 0.83 & 0.71 & 0.32 & 0.35 & 0.19 & 0.48 \\
        \midrule
         \rowcolor{metabg}
        MoS-Image-S & 3B & 0.82 & 0.82 & 0.26 & 0.38 & 0.20 & 0.50 \\
         \rowcolor{metabg}
        MoS-Image-L & 5B & \underline{0.85} & \underline{0.87} & 0.26 & \underline{0.41} & 0.19 & \underline{0.52} \\
        \bottomrule
    \end{tabular}
\end{table}
\begin{table}[!h]
    \centering
        \footnotesize
    \setlength{\tabcolsep}{0.15em}  
    \renewcommand{\arraystretch}{1.0}
\caption{\textbf{Performance of Foundational Image Editing Models on ImgEdit Benchmark \citep{ye2025imgedit}.} 
Greyed rows indicate models lacking clear configuration references, which prevents fair comparison.}    
    \label{tab:imgedit}
    \begin{tabular}{lccccccccccc}
        \toprule
        Model & \#Param  & Add$^\uparrow$ & Adjust$^\uparrow$ & Extract$^\uparrow$ & Replace$^\uparrow$ & Remove$^\uparrow$ & Back.$^\uparrow$ & Style$^\uparrow$ & Hybrid$^\uparrow$ & Action$^\uparrow$ & Overall$^\uparrow$ \\
        \midrule
        \textcolor{gray}{FLUX.1 Kontext {[}Pro{]}\citep{batifol2025flux}} & \textcolor{gray}{-} & \textcolor{gray}{4.25} & \textcolor{gray}{4.15} & \textcolor{gray}{2.35} & \textcolor{gray}{4.56} & \textcolor{gray}{3.57} & \textcolor{gray}{4.26} & \textcolor{gray}{4.57} & \textcolor{gray}{3.68} & \textcolor{gray}{4.63} & \textcolor{gray}{4.00} \\
        \textcolor{gray}{GPT Image 1 [High] \citep{openai_gpt_image_1_2025}} & \textcolor{gray}{-} & \textcolor{gray}{4.61} & \textcolor{gray}{4.33} & \textcolor{gray}{2.90} & \textcolor{gray}{4.35} & \textcolor{gray}{3.66} & \textcolor{gray}{4.57} & \textcolor{gray}{\textbf{4.93}} & \textcolor{gray}{3.96} & \textcolor{gray}{\textbf{4.89}} & \textcolor{gray}{4.20} \\
        Qwen-Image \citep{wu2025qwen} & \textbf{20B} & 4.38 & 4.16 & \textbf{3.43} & 4.66 & 4.14 & 4.38 & 4.81 & 3.82 & 4.69 & 4.27 \\ \midrule
        MagicBrush~\citep{zhang2023magicbrush} & 0.86B & 2.84 & 1.58 & 1.51 & 1.97 & 1.58 & 1.75 & 2.38 & 1.62 & 1.22 & 1.90 \\
        Instruct-Pix2Pix \citep{brooks2023instructpix2pix}& 0.86B & 2.45 & 1.83 & 1.44 & 2.01 & 1.50 & 1.44 & 3.55 & 1.20 & 1.46 & 1.88 \\
        AnyEdit~\citep{yu2025anyedit} &  0.86B & 3.18 & 2.95 & 1.88 & 2.47 & 2.23 & 2.24 & 2.85 & 1.56 & 2.65 & 2.45 \\
        UltraEdit~\citep{zhao2024ultraedit}&  2.5B & 3.44 & 2.81 & 2.13 & 2.96 & 1.45 & 2.83 & 3.76 & 1.91 & 2.98 & 2.70 \\
        OmniGen~\citep{xiao2025omnigen} & 3.8B & 3.47 & 3.04 & 1.71 & 2.94 & 2.43 & 3.21 & 4.19 & 2.24 & 3.38 & 2.96 \\
        ICEdit~\citep{zhang2025context} & 0.2B & 3.58 & 3.39 & 1.73 & 3.15 & 2.93 & 3.08 & 3.84 & 2.04 & 3.68 & 3.05 \\
        Step1X-Edit~\citep{liu2025step1x_edit} & 12B & 3.88 & 3.14 & 1.76 & 3.40 & 2.41 & 3.16 & 4.63 & 2.64 & 2.52 & 3.06 \\
        BAGEL~\citep{deng2025emerging} & 14B & 3.56 & 3.31 & 1.70 & 3.30 & 2.62 & 3.24 & 4.49 & 2.38 & 4.17 & 3.20 \\
        UniWorld-V1~\citep{lin2025uniworld} & 12B & 3.82 & 3.64 & 2.27 & 3.47 & 3.24 & 2.99 & 4.21 & 2.96 & 2.74 & 3.26 \\
        OmniGen2~\citep{wu2025omnigen2} & 4B & 3.57 & 3.06 & 1.77 & 3.74 & 3.20 & 3.57 & \textbf{4.81} & 2.52 & \textbf{4.68} & 3.44 \\
        \midrule
        \rowcolor{metabg}
        MoS-Image-S & 3B & 4.40 & 4.02 & 2.39 & 4.80 & 4.60 & 4.52 & 4.68 & 3.80 & 4.31 & 4.17 \\
        \rowcolor{metabg}
        MoS-Image-L & 5B & \textbf{4.63} & \textbf{4.47} & 2.04 & \textbf{4.85} & \textbf{4.73} & \textbf{4.85} & 4.71 & \textbf{4.16} & 4.52 & \textbf{4.33} \\
        \bottomrule
    \end{tabular}
\end{table}
\begin{table}
    \centering
        \footnotesize
    \setlength{\tabcolsep}{1.1em}  
    \renewcommand{\arraystretch}{1.0}
\caption{\textbf{Performance of Foundational Image Editing Models on GEdit Benchmark \citep{liu2025step1x_edit}.} 
Greyed rows indicate models lacking clear configuration references, which prevents fair comparison.}    
    \label{tab:gedit}
\begin{tabular}{lcccc}
\toprule
Model                         & \#param & G-Semantic Consistency$^\uparrow$ & G-Perceptual Quality $^\uparrow$ & G.-Overall$^\uparrow$ \\
                         \midrule
\textcolor{gray}{Gemini 2.0 \citep{google2025gemini2}}               &    \textcolor{gray}{-}     & \textcolor{gray}{6.73}               & \textcolor{gray}{6.61}                  & \textcolor{gray}{6.32}       \\
\textcolor{gray}{FLUX.1 Kontext {[}Pro{]}\citep{batifol2025flux}} &         & \textcolor{gray}{7.02}               & \textcolor{gray}{7.60}                  & \textcolor{gray}{6.56}       \\
\textcolor{gray}{GPT Image 1 [High] \citep{openai_gpt_image_1_2025}} & \textcolor{gray}{-}& \textcolor{gray}{7.85}               & \textcolor{gray}{7.62}                  & \textcolor{gray}{7.53}       \\
Qwen-Image \citep{wu2025qwen} & 20B & 8.00              & \textbf{7.86}                  & 7.56       \\
\midrule
Instruct-Pix2Pix \citep{brooks2023instructpix2pix}       &  0.86B       & 3.58               & 5.49                  & 3.68       \\
AnyEdit~\citep{yu2025anyedit} &  0.86B       & 3.18               & 5.82                  & 3.21       \\
MagicBrush~\citep{zhang2023magicbrush} & 0.86B    & 4.68               & 5.66                  & 4.52       \\
UniWorld-V1~\citep{lin2025uniworld} & 12B      & 4.93               & 7.43                  & 4.85       \\
OmniGen~\citep{xiao2025omnigen} & 3.8B        & 5.96               & 5.89                  & 5.06       \\
OmniGen2~\citep{wu2025omnigen2} & 4B     & 7.16               & 6.77                  & 6.41       \\
BAGEL~\citep{deng2025emerging} & 14B    & 7.36               & 6.83                  & 6.52       \\
Step1X-Edit~\citep{liu2025step1x_edit}              &     12B    & 7.66               & 7.35                  & 6.97       \\ \midrule
\rowcolor{metabg}
MoS-Image-S                   &   3B      &        8.00            &        7.34               &    7.41        \\
\rowcolor{metabg}
MoS-Image-L                   &   5B      &       \textbf{8.54}             &        7.64              &     \textbf{7.86}       \\
\bottomrule
\end{tabular}
\end{table}

\begin{algorithm}[t]
\caption{Training procedure of MoS}
\label{alg:mos_training}
\begin{algorithmic}[1]
\Require Paired training data $(z_0, c)$; understanding tower $\mathcal{U}$; generation tower $\mathcal{G}$; router $\mathcal{R}$; number of layers $m, n$; top-$k_\epsilon$ selection function.
\Statex

\State \textbf{1. Encode input.}
\State Extract hidden states from the understanding tower:
\[
\mathcal{U}(c) = \{\mathcal{S}^c_i \mid i \in [1, m]\}
\]

\State \textbf{2. Sample diffusion step.}
\State Randomly sample timestep $t \sim \text{Uniform}(1, T)$ and obtain noisy latent $z_t$.

\State \textbf{3. Predict routing weights.}
\State Compute router logits:
\[
\mathcal{W} = \mathcal{R}(c, t, z_t)
\]
where $\mathcal{W} \in \mathbb{R}^{m \times n}$ and $w_{ij}$ denotes the routing weight from the $i$-th understanding layer to the $j$-th generation layer.

\State Normalize logits: 
\[
\overline{\mathcal{W}} = \text{softmax}(\mathcal{W})
\]
where the softmax operation is applied to each column $w_{1:m,j}$ of $\mathcal{W}$.

\State \textbf{4. Construct conditional signal for each generation block.}
\For{$j = 1$ to $n$}
    \State Select indices of top-$k$ elements under $\epsilon$-greedy rule:
    \[
    I_j = \text{top-}k_\epsilon(\overline{w}_{1:m, j})
    \]
    \State Compute the conditional context:
    \[
    \mathbf{S}^c_j = \sum_{i \in I_j} \overline{w}_{ij} \cdot \mathcal{S}^c_i.
    \]
    \State Fuse with the generation tower features:
    \[
    \mathbf{H}_j = \text{Concat}\big(\text{Proj}(\mathbf{S}^c_j), \mathbf{S}^z_j\big)
    \]
    \State Perform Generation Tower Block-$j$ on $\mathbf{H}_j$ to update $\mathbf{S}^z_j$ to $\mathbf{S}^z_{j+1}$.
\EndFor

\State \textbf{5. Compute loss.}
\State Apply the diffusion objective (e.g., $\ell_2$ loss) between predicted and ground-truth $z_0$.
\end{algorithmic}
\end{algorithm}
\begin{algorithm}[t]
\caption{Inference procedure of MoS}
\label{alg:mos_inference}
\begin{algorithmic}[1]
\Require Conditioning input $c$; understanding tower $\mathcal{U}$; generation tower $\mathcal{G}$; router $\mathcal{R}$; number of steps $T$; number of top connections $k$; number of layers $m, n$.
\Ensure Generated sample $\hat{z}_0$
\Statex

\State \textbf{1. Encode conditioning signal.}
\State Obtain hidden states from the understanding tower:
\[
\mathcal{U}(c) = \{\mathcal{S}^c_i \mid i \in [1, m]\}.
\]

\State \textbf{2. Initialize latent.}
\State Sample initial noise $\hat{z}_1 \sim \mathcal{N}(0, \mathbf{I})$.

\State \textbf{3. Iterative denoising.}
\For{$t = 1$ down to $0$}
    \State Compute router logits:
    \[
    \mathcal{W} = \mathcal{R}(c, t, \hat{z}_t)
    \]
    \State Normalize logits: 
    \[
    \overline{\mathcal{W}} = \text{softmax}(\mathcal{W})
    \]
    \For{$j = 1$ to $n$}
        \State Select top-$k$ indices under $\epsilon$-greedy rule:
        \[
        I_j = \text{top-}k_\epsilon(\overline{w}_{1:m, j})
        \]
        \State Compute the conditional context:
        \[
        \mathbf{S}^c_j = \sum_{i \in I_j} \overline{w}_{ij} \cdot \mathcal{S}^c_i.
        \]
        \State Fuse with the generation tower features:
        \[
        \mathbf{H}_j = \text{Concat}\big(\text{Proj}(\mathbf{S}^c_j), \mathbf{S}^z_j\big)
        \]
        \State Perform Generation Tower Block-$j$ on $\mathbf{H}_j$ to update $\mathbf{S}^z_j$ to $\mathbf{S}^z_{j+1}$.
    \EndFor
    \State Predict $\hat{z}_{t-1}$ with $ \hat{v}_t =\mathcal{G}(z_t, \mathcal{W}, \mathcal{U}(c))$ following the diffusion sampling rule.
\EndFor

\State \textbf{4. Decode.}
\State Obtain final output $\hat{z}_0$ via the decoder of $\mathcal{G}$.
\end{algorithmic}
\end{algorithm}
\begin{figure}
    \centering
    \includegraphics[width=.98\linewidth]{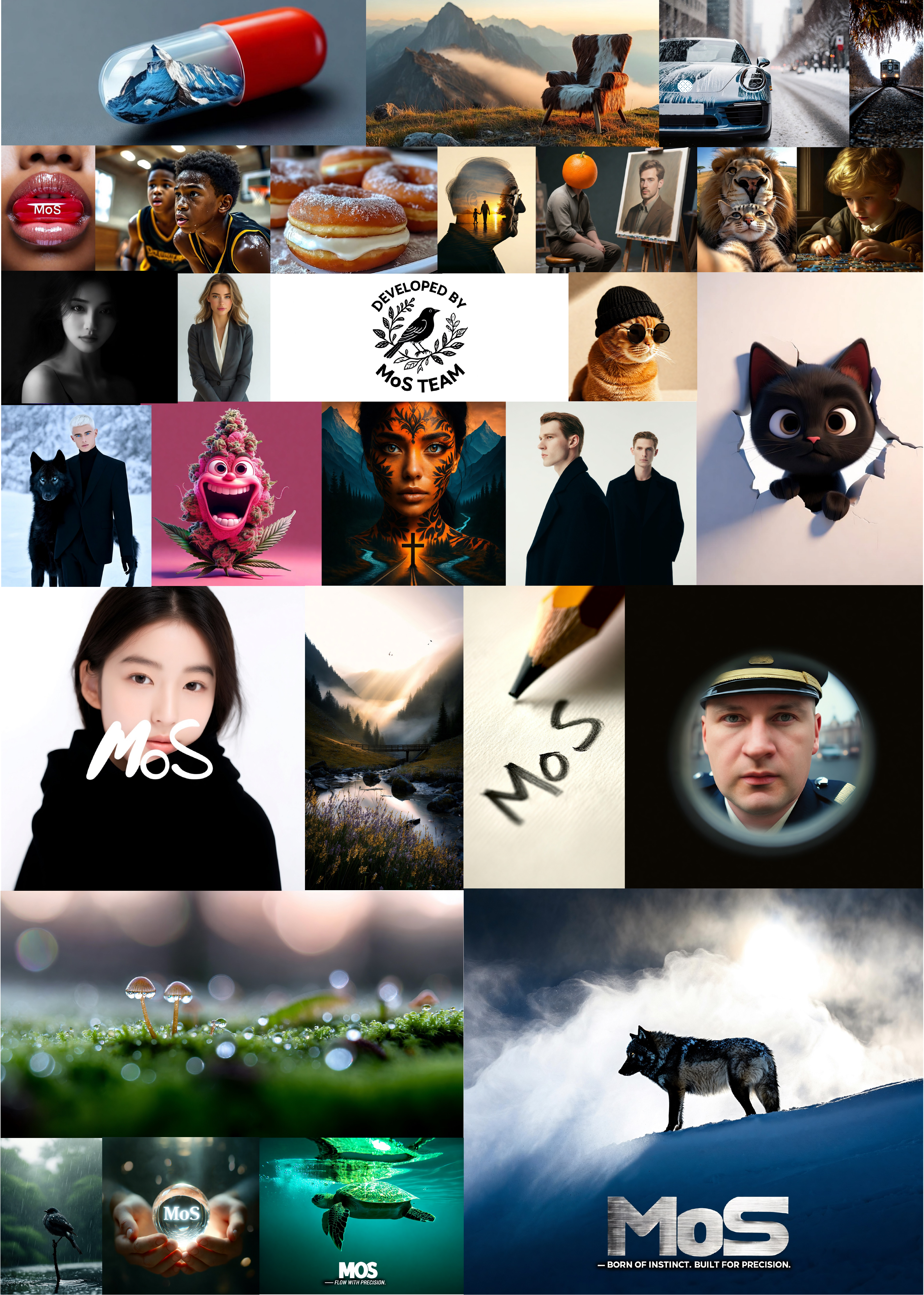}
    \caption{\textbf{Visualization of MoS-L on text-to-image generation.} The samples are produced under a dynamic resolution setting, with the maximum side length capped at 2048 pixels. }
    \label{fig:vis_t2i}
    \vspace{-2em}
\end{figure}
\begin{figure}
\centering
\includegraphics[width=.95\linewidth]{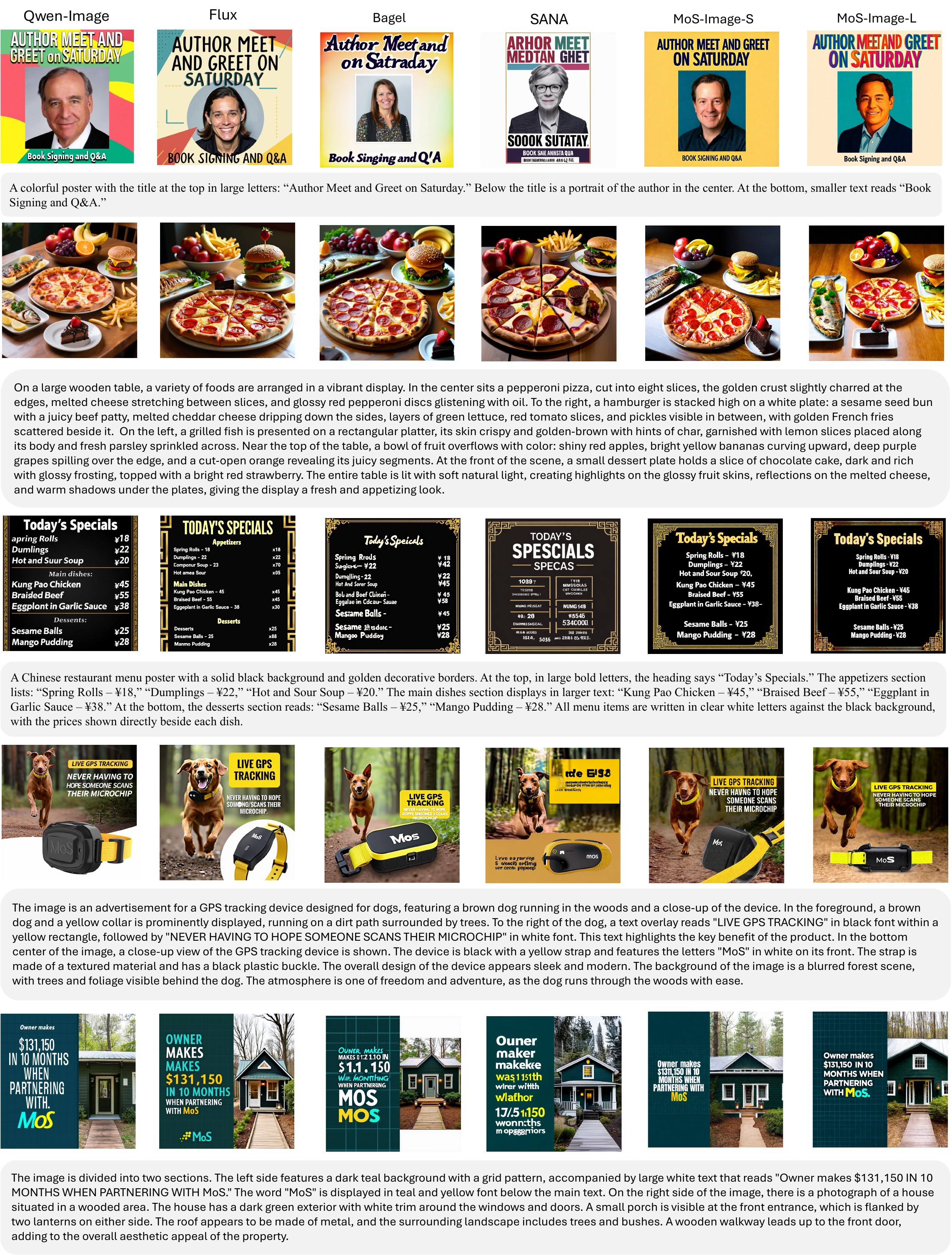}
\caption{\textbf{Visualization of MoS-L/S and baseline methods on text-to-image generation.} All models are evaluated with their default parameters in \texttt{Diffusers}. We present results on challenging cases. These include scenarios such as arranging foods with distinct categories, colors, and patterns; posters combining natural objects with visual text; and purely textual prompts, such as generating a menu. MoS-L demonstrates competitive performance in these demanding settings. Zoomed-in view for better clarity.}
    \label{fig:vis_t2i_comapare_with_others}
\end{figure}
\begin{figure}
\centering
\includegraphics[width=.95\linewidth]{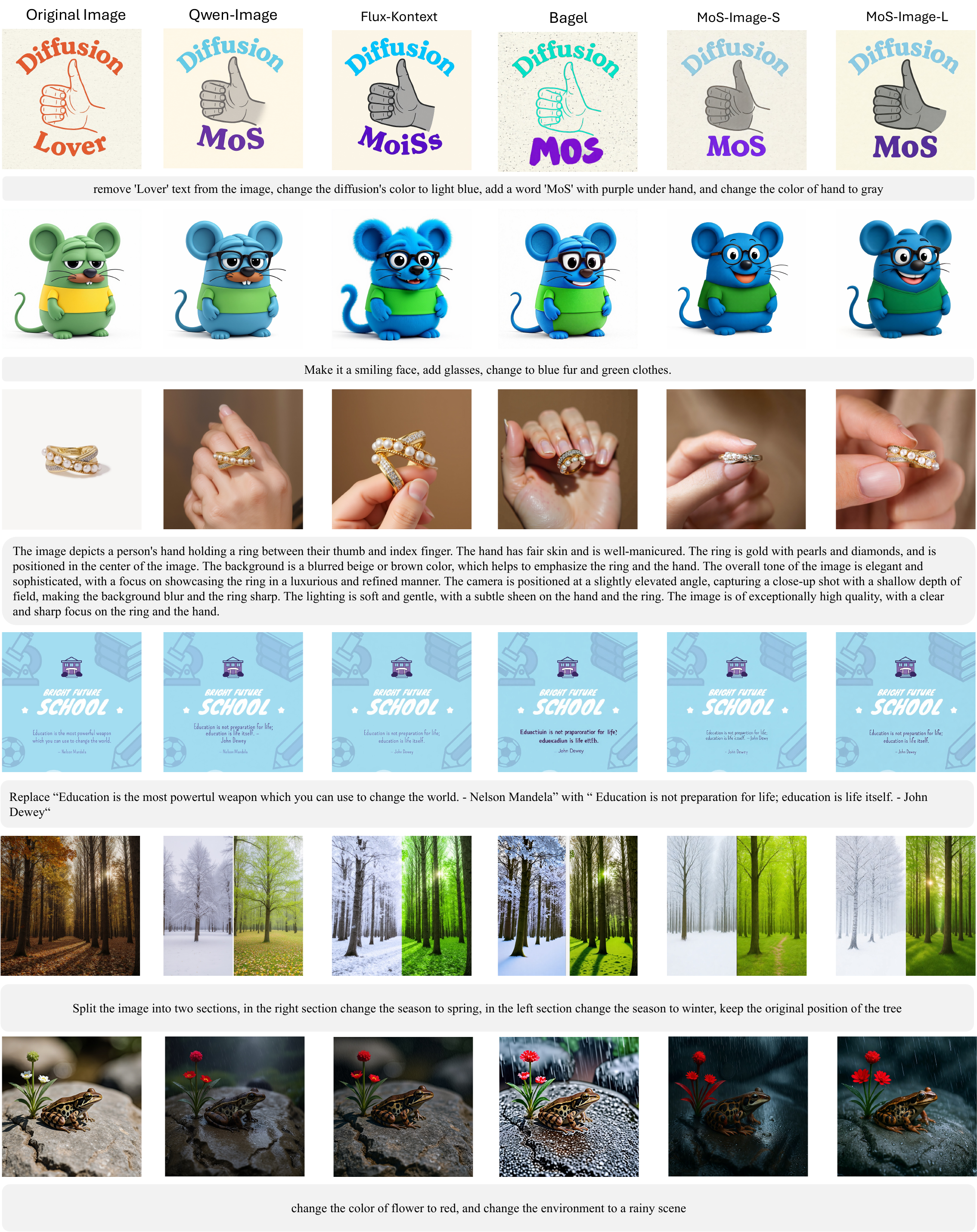}
\caption{\textbf{Visualization of MoS-L/S and baseline methods on instruction-based image editing.} All models are evaluated using their default parameters in \texttt{Diffusers}. We showcase results on hybrid instructions and the cases involving visual text editing. Zoomed-in for better clarity.}
    \label{fig:vis_edit_comapare_with_others}
\end{figure}
\end{document}